%% file: main.tex
\documentclass[10pt]{article} % For LaTeX2e
\usepackage[preprint]{tmlr}
\input{math_commands.tex}

\usepackage{hyperref}
\usepackage{url}
\usepackage{booktabs} 
\usepackage{adjustbox}
\usepackage{enumitem}
\usepackage{subcaption}
\usepackage{longtable}
\usepackage[most]{tcolorbox} % preamble

\title{Evaluating LLM Understanding via Structured \\ Tabular Decision Simulations}

\author{\name Sichao Li \thanks{Equal contribution.} \thanks{Corresponding author.} \email scli@city.edu.mo \\
      \addr City University of Macau  
      \AND
      \name Xinyue Xu \footnotemark[1] \email xinyue.xu@connect.ust.hk \\
      \addr The Hong Kong University of Science and Technology
      \AND
      \name Xiaomeng Li \email eexmli@ust.hk\\
      \addr The Hong Kong University of Science and Technology
      }

\def\month{MM}  % Insert correct month for camera-ready version
\def\year{YYYY} % Insert correct year for camera-ready version
\def\openreview{\url{https://openreview.net/forum?id=XXXX}} % Insert correct link to OpenReview for camera-ready version

\newtheorem{theorem}{Theorem}[section]

\newtheorem{remark}[theorem]{Remark}

\begin{document}

\maketitle

\begin{abstract}
Large language models (LLMs) often achieve impressive predictive accuracy, yet correctness alone does not imply genuine \textit{understanding}. True LLM understanding, analogous to human expertise, requires making consistent, well-founded decisions across multiple instances and diverse domains, relying on relevant and domain-grounded decision factors.
We introduce \textbf{Structured Tabular Decision Simulations (STaDS)}, a suite of expert-like decision settings that evaluate LLMs as if they were professionals undertaking structured decision ``exams''.
In this context, \textbf{understanding} is defined as the ability to identify and rely on the correct \textbf{decision factors}, features that determine outcomes within a domain.
STaDS jointly assesses understanding through: (i) question and instruction comprehension, (ii) knowledge-based prediction, and (iii) reliance on relevant decision factors.
By analyzing 9 frontier LLMs across 15 diverse decision settings, we find that (a) most models struggle to achieve consistently strong accuracy across diverse domains; (b) models can be \textit{accurate yet globally unfaithful}, and there are frequent mismatches between stated rationales and factors driving predictions. Our findings highlight the need for global-level understanding evaluation protocols and advocate for novel frameworks that go beyond accuracy to enhance LLMs' understanding ability.
\end{abstract}

\section{Introduction}\label{sec:intro}
Large language models (LLMs) are increasingly deployed as \textbf{surrogate professionals} due to their strong predictive performance, acting as physicians for medical triage, analysts for financial risk assessment, or policy advisors for legislative decisions \cite{abd2023large, brown2020language, dong2022survey, zhao2023survey}. In such applications, users expect models to reason and make decisions with the reliability of domain experts. Yet current evaluations overwhelmingly focus on surface metrics like accuracy or task completion. What is largely missing is an assessment of the model's \textit{understanding ability}: its internal competence to grasp and apply the principles that govern a decision task, going \textit{beyond making a single correct prediction}.  Understanding in this context is cognitive rather than purely behavioral, which makes it challenging to measure explicitly.

\begin{figure}
    \centering
    \includegraphics[width=0.8\linewidth]{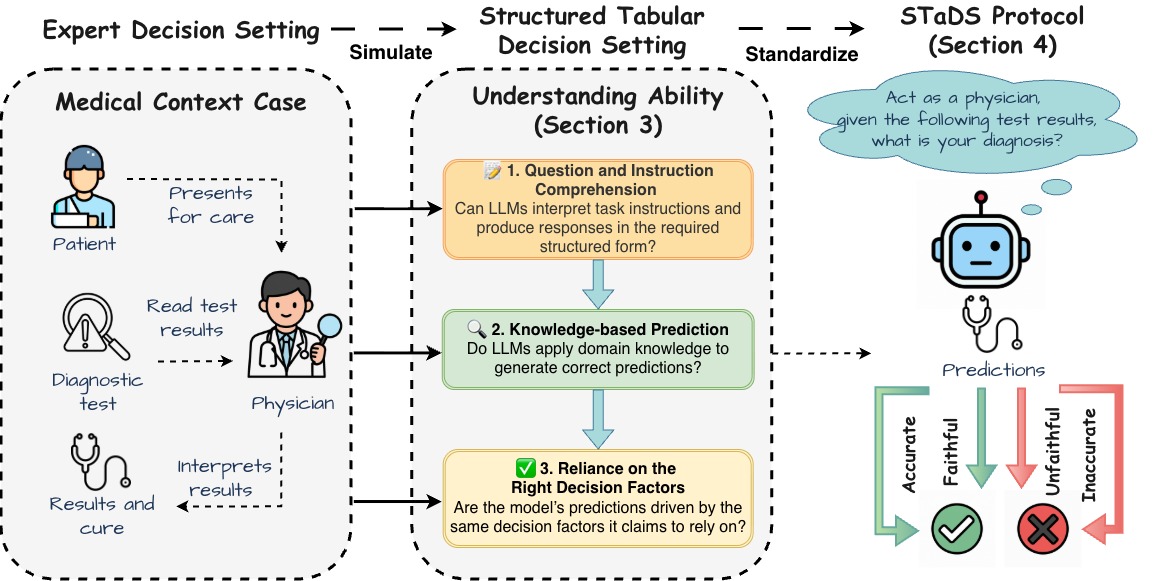}
    \caption{This diagram illustrates how the STaDS protocol simulates expert decision-making processes in structured tabular decision settings. The protocol evaluates LLMs' understanding ability through three key dimensions: (1) Question and Instruction Comprehension, assessing task interpretation and output adherence; (2) Knowledge-based Prediction, evaluating the model's application of domain knowledge for accurate predictions; and (3) Reliance on the Right Decision Factors, determining whether predictions align with the factors the model claims to rely on. The diagram depicts how these dimensions together form a principled basis for understanding and evaluating LLMs.}
    \label{fig:motivation}
\end{figure}

While recent work has begun to ask whether LLMs ``reason faithfully'', especially through chain-of-thought (CoT) rationales, such analyses mostly operate at the level of a single problem instance: does the explanation accompanying an answer reflect the steps that actually produced it? \citep{wei2022chain, lewkowycz2022solving, barez2025chain, yu2024natural} These studies expose important failures of \textbf{\textit{local}} reasoning faithfulness, but they do not tell us whether a model behaves like a reliable expert across many decisions in a domain \citep{jacovi2020towards, arcuschin2025chain}. In this work, we extend the concept of faithfulness from local reasoning traces to \textbf{\textit{global}} decision faithfulness, exploring whether the model's predictions across multiple cases consistently rely on a governing rule, driven by meaningful decision factors, rather than superficial correlations (see our motivation in Fig.~\ref{fig:motivation}).

\textbf{This limitation motivates a shift in perspective.} We propose evaluating the \textbf{understanding ability} of LLMs as a distinct axis of competence by simulating human-like decision settings, through the lens of interpretability and explainability. 
\begin{quote}
    By understanding ability, we refer to the capacity of an LLM to capture the underlying concepts of a task, generalize across diverse instances, and base its decisions on conceptually meaningful features \citep{mayer1989models, bereiter2005education}.
\end{quote}
Human experts exemplify this ability: they demonstrate understanding not merely by producing a correct answer once, but by consistently applying underlying domain concepts across diverse cases. For instance, a physician is expected to base diagnoses on established medical knowledge rather than incidental correlations. Similarly, a reliable LLM should leverage its pre-learned knowledge representations to generate well-founded, conceptually grounded decisions.

To probe this dimension, we introduce \textbf{Structured Tabular Decision Simulations (STaDS)} protocol: a systematic evaluation framework that casts expert-like decision problems into tabular form (details see Sec.~\ref{subsec:why-tab}), enabling \textit{controlled} assessment of both predictive performance and explanation faithfulness. Unlike reasoning, which emphasizes step-by-step justifications where models may make errors on intermediate steps \citep{turpin2023language}, tabular simulations focus on an end-to-end evaluation on whether LLMs capture the underlying decision rules or policies that govern outcomes (\textit{interpretability}). 
Moreover, tabular features and labels are unambiguous and semantically well-defined: each corresponds to a clearly grounded concept (e.g., disease diagnosis, loan default, voting preference) that has direct meaning in the application domain. This stands in contrast to \textit{free-text rationales}, which may admit multiple plausible interpretations, or \textit{image-based tasks}, where concepts are often loosely bounded and harder to specify, and typically require extensive labeling, human validation, and additional grounding efforts to ensure \textit{explainability} \citep{kim2024ambiguous, zarlenga2023tabcbm, li2023evaluating}. In this way, the setting enables quantitative assessment of understanding ability.

\noindent \textbf{Research Questions and Contributions.}
Ultimately, we target two central research questions:
\begin{quote}
    \textit{\textbf{RQ1:} To what extent do LLMs demonstrate understanding ability by generalizing across diverse tabular decision settings, beyond surface-level correlations?}
\end{quote}

\begin{quote}
   \textit{\textbf{RQ2:} To what extent do LLMs demonstrate globally faithful behavior, consistently identifying the decision factors that govern outcomes within a domain?}
\end{quote}

\noindent To answer these questions, we present a unified evaluation framework and empirical study of LLMs' understanding ability. Our primary contributions are as follows:
\begin{itemize}
    \item \textbf{STaDS Protocol:} We propose the \textbf{Structured Tabular Decision Simulation (STaDS)} protocol as a comprehensive evaluation of LLMs' understanding ability through structured, expert-like decision settings. STaDS operationalizes understanding along three complementary axes: (i) \textit{Question and instruction comprehension}: the ability to interpret task instructions and follow structured output specifications; (ii) \textit{Knowledge-based prediction}: the capacity to apply intrinsic and in-context knowledge to produce accurate predictions; and (iii) \textit{Reliance on the right decision factors}: the degree to which a model's predictions are driven by the same decision factors it claims to rely on. Unlike conventional evaluation frameworks, STaDS enables an end-to-end assessment that minimizes prompt bias and avoids errors from unfaithful reasoning steps.
    It provides a structured, reproducible, and extensible setting for LLMs, bridging concepts of interpretability (internal mechanisms) and explainability (post-hoc justifications) from eXplainable AI (XAI) to LLM evaluation \citep{bender2021dangers}. 
    \item \textbf{STaDS Metrics and Benchmarks.} Building on these three axes, we define targeted metric suites for each: (1) \textbf{Comprehension Fidelity}: captured by Len-F1, UnkLbl\%, and the format-related component of Penalized Accuracy to quantify instruction adherence; (2) \textbf{Predictive Competence}: measured through zero/few-shot Accuracy, Macro-F1, and overall Penalized Accuracy to assess knowledge grounding; and (3) \textbf{Decision Faithfulness}: evaluated via LAO-based feature reliance, Self-Decision Faithfulness, and SelfAtt@k to determine whether stated and actual decision factors align at the global level. To support systematic evaluation, we release a curated suite of \textit{15} real-world tabular datasets spanning healthcare, finance, and public policy, summing up to \textit{160k} tasks, where each task represents a real-world decision environment. 
    This is accompanied by \textit{standardized instruction templates} designed to minimize prompt bias and promote reproducibility.  
    \item \textbf{Empirical Insights.} We conduct a large-scale study of \textit{9} state-of-the-art LLMs, including advanced closed-source models (GPT, Gemini \citep{gpt4, gemini}) and leading open-source models (LLaMA, Mistral, DeepSeek, Qwen, Gemma), across all benchmarks \citep{llama3, deepseek, qwen, gemma}. Our analysis reveals a spectrum of behaviours, from models that are \textit{neither accurate nor faithful}, to those that are \textit{accurate but unfaithful}, and a small subset that achieve \textit{both accuracy and faithfulness}, highlighting persistent challenges in building understanding ability in LLMs.  
\end{itemize}

\section{Background \& Related Work}
We first differentiate the notion of \emph{understanding} in LLMs from existing terms:

\paragraph{Explainability \& Interpretability for LLMs.}
Explainability involves providing \textit{human-understandable justifications} for model decisions, typically through post-hoc methods linking inputs to outputs \citep{adadi2018peeking, li2022interpretable}. Interpretability, conversely, emphasizes \textit{transparent internal mechanisms} such as weights and attention interactions, making the model's reasoning intrinsically comprehensible \cite{ das2020opportunities, ali2023explainable}. In the LLM setting, explainability has been pursued through prompt-based rationales \citep{liu2023towards}, contrastive saliency for token-level influence \citep{min2023factscore}, and gradient-free feature attribution methods for structured tasks \citep{sui2024table}. While these approaches generate plausible explanations, they often fail to guarantee \emph{faithfulness}, i.e., alignment with the actual causal drivers of model predictions \citep{jacovi2020towards, agarwal2022openxai}. Evaluation practices also remain fragmented, relying on human judgments or narrow benchmarks. STaDS bridges these gaps by adapting concepts from XAI (eXplainable AI) to LLMs in a structured and model-agnostic way.
\textit{External} feature attributions are evaluated via LAO to capture human-understandable justifications, which are then compared to internally \textit{stated reasons} for agreement.

\paragraph{Reasoning \& Unfaithfulness.} Chain-of-thought prompting is widely used to evaluate logical reasoning in free-form text \citep{wang2024chain}, where advanced models generate step-by-step intermediate reasoning. However, such reasoning can be unreliable: models may produce errors in intermediate steps or exploit latent shortcuts that yield correct answers for the wrong reasons. Recent studies further highlight evidence of \emph{unfaithfulness} in both thinking and non-thinking frontier models, showing that they sometimes provide correct or incorrect answers accompanied by fabricated (Implicit Post-Hoc Rationalization) or illogical justifications (Unfaithful Illogical Shortcuts) \citep{barez2025chain, lanham2023measuring, arcuschin2025chain}. We distinguish this reasoning-level unfaithfulness from the \textit{behavioral inconsistency} targeted in STaDS, which captures differences between what a model claims and how it actually decides.
Rather than examining intermediate reasoning steps, STaDS evaluates whether a model’s \emph{global feature reliance} aligns with its stated attributions, shifting the focus from local reasoning chains to end-to-end decision faithfulness.

\paragraph{Step-level Reasoning \& Global Attribution Reasoning.} \cite{arcuschin2025chain} evaluates \emph{Unfaithful Illogical Shortcuts} through three steps: answer correctness, step criticality, and step unfaithfulness. Their analysis centers on intermediate reasoning steps that accumulate toward a \textbf{single} decision, which makes models vulnerable to biases or fabricated justifications at the \textbf{step level}. In contrast, STaDS deliberately avoids over-reliance on step-level reasoning. Instead, it emphasizes a holistic perspective: evaluating whether a model's \textbf{overall attribution ranking}, elicited via self-attribution, faithfully reflects the features governing a \textbf{set} of decisions. We acknowledge that self-stated attribution rankings may not perfectly reflect an LLM’s global attribution knowledge. 

% To alleviate this concern, we apply prompt sampling. Further discussion and empirical results are provided in Sec.~\ref{}.

\paragraph{Tabluar Relevant Tasks.}
A growing body of work develops table-centric models tailored for structured data, including TAPAS \citep{herzig2020tapas}, TURL \citep{deng2022turl},  TableLlama \citep{zhang2023tablellama}, and TabPFN \citep{hollmann2025tabpfn}. These models support tasks such as entity linking, column annotation, and fact extraction \citep{deng2022turl, zhang2023tablellama}, and more broadly span table interpretation, augmentation, question answering, fact verification, and dialogue generation. 
While these methods advance training efficiency and correctness across diverse applications, they often rely on specialized architectures restricted to particular table formats or tasks \citep{sui2024table}. By contrast, STaDS treats tabular data as a \emph{probing decision simulation setting}, where explicit columns enable controlled ablations, systematic perturbations, and unambiguous attribution scoring. Crucially, STaDS is \textbf{not intended as} another table-specific architecture, but rather as a complementary evaluation framework that leverages tabular data to probe whether LLMs behave like it states, beyond merely producing correct answers
\footnote{Recent table-specific models such as TableLlama, built on LLaMA-2 (7B), can handle contexts of up to 8K tokens, yet remain unable to process the longer tabular inputs considered in this work.}. 
% Specifically, table interpretation seeks to uncover semantic attributes of relational tables and transform them into machine-understandable knowledge; table augmentation extends partial tables with additional rows or columns; table question answering retrieves answers with optional highlighted cells or passages as evidence; fact verification determines whether tabular data supports or refutes a claim; and table-grounded dialogue generation produces conversational responses based on table content and dialogue history.

\paragraph{In-Context Learning (ICL).}
ICL enables LLMs to solve new tasks by \textit{learning} from demonstrations provided in prompts without parameter updates, rather than focus on understanding \cite{dong2022survey, brown2020language}. Follow-up work probes sensitivities for understanding ICL by measuring influence factors, such as demonstration selection, order, and formatting \cite{wang2023label, wei2023larger, akyurek2022learning, min2022rethinking}, which can be improved by our framework. These efforts highlight that ICL competence is fragile, but they rarely interrogate whether predictions are faithful to underlying reasoning. Moreover, ICL generally evaluates the likelihood of a \textbf{single} candidate answer, whereas STaDS systematically queries a \textbf{sequence} of labels, offering a more comprehensive measure of understanding.

\paragraph{Benchmark Landscape.}
Existing benchmark suites typically assess competence (e.g., GLUE, MMLU) \citep{wang2018glue, hendrycks2020measuring}, reasoning (e.g., GSM8K, DROP) \citep{cobbe2021training, dua2019drop}, or explanation plausibility (e.g., ERASER, e-SNLI) \citep{deyoung2019eraser, camburu2018snli} in isolation. However, none addresses the interplay among understanding, interpretability, and explainability \emph{within} an controllable decision simulation setting. STaDS fills this final gap by introducing the first benchmark that spans diverse real-world tabular domains while jointly evaluating predictive competence, attributional faithfulness, and explanation agreement at a global level.

\section{STaDS Protocol: Formalizing Understanding with Tabular Decision Simulations}

We introduce the STaDS protocol as a systematic evaluation framework for evaluating the understanding ability of LLMs. 
\subsection{What is Understanding?}\label{subsec:what-is-understanding}
Understanding is a broad and abstract notion. 
It has been described as \textit{``a cognitive process in which concepts are used to model an abstract or physical object, establishing a relation between the knower and the object of understanding by \citeauthor{mayer1989models}''}. Understanding implies abilities and dispositions with respect to an object of knowledge that are sufficient to support intelligent behavior \citep{ bereiter2005education}. We adopt this concept and refer understanding to as: 
\begin{quote}
    \textbf{The model's internal competence to grasp and apply the underlying concepts and principles that govern a decision task.}
\end{quote}

We characterize understanding as a multidimensional capacity from cognitive science and we therefore decompose understanding into three dimensions:

\begin{enumerate}
    \item \textbf{Question and instruction comprehension}: This dimension is the ability to correctly \textbf{interpret a task}: to recognize what is being requested, identify the goal state, and determine the appropriate form of response. In cognitive terms, this is often described as constructing a situation model or problem representation ``\textit{what is going on in this task, and what is being asked of me?}'' \cite{chi2014nature}. Successful comprehension requires mapping linguistic instructions to an internal representation of required actions, not just decoding words. In human learners, failure at this stage (misreading the question, misunderstanding constraints) is considered a failure of understanding even before any attempt at problem solving \cite{chi1981categorization}. Interpreting and adhering to instructions is therefore evidence of understanding at the level of task framing: the LLMs (knower) demonstrates it knows what problem it is solving and what form a valid answer should take.  
    \item \textbf{Knowledge-based prediction}: This dimension is the capacity to apply relevant prior knowledge to produce correct inferences or decisions in a new context. This corresponds to what is classically called transfer in cognitive science, the ability to take learned knowledge or principles from one setting and apply them appropriately in another \cite{bransford2000people}. Transfer is widely treated as a defining marker of genuine understanding, because it shows that performance is not tied to rote pattern matching or memorized responses but instead reflects grasp of underlying relationships and principles. Under this view, producing accurate, generalizable predictions across varied instances is behavioral evidence that the system can align internal knowledge with the current decision context productively, and use that alignment productively, known as a \textit{hallmark of deep rather than superficial understanding} \cite{chi2014nature}. 
    \item \textbf{Reliance on the right decision factors}: This dimension is the extent to which the model's decisions are driven by the same task-relevant factors it identifies as important. Expert performance is not only accurate; it is principled. Experts are able to justify their decisions by referencing structurally relevant features of the situation, and those justifications reflect the same internal criteria they actually used to make the decision \cite{chi1981categorization}. Novices, by contrast, often give explanations that are either post hoc, superficial, or anchored to salient but non-diagnostic surface cues. The alignment between (i) what an LLM claims matters and (ii) what actually drives its choices is therefore treated as evidence of decision faithfulness: it indicates that the LLM is guided by appropriate conceptual features of the problem space, rather than by opportunistic heuristics or randomness.
\end{enumerate}
Taken together, these three dimensions mirror how human understanding is evaluated in cognitive science and educational assessment. A competent human decision-maker is expected to demonstrate \textit{all} and we adopt the same structure when assessing LLM understanding.

\subsubsection{What are Decision Factors?}
In structured decision settings, decision factors are \textit{the explicit, semantically grounded variables that influence outcomes within a domain}.
Each factor corresponds to a domain-grounded explanatory variable, such as age, income, or tumor size, that represents part of the evidence a competent decision-maker would consider.
Decision factors are thus the building blocks of rational decision-making: \textit{they encode the domain's causal or normative structure and collectively define the reasoning space in which expertise operates.}
Hence, evaluating understanding in STaDS means probing whether the model correctly identifies, prioritizes, and relies on the appropriate decision factors, mirroring how human experts justify their judgments with causally meaningful reasoning.

\subsection{Why Tabular Decision Simulations?}\label{subsec:why-tab}
Tabular decision simulations provide a principled setting for evaluating LLMs’ understanding ability. Their design offers several advantages over other data formats:
\begin{enumerate}
    \item \textbf{Instance-level structure.} Each row corresponds to a complete, self-contained decision instance: the set of feature values in that row provides all the information required to determine an outcome. This framing mirrors how experts make case-by-case judgments in real-world domains. Crucially, it prevents models from exploiting dataset-level artifacts or spurious correlations, since each prediction must be grounded in the features of a single case. 
    Unlike tasks in vision or other multimodal reasoning, which often require additional perceptual processing or contextualization, tabular data provides a uniquely ``decision-ready'' input format. Each row presents a well-bounded context, with explicit and interpretable features, and an associated ground-truth label.

    \item \textbf{Global-level faithfulness.} Because all attributes are explicitly named, defined, and consistently shared across rows, tabular data naturally support analysis of \textit{global feature importance}, an established goal in XAI tasks \cite{Samek2017Explain, ali2023explainable}. This \textbf{distinguishes} tabular simulations from faithfulness evaluation in conventional reasoning tasks, where faithfulness is typically examined at the level of (i) individual decisions with (ii) intermediate reasoning steps. In contrast, tabular simulations enable evaluation of whether a model’s decision-making aligns with coherent, domain-wide patterns of feature reliance, providing a bridge between local prediction accuracy and global reasoning consistency.

    \item \textbf{Clear decision setting.} While many evaluation tasks adopt binary questions (e.g., yes/no in open-ended text or image-based object detection), such questions are typically \emph{constructed} for the benchmark \citep{li2023evaluating, arcuschin2025chain}. In contrast, tabular data naturally encode decision outcomes as binary or multi-class classification labels, which can be directly transformed into answers without additional design. This framing avoids the biases inherent in multiple-choice formats, where prior work shows that models can flip answers in up to 36\% of cases \citep{arcuschin2025chain}. While our present focus is on classification, extending the protocol to regression tasks represents a natural direction for future work.

    \item \textbf{Explainable end-to-end evaluation.} Because both features and labels carry explicit, domain-grounded meanings (e.g., age, income, or medical indicators), they are directly interpretable to humans without requiring additional segmentation or concept mapping. Tabular simulations capture the entire reasoning pipeline end-to-end, from structured inputs to predicted outputs, without relying on access to internal parameters or querying intermediate reasoning steps, thereby avoiding extraneous sources of bias. This design ensures that comprehension, grounding, application, and output fidelity are jointly assessed within a unified and realistic decision setting.

    \item \textbf{Generality with systematic probing.} Tabular data appear across nearly every real-world domain, from healthcare and finance to science and policy, making them a natural substrate for evaluating whether LLMs can act like domain experts. Their structured format also enables systematic perturbations, such as varying the number of rows, masking attributes, or ablating features, which provide direct tests of whether predictions truly depend on the claimed decision factors. Together, this breadth and manipulability establish tabular simulations as both widely applicable and quantitatively rigorous for assessing attributional faithfulness.
\end{enumerate}
In this way, tabular decision simulations provide a distinct, structured, and reproducible environment where understanding ability can be quantitatively assessed against explainable, domain-relevant ground truth.

\subsection{STaDS Protocol Tasks}
In this section, we operationalize the process of understanding and the relationship between LLMs and structured decision tasks (the ``object of understanding'') through the dimensions introduced in Section \ref{subsec:what-is-understanding}. These dimensions are assessed through \textbf{observable behavioral indicators}, which are identified through violations of task specifications. These violations reveal different aspects of a model's understanding ability.

\paragraph{Behavioral Indicators for Understanding}
Violations of the output specification provide diagnostic signals for different aspects of understanding ability:
\begin{enumerate}
\item \textbf{Question and instruction comprehension violations}: producing the wrong number of predictions, misaligned outputs, or irrelevant text indicates that the model has misinterpreted task instructions or failed to identify the required outputs. These violations assess whether the model has correctly \textbf{comprehended the task} by interpreting and following instructions properly. Misunderstandings at this level suggest that the model has not fully grasped what is being asked.

\item \textbf{Knowledge-based prediction violations}: generating labels in the correct format but with consistently low accuracy, or producing invalid labels outside $\mathcal{Y}$, suggests that the model has failed to ground prior knowledge in the domain or apply it effectively to the task. These violations reflect whether the model has correctly \textbf{applied learned knowledge} to generate correct and consistent predictions. Failures here indicate the model's inability to ground its predictions in the relevant domain knowledge.

\item \textbf{Reliance on the right decision factors violations}: providing self-claimed feature rankings inconsistent with actual feature reliance reflects that models produces unfaithful rationales despite having task-relevant knowledge. These violations measure, \textbf{decision faithfulness}, whether the model is reasoning based on the correct, relevant factors. A model with low decision faithfulness may make accurate predictions but fail to justify them using the right reasoning or decision criteria.
\end{enumerate}

\paragraph{STaDS Evaluation Tasks}
STaDS evaluates understanding ability through the three dimensions of understanding defined above, each assessed through distinct tasks:
\begin{enumerate}
    \item \textbf{Comprehension Fidelity}: This task evaluates whether the model can correctly interpret task instructions and adhere to output specifications. It probes whether the model understands the task by checking whether it produces the correct number of predictions, follows the specified output format, and provides relevant responses. Misalignments or format violations here assess the model's \textbf{question and instruction comprehension ability}.
    \item \textbf{Predictive Competence.} This task measures whether the model can generate accurate predictions based on its learned knowledge. It involves producing predictions for masked rows under zero-shot and few-shot settings, testing whether the model can effectively apply domain knowledge and generalize across different instances. Violations of this task (e.g., generating invalid or incorrect labels) assess the model's ability to apply relevant knowledge to make \textbf{correct predictions}.
    \item \textbf{Decision faithfulness.} 
    After generating predictions, the model is asked to provide a feature-importance ranking (self-attribution). This ranking is compared with behavioral attributions obtained through systematic perturbations, such as Leave-Any-Out (LAO) analysis. This task probes whether the model's stated rationale aligns with its actual feature reliance when making predictions, testing its \textbf{decision faithfulness}. Misalignments indicate that the model's explanation does not reflect the actual decision factors that influenced its predictions.
\end{enumerate}

\section{STaDS Metrics: Quantifying Comprehension, Competence, and Faithfulness}
To rigorously evaluate understanding, STaDS introduces a suite of metrics that quantify all dimensions discussed above. We first formalize the input and output.
\paragraph{Input formulation.}  
A tabular decision task is represented as $\mathcal{D} = \{(x_i, y_i)\}_{i=1}^N$, where each $x_i \in \mathbb{R}^d$ is a feature vector and $y_i \in \mathcal{Y}$ is the corresponding label. To evaluate the model, $\mathcal{D}$ is rendered into a structured prompt $\mathcal{C} = (I, T, S_k)$ at inference time, consisting of:
\begin{itemize}
    \item $I$: a natural language instruction specifying the professional role, task, and an attribute glossary mapping features to domain concepts;  
    \item $T$: a textual rendering of the structured table, where target labels are masked as \texttt{class=?};  
    \item $S_k = \{(x^{(j)}, y^{(j)})\}_{j=1}^k$: an optional set of $k$ in-prompt demonstrations.
\end{itemize}

\paragraph{Output specification.}  
The LLM $f_\theta$ is required to output predictions for the masked rows.
\[
\hat{\mathbf{y}} \;=\; \bigl(\hat{y}_1,\dots,\hat{y}_{n_{\text{p}}}\bigr)
      \;=\; f_{\theta}(C),
\]
where $n_p$ is the number of predictions generated.
Outputs are expected to follow strict formatting rules (e.g., exact number of predictions, valid label set, no additional text). These constraints ensure that performance reflects genuine task comprehension and not prompt-formatting artifacts.

\paragraph{Other Notations.}  
We evaluate model predictions against the ground-truth labels available for each task.  
Let $n_p$ denote the number of predictions produced by the model and $n_g$ the number of ground-truth labels available for evaluation.  
To  enable fair comparison\footnote{LLMs may produce incorrect number of predictions even if specified.}, we define the \emph{aligned evaluation length} as:
\[
n_a = \min(n_p, n_g),
\]
so that only the first $n_a$ prediction–label pairs $(\hat{y}_i, y_i)$ are considered.  
We define the following sets:  
\begin{itemize}
    \item \textbf{Valid label set ($\mathcal{Y}_{\text{valid}}$)}: the complete set of permissible label values specified by the task (e.g., ${0,1}$ for binary classification, or ${0,1,2}$ for 3-class). Any $\hat{y}_i \notin \mathcal{Y}_{\text{valid}}$ is treated as an invalid prediction.  
    \item \textbf{Ground-truth label set ($\mathcal{Y}_{\text{gt}}$)}: the set of unique ground-truth labels among the aligned pairs $n_a$. $$\mathcal{Y}_{\text{gt}} = \{y_i : 1 \leq i \leq n_a\}$$ 
    \item \textbf{Predicted label set($\hat{\mathcal{Y}}$)}: : the set of valid predicted labels among the aligned pairs $n_a$. 
    $$\hat{\mathcal{Y}} = \{\hat{y}_i : \hat{y}_i \in \mathcal{Y}_{\text{valid}}, 1 \leq i \leq n_a\}$$
\end{itemize}

\paragraph{Zero/Few Shot Settings.}
STaDS explicitly considers performance in both zero-shot and few-shot settings, since the gap between them reflects the extent to which models rely on intrinsic knowledge versus in-context adaptation. 
\begin{itemize}
  \item \textbf{Zero-shot ($k=0$):} The model receives prompt $\mathcal{C}$ with no demonstrations and must predict all rows whose labels are masked as \texttt{class=?}. This setting tests whether the model can ground its predictions directly in the instructions and table structure, reflecting its intrinsic knowledge grounding \cite{chen2023self}. In other words, does the model already know enough about the decision domain to perform competently without examples?

  \item \textbf{Few-shot ($k>0$):} The model receives $\mathcal{C}$ with $k$ labelled demonstrations injected into the prompt. This probes whether the model can align row-level features with labels when given exemplars, i.e., whether it can perform in-context learning in a manner analogous to how humans adapt to case-based examples \cite{petroni2019language}.
\end{itemize}

\paragraph{General Metrics.}
We adopt conventional classification metrics Accuracy (Acc), Macro‑F1, and Label‑Set Jaccard (Set‑Jacc) for reference. These capture baseline task performance and provide a point of comparison to existing tabular classification benchmarks. 

For each $c\in\mathcal{Y}_{\text{gt}}$, compute precision, recall, and $F1_c$ on the aligned set. 
\small{
\begin{align*}
  & \text{Acc} \;=\; \frac{1}{n_a}\sum_{i=1}^{n_a}1[\hat y_i = y_i], \quad
   \text{Macro-F1} \;=\; \frac{1}{|\mathcal{Y}_{\text{gt}}|}\sum_{c\in\mathcal{Y}_{\text{gt}}} F1_c.
\end{align*}
}

\noindent The label-set Jaccard is given by:
\small{
\begin{equation*}
  \text{Set‑Jacc} \;=\; \frac{|\hat{\mathcal{Y}}\cap \mathcal{Y}_{\text{gt}}|}{|\hat{\mathcal{Y}}\cup \mathcal{Y}_{\text{gt}}|}.
\end{equation*}}

\paragraph{Comprehension Fidelity Metrics.}
\label{sec:cf-metrics}
We explicitly penalize \textit{over/under-production} as well as \textit{unknown labels}, as these indicate a failure to understand and follow task instructions.

\begin{itemize}
    \item \textbf{Length F1 (Len‑F1)}: Len‑F1 measures output‑length fidelity, where the model produces incorrect number of expected predictions.
    Let $P_L={n_a}/{n_p}$\;(set $0$ if $n_p\!=\!0$) denote precision with respect to output length, and $R_L={n_a}/{n_g}$\;(set $0$ if $n_g\!=\!0$) denote recall with respect to the number of ground-truth labels. Then,
    \small{\begin{equation}
      \text{Len-F1} \;=\; 
      \begin{cases}
         \displaystyle \frac{2P_L R_L}{P_L + R_L}, & \text{if } P_L+R_L>0,\\[6pt]
         0, & \text{otherwise.}
      \end{cases}
    \end{equation}}
    \item \textbf{Unknown‑Label Rate (UnkLbl\%)}
    UnkLbl\% refers to the fraction of all produced predictions outside the valid label set:
    \small{\begin{equation}
      \text{UnkLbl\%} \;=\; \frac{1}{n_p}\sum_{i=1}^{n_p}\mathbf{1}[\hat y_i\notin \mathcal{Y}_{\text{valid}}]
      \quad (0 \text{ if } n_p=0).
    \end{equation}}    
\end{itemize}

\paragraph{Predictive Competence Metrics.}
\label{sec:pc-metrics}
We integrates correctness, output-length fidelity, and label validity into a single measure of predictive quality as \textbf{Penalized Accuracy (PenAcc)}:
\begin{equation}
      \text{PenAcc} \;=\; \text{Acc} - (\alpha \times (1-\text{Len-F1})+\beta \times\text{UnkLbl\%}),
\end{equation}
where $\alpha, \beta > 0$ are penalty weights.

\noindent \textbf{Remark.} This metric suite allows us to rigorously assess (i) correct output length; (ii) valid label; and (iii) accurate predictions. PenAcc folds these requirements into a single score. An ideal model achieves $\text{Acc}=\text{PenAcc}$; any non-zero penalty ($\Delta_{\text{acc}} = \text{Acc}-\text{PenAcc} > 0$) indicates a format violation. Consequently, PenAcc serves as a concise, end-to-end indicator of whether an LLM both \emph{understands} the prompt specification and \emph{solves} the prediction task.

\paragraph{Decision Faithfulness Metrics.}
Decision faithfulness is evaluated by comparing a model’s \emph{stated reasons} for its predictions with its \emph{actual feature reliance}. Following principles from XAI, we employ two complementary attribution methods:

\begin{itemize}
  \item \textbf{Self-claimed Attribution (Interpretability).} Given the same context, the model is prompted to produce a ranking $\pi_{\text{self}}$ over the $m$ features, indicating which attributes it \emph{believes} were most influential for its decision. This refers to the model's own understanding of the features that influenced its decision.

  \item \textbf{Leave-Any-Out (LAO) Attribution (Explainability).} For each feature $j \in \{1,\ldots,m\}$, we re-evaluate the model under identical prompting while \emph{ablating} that feature from all rows~\citep{koh2017understanding}:
    \[
      \Delta_j \;=\; \text{Perf}(\mathcal{D})
                     \;-\;
                     \text{Perf}\!\bigl(\mathcal{D} \setminus x_{[:,j]}\bigr),
    \]
    where $\text{Perf}$ is the same predictive metric used for predictive competence (e.g., Accuracy or F1). This yields an attribution vector $\boldsymbol{\Delta} = (\Delta_1,\ldots,\Delta_m)$ and an induced ranking $\pi_{\text{LAO}}$, where larger $\Delta_j$ indicates a stronger reliance on feature $j$\footnote{Ablation can target individual features or pre-defined feature \emph{groups} to capture higher-order interactions.}. This provides a post-hoc justification for the model's decision based on the impact of removing specific features.
    
\end{itemize}

\noindent
Comparing $\pi_{\text{self}}$ and $\pi_{\text{LAO}}$ helps assess the extent to which the model's stated explanations (self-claimed attribution) faithfully reflect its actual decision-making reliance (behavioral attribution). We then formalize the following metrics for global faithfulness.

\begin{itemize}
    \item \textbf{Self-Attribution Recall (\textsc{SelfAtt@\,$k$})}: This metric measures how well the model's self-reported important features cover the ground-truth feature set. Specifically, given the ground-truth feature set $\mathcal{S}_m$ and $\mathrm{Top}_k(\pi_{\text{self}})$ as the first $k$ distinct features in
    $\pi_{\text{self}}$, then \textsc{SelfAtt@\,$k$} is defined as:
    \small{\[
      \textsc{SelfAtt@}k
      \;=\;
      \frac{|\mathcal{S}_m \cap \mathrm{Top}_k(\pi_{\text{self}})|}
           {|\mathcal{S}_m|},
      \quad
      k = |\mathcal{S}_m|\text{ by default}.
    \]}

    \item \textbf{Self-Decision Faithfulness (\textsc{Self-Faith}, $\rho$)}: This metric measures the agreement between the model's self-claimed attribution ranking and its actual reliance on features. Specifically, we calculate Spearman’s rank correlation coefficient ($\rho$) between the behavioral ranking derived from LAO scores ($\pi_{\text{LAO}}$) and the self-claimed ranking ($\pi_{\text{self}}$). A higher agreement between these rankings indicates how faithfully the model reports its own decision rule.
    \small{
    \begin{align*}
		\text{Spearman’s } \rho 
		&\;=\;
		1 - \frac{6}{m(m^{2}-1)}
		\sum_{i=1}^{m} (r_i - s_i)^{2},
	\end{align*}
    }where $r_i$ and $s_i$ denote the respective ranks of feature $i$ in $\pi_{\text{self}}$ and $\pi_{\text{LAO}}$.
    $p$-values for $\rho$ (and $\tau$) can be obtained through permutation tests.

    \item \textbf{LAO Magnitude ($\sigma_{\text{LAO}}$)}: This metric captures the dispersion of the model's behavioral reliance across features, reflecting the concentration and interpretability of its decision rationale. It is computed as the standard deviation of the LAO performance changes across all features:
    \small{\[
      \sigma_{\text{LAO}}
      \;=\;
      \sqrt{\frac{1}{m-1}\sum_{j=1}^{m}(\Delta_j-\bar{\Delta})^{2}},
      \quad
      \bar{\Delta}=\frac{1}{m}\sum_{j=1}^{m}\Delta_j,
    \]}where $\boldsymbol{\Delta}=(\Delta_1,\dots,\Delta_m)$. A small $\sigma_{\text{LAO}}$ indicates that feature effects are evenly distributed, suggesting reliance on many weak signals. A large $\sigma_{\text{LAO}}$ indicates \textit{sparse and concentrated reliance}, often regarded as more interpretable and human-comprehensible in XAI literature.
\end{itemize}

% \noindent \textbf{Interpretation Protocol.} This metric suite allows us to rigorously assess \textit{attributional faithfulness} along three key axes:
% \begin{enumerate}
%     \item \textit{Self-attribution recall:} does the model's explanation match the correct number of features?
%     \item \textit{Behavioural sparsity:} are LAO effects sufficiently concentrated for human inspection?
%     \item \textit{Self-faithfulness:} do stated reasons align with causal reliance? 
% \end{enumerate}

\subsection{Integrated View}
Taken together, these three key dimensions, \textbf{Comprehension Fidelity}, \textbf{Predictive Competence} and \textbf{Decision Faithfulness}, allow us to map the model's behavior into distinct regions of understanding. These dimensions jointly reveal how well the model performs the task and whether it does so for the right understanding.
\begin{itemize}
\item \textbf{Accurate \& Faithful:} Reliable expert-like decision-maker, producing correct predictions grounded in faithful decision factors.
\item \textbf{Accurate \& Unfaithful:} Correct predictions are made, but the model's stated rationale is misleading, suggesting that while the predictions are accurate, the global rule behind them is flawed from the true decision factors.
\item \textbf{Inaccurate \& Faithful:} Grounded but honest decision-maker, whose reasoning aligns with true decision factors, but fails to generalize or make correct predictions.
\item \textbf{Inaccurate \& Unfaithful:} A failure of both competence and faithfulness, where the model neither understands the task nor applies the correct global rule.
\end{itemize}

\begin{figure}
    \centering
    \includegraphics[width=\linewidth]{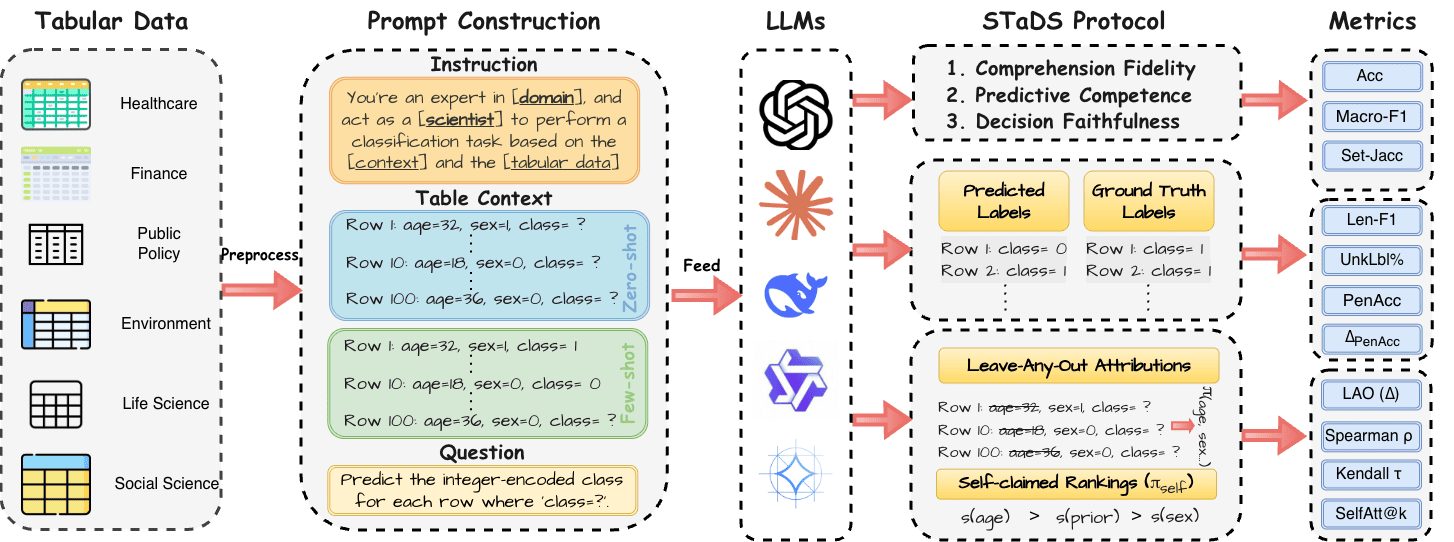}
    \caption{The diagram illustrates the process flow from tabular data preprocessing across various domains to prompt construction for LLMs. 
    The model is tasked with predicting labels based on structured input rows and is evaluated along three key dimensions of understanding. \textit{Question and Instruction Comprehension} is measured by Len-F1, UnkLbl\%, and $\Delta_{\text{PenAcc}}$, \textit{Predictive Competence} is quantified through Accuracy, Macro-F1, and PenAcc, and \textit{Reliance on the Right Decision Factors} is measured by SelfAtt@k, Self-Decision Faithfulness and LAO Magnitude.}
    \label{fig:main-fig}
\end{figure}

\section{STaDS Benchmark \& Experimental Setup}\label{sec:exp-setup}
We next describe the datasets, models, prompting strategy (e.g., table formatting, instruction, and question), LAO-attribution protocol, and implementation details used to instantiate the evaluation. The overview is illustrated in Fig. \ref{fig:main-fig}.

\noindent \textbf{Benchmark datasets.}
We employ 15 tabular datasets spanning diverse domains (see professional roles in Table~\ref{tab:data}), encompassing binary and multi-class classification settings as well as both balanced and imbalanced scenarios. Sample sizes range from $150$ to $100{,}000$ (tokens up to 128k), and the number of features varies from 4 to 25. For the few-shot setting, each dataset is stratified into 80\% \emph{train} (used for in-prompt demonstrations) and 20\% \emph{test} (hidden rows for evaluation). 

\begin{table}[!t]
\centering
\small
\begin{adjustbox}{max width=\linewidth}
\begin{tabular}{lcccc}
\toprule
\textbf{Dataset} & \textbf{Sample ($N$)} & \textbf{Feature ($m$)} & \textbf{Task} & \textbf{Acting Role}\\
\midrule
Adult Income \citep{yeh2009comparisons}          & 32561 & 14 & Binary & labour–economist\\
Breast Cancer \citep{xin2022exploring}           &   277   &  9 & Binary & clinical oncologist\\
Car Evaluation \citep{asuncion2007uci}           &  1728 &  6 & Multi  & automotive specialist\\
COMPAS \citep{jordan2015effect}                 &  6172 & 25 & Binary & criminal risk-assessment analyst\\
Congressional Voting \citep{congressional_voting}                           &   232   & 16 & Binary & legislative political scientist\\
Gaussian Synthetic \citep{agarwal2022openxai}                              &  5000 & 20 & Binary & applied statistician\\
German Credit \citep{asuncion2007uci}            &  1000 & 20 & Binary & bank credit risk analyst\\
Give Me Some Credit \citep{freshcorn2022gmsc}                            &102209 & 10 & Binary & consumer credit risk analyst\\
Framingham Heart  \citep{who_cvd_2021}                              &  3658 & 15 & Binary & cardiovascular epidemiology analyst\\
HELOC (FICO)~\citep{holter2018fico}                                    &  9871 & 23 & Binary & home-equity lending risk analyst\\
Iris \citep{Unwin2021TheID}                                           &    150  &  4 & Multi  & botanical data analyst\\
MONK 1 / 2 / 3 \citep{thrun1991monk}                                 &    432  &  6 & Binary & data analyst\\
Pima Indians Diabetes \citep{smith1988using}     &    768  &  8 & Binary & diabetes researcher\\
\bottomrule
\end{tabular}
\end{adjustbox}
\caption{Summary of the tabular datasets, listing sample size ($N$), feature count ($m$), task type (binary/multi-class classification), and the corresponding professional roles adopted by the model in each prompt to ensure domain-relevant reasoning.}
\label{tab:data}
\end{table}

\noindent \textbf{LLMs \& Hardware.}
We evaluate $9$ LLMs. Open-source models: including Llama3-8B-Instruct (Llama3-8B) and Llama3-3B~\citep{llama3}, Mistral-7B-Instruct-v0.3 (Mistral-7B)~\citep{mistral}, DeepSeek-Llama-8B (DeepSeek-Llama-8B)~\citep{deepseek}, Qwen3-8B~\citep{qwen}, and Gemma-1B/4B-it (Gemma-1B/4B)~\citep{gemma} are run on 8$\times$NVIDIA RTX 3090 cluster, GH200, and 8$\times$A100 cluster. Commercial baselines Gemini-2.5-Pro and GPT-4.1-mini are accessed via their public APIs. All models are decoded with a unified configuration (temperature 0.2, top-p 1.0, new maximum tokens 8192).

\begin{table}[htb]
\centering
\small
\begin{adjustbox}{max width=\linewidth}
\begin{tabular}{l|c|cc|c|cc|c}
\toprule
Dataset & Tokens & \multicolumn{2}{c|}{Zero-shot} & Best Z & \multicolumn{2}{c|}{Few-shot} & Best F \\
& (100 rows) & Acc & Macro-F1 & Model & Acc & Macro-F1 & Model \\
\midrule
Adult Income & 10K & 0.700 & 0.688 & Gemini-2.5-Pro & 0.737 & 0.708 & Gemini-2.5-Pro \\
Breast Cancer & 17K & 0.729 & 0.524 & GPT-4.1-mini & 0.732 & 0.525 & GPT-4.1-mini \\
Car Evaluation & 16K & 0.419 & 0.243 & Gemini-2.5-Pro & 0.600 & 0.616 & Gemini-2.5-Pro \\
Compas & 39K & 0.816 & 0.810 & Gemini-2.5-Pro & 0.716 & 0.714 & Gemini-2.5-Pro \\
Congression Votes & 36K & 0.534 & 0.348 & Llama3-3B & 0.638 & 0.636 & Gemini-2.5-Pro \\
Synthetic & 48K & 0.550 & 0.448 & Gemini-2.5-Pro & 0.880 & 0.873 & GPT-4.1-mini \\
German Credit & 13K & 0.660 & 0.616 & GPT-4.1-mini & 0.889 & 0.862 & Gemini-2.5-Pro \\
Give Me Some Credit & 14K & 0.830 & 0.832 & DeepSeek-Llama-8B & 0.917 & 0.916 & GPT-4.1-mini \\
Heart Disease & 12K & 0.640 & 0.614 & GPT-4.1-mini & 0.700 & 0.697 & GPT-4.1-mini \\
HELOC & 25K & 0.670 & 0.670 & Gemini-2.5-Pro & 0.885 & 0.883 & Gemma-4B \\
Iris & 7K & 0.787 & 0.787 & Gemini-2.5-Pro & 1.000 & 1.000 & Gemini-2.5-Pro \\
Monk\_1 & 18K & 0.620 & 0.613 & Gemini-2.5-Pro & 0.759 & 0.758 & Qwen3-8B  \\
Monk\_2 & 18K & 0.674 & 0.409 & DeepSeek-Llama-8B & 0.713 & 0.601 & Qwen3-8B  \\
Monk\_3 & 18K & 0.579 & 0.596 & Gemini-2.5-Pro & 0.644 & 0.642 & Gemini-2.5-Pro \\
Pima & 31K & 0.758 & 0.784 & Gemini-2.5-Pro & 0.820 & 0.814 & Gemini-2.5-Pro \\
\bottomrule
\end{tabular}
\end{adjustbox}
\caption{Prediction results on tabular benchmarks. We report the best accuracy and macro-F1 across models for zero-shot and few-shot settings, along with the model achieving the best score and the approximate number of tokens (for 100 examples).}
\label{tab:icu-main}
\end{table}

\subsection{Prompt Construction}
\label{sec:prompts}
Inspired by the structured prompting strategy of \citet{zhang2023tablellama},  
we design a \textbf{single, deterministic template} composed of four consecutive blocks for STaDS, as shown below:
\begin{center}
\begin{minipage}{0.9\linewidth}
\ttfamily
Below is an instruction that describes a task, paired with an input table that provides further context. \\
Write a response that appropriately completes the request. \\
⟨Instruction⟩ \\
⟨Input⟩ \\
⟨Question⟩ \\
⟨Response⟩
\end{minipage}
\end{center}

\subsubsection{Instruction Template.}
Each prompt begins with a concise, self-contained instruction specifying \emph{who} the model should act as and \emph{what} task it is to perform. Specifically, we include the following fields:
\texttt{⟨DATASET⟩}, \texttt{⟨ROLE⟩}, \texttt{⟨TASK TYPE⟩}, \texttt{⟨TARGET ENCODING⟩}, and an \texttt{⟨ATTRIBUTE GLOSSARY⟩}. Below are descriptions:
\begin{description}[leftmargin=3em]
    \item[\texttt{⟨DATASET⟩}] The name of the benchmark dataset.  
    \emph{Example:} \texttt{Breast Cancer}.
    
    \item[\texttt{⟨ROLE⟩}] The professional identity that the model is instructed to assume, chosen to reflect domain expertise.  
    \emph{Example:} \texttt{clinical oncologist}.
    
    \item[\texttt{⟨TASK TYPE⟩}] The prediction setting (e.g., binary or multi-class classification) together with its domain-specific description.  
    \emph{Example:} \texttt{binary classification — predicting whether a breast cancer patient will experience recurrence or not}.
    
    \item[\texttt{⟨TARGET ENCODING⟩}] The mapping between integer-coded labels and their semantic meanings. This ensures the model outputs strictly integer predictions while preserving human interpretability.  
    \emph{Example:} \texttt{\{0: no-recurrence-events, 1: recurrence-events\}}.
    
    \item[\texttt{⟨ATTRIBUTE GLOSSARY⟩}] A glossary listing each input feature, its semantic description, and categorical encodings (if applicable). This grounds the tabular features in explicit domain knowledge.  
    \emph{Example (Breast Cancer dataset, partial):}  
    \begin{itemize}[leftmargin=1.5em]
        \item \texttt{age}: Age group of the patient \{0: 10--19, 1: 20--29, \ldots, 8: 90--99\}
        \item \texttt{menopause}: Menopausal status \{0: lt40, 1: ge40, 2: premeno\}
        \item \texttt{tumor-size}: Tumor size intervals in mm \{0: 0--4, 1: 5--9, \ldots, 11: 55--59\}
        \item \texttt{node-caps}: Capsular invasion \{0: no, 1: yes\}
        \item \texttt{irradiat}: Radiation therapy received \{0: no, 1: yes\}
    \end{itemize}
\end{description}

The instruction typically opens with:
\begin{center}
\begin{minipage}{0.9\linewidth}
\ttfamily
Act as a professional ⟨ROLE⟩, \\
Your task is to perform ⟨TASK TYPE⟩, predicting whether ... or not. \\
⟨TARGET ENCODING⟩ ⟨ATTRIBUTE GLOSSARY⟩ \\
For every row where ``class=?'', predict its integer target, relying solely on your pre-trained knowledge. \\
Return one integer per row, in the exact same order as the rows appear. \\
The number of predictions must equal the number of rows with ``class=?''. 
\end{minipage}
\end{center}

\subsubsection{Tabular Input.}
The dataset is rendered as plain text, with one row per line:
\small{\[
\texttt{Row Num: attribute\_1 = 2, attribute\_2 = 0, \\ ..., class = ?}
\] }
All categorical variables are pre-encoded as integers, and rows requiring prediction are marked with \texttt{class=?}.  
For \textit{LAO experiments}, the specified feature(s) are physically removed from each row, leaving all other attributes unchanged.

\subsubsection{Question.}
The task is restated in a single sentence, explicitly specifying \emph{the exact number $N$} of unknown rows for clarity:  
\begin{quote}
\small
\textit{Predict the integer-encoded class for the $N$ rows where
\texttt{class=?}. Output \emph{exactly} $N$ predictions in the same order.}
\end{quote}
The length requirement links directly to the \textsc{Len-F1} metric.

\subsubsection{Output Format.}
The model is instructed to return a list of integer labels. For instance,
\small{\[
\texttt{[\,0, 2, 1, …, 3\,]}
\]}
i.e., a Python-style list of $N$ comma-separated integers and \emph{no additional
text}.  Any deviation triggers the label penalties in Sec.~\ref{sec:pc-metrics}.

\begin{remark}
\label{rem:formatting}
Despite explicit instructions, some LLMs often generate additional explanations or non-standard formatting. To ensure fair evaluation, we post-process all raw outputs using \textsc{GPT-4-mini} to obtain clean, standardized predictions.
\end{remark}

\paragraph{Self-attribution protocol.}
To elicit the self-attribution ranking ($\pi_{\text{self}}$), we provide the full table and ask the model to order all input features by their importance for predicting the target variable. 

% Because LLMs are often sensitive to surface form, we designed \textbf{five} semantically equivalent but stylistically distinct prompts, including academic, instructional, analytic, evaluative, and concise styles (see Table \ref{tab:selfattr_prompts}).

Each prompt explicitly enforces a strict output format: a single comma-separated line listing all valid feature names in descending order of importance, with the target label \texttt{class} excluded. 
This reduces ambiguity and prevents additional text such as numbering, bullets, or explanations. LLMs might be prompted as follows to rank all the features:

\begin{quote}
\small
\textit{Rank all the features in order of their importance for predicting the target variable, from most important to least...}
\end{quote}
We expect the model to return a list of feature names. For instance,
\[
\texttt{[\,attribute\_3, attribute\_1, attribute\_2, …, attribute\_7\,]}
\]

\begin{table}[t]
\small
\centering
\begin{tabular}{p{3cm}ccc}
\toprule
Model & $\sigma_{\text{LAO}}$ & \textsc{Self-Faith} & \textsc{SelfAtt@\,$k$} \\
\midrule
Gemma-1B & 0.17 & NaN & 0.00 \\
Gemini-2.5-Pro & 0.07 & 0.25 (0.38) & 1.00 \\
DeepSeek-Llama-8B & 0.24 & 0.24 (0.41) & 1.00 \\
Llama3-8B & 0.00 & $-0.05$ (0.89) & 0.73 \\
Qwen3-8B  & 0.11 & $-0.17$ (0.67) & 0.44 \\
GPT-4.1-mini & 0.01 & $-0.02$ (0.96) & 1.00 \\
Llama3-3B & 0.11 & $-0.34$ (0.24) & 1.00 \\
Mistral-7B & 0.01 & $-0.54$ (0.08) & 0.73 \\
\bottomrule
\end{tabular}
\caption{\label{tab:au_metrics} Decision faithfulness  metrics between self‑attribution rank ($\pi_{\text{self}}$) and LAO-attribution rank ($\pi_{\text{LAO}}$); Adult Income Dataset ($m=14$). NaN indicates $\pi_{\text{self}}$ empty.}
\end{table}

\begin{figure*}[!t]
    \centering
    \includegraphics[width=\linewidth]{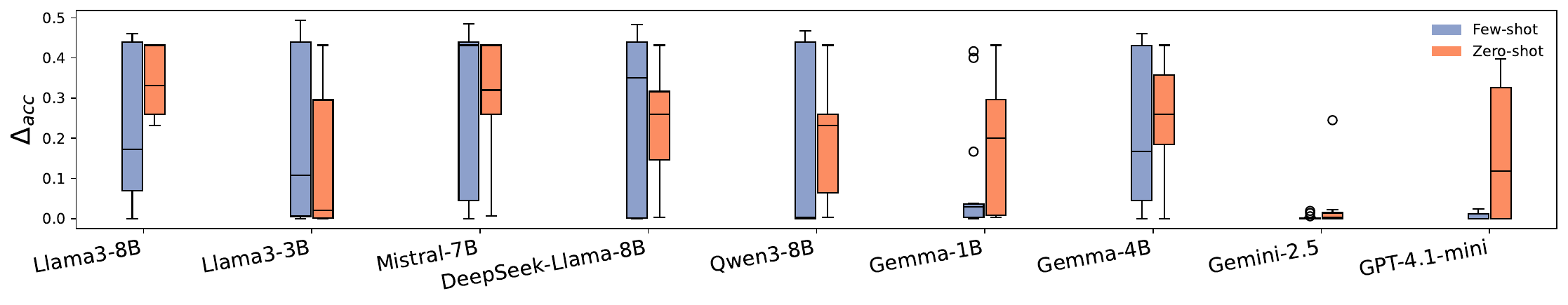}
    \caption{Box plots illustrate the distribution of $\Delta_\text{acc} = \mathrm{Acc} - \mathrm{PenAcc}$ for each model across all benchmark datasets. Blue and orange correspond to few-shot and zero-shot settings, respectively. Frontier models cluster near zero $\Delta_\text{acc}$, while several open-source checkpoints incur format penalties, especially in few-shot setting, indicating heightened prompt sensitivity.}
    \label{fig:acc_delta}
\end{figure*}

\begin{figure}[!ht]
\centering
    \begin{subfigure}[t]{.5\textwidth}
        \centering
        \includegraphics[width=\textwidth]{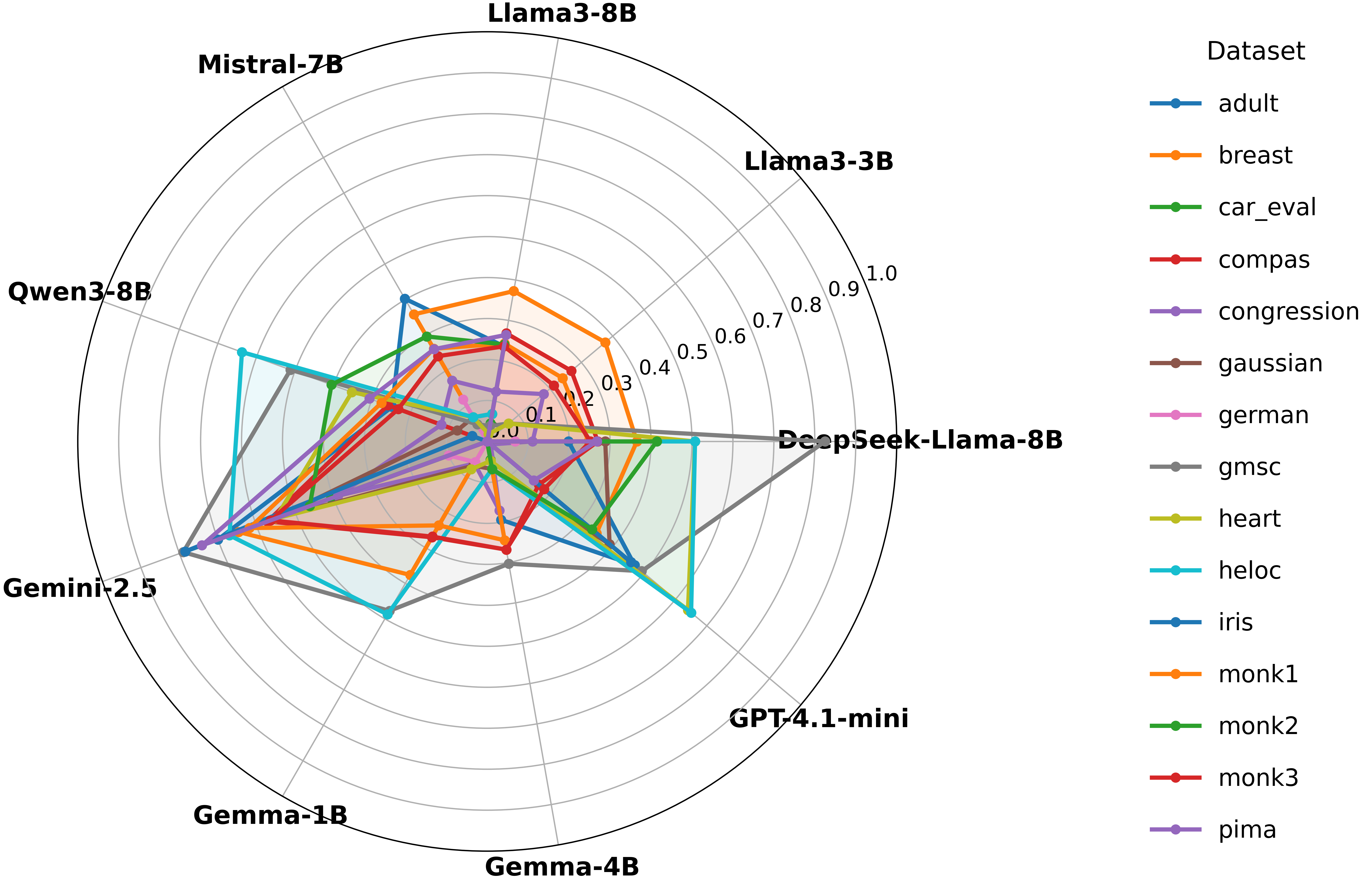}
        \caption{Zero-shot across benchmark datasets}
    \end{subfigure}%
    \hfill
    \begin{subfigure}[t]{.5\textwidth}
        \centering
        \includegraphics[width=\textwidth]{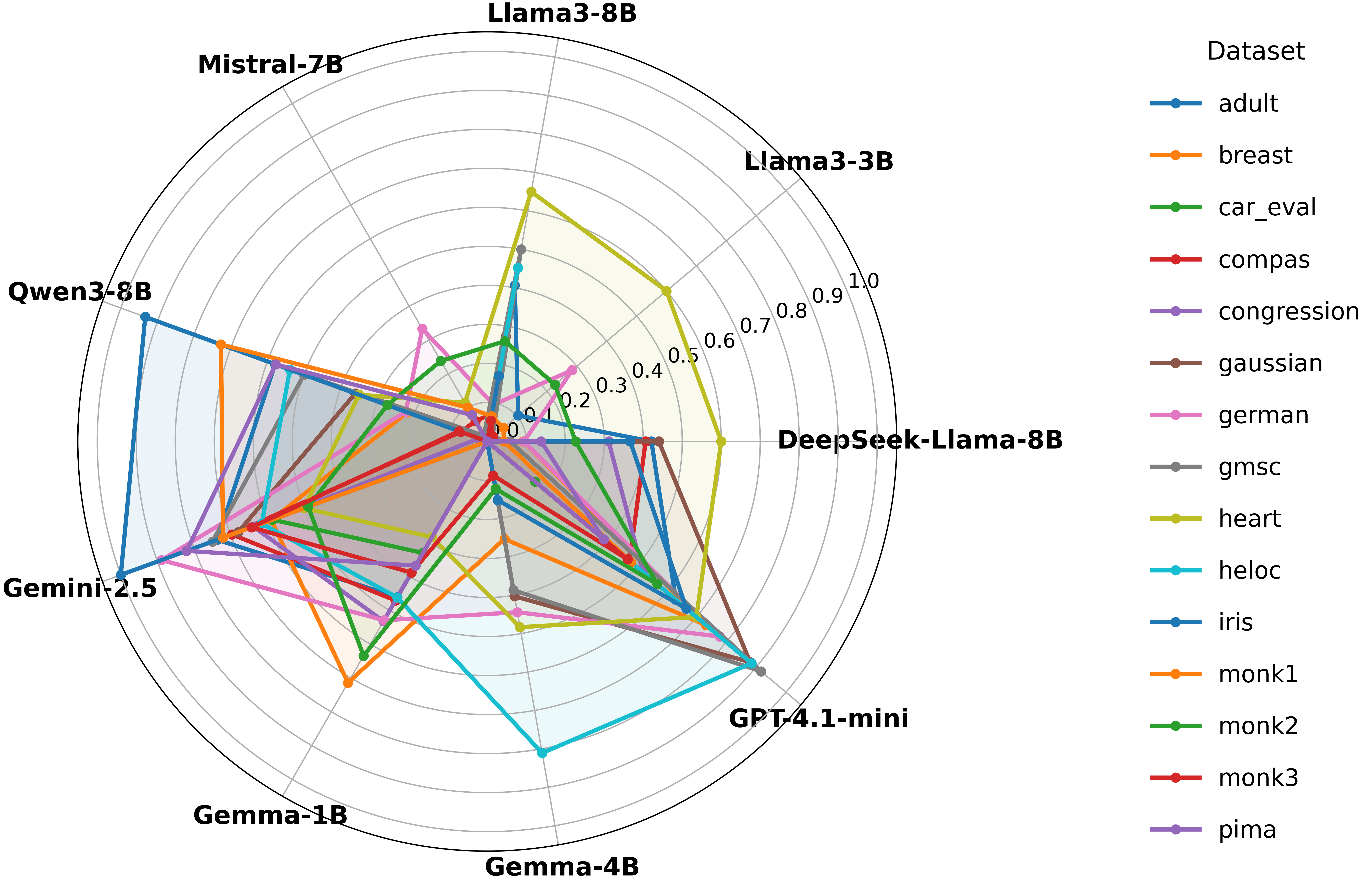}
        \caption{Few-shot across benchmark datasets}
    \end{subfigure}%
\caption{Spider plots of Penalized Accuracy ($\alpha = 0.5, \beta = 0.5$) across models and datasets in (a) zero-shot and (b) few-shot settings. Each axis is a model; each colored trace is a dataset. Higher values indicate stronger accuracy and instruction-following. Few-shot generally inflates the polygons (with higher PenAcc) across datasets, with Gemini-2.5-Pro showing the most uniform gains.}
\label{fig:spider}
\end{figure}

\begin{figure*}[!t]
\centering
    \begin{subfigure}[t]{.5\textwidth}
        \centering
        \includegraphics[width=\textwidth]{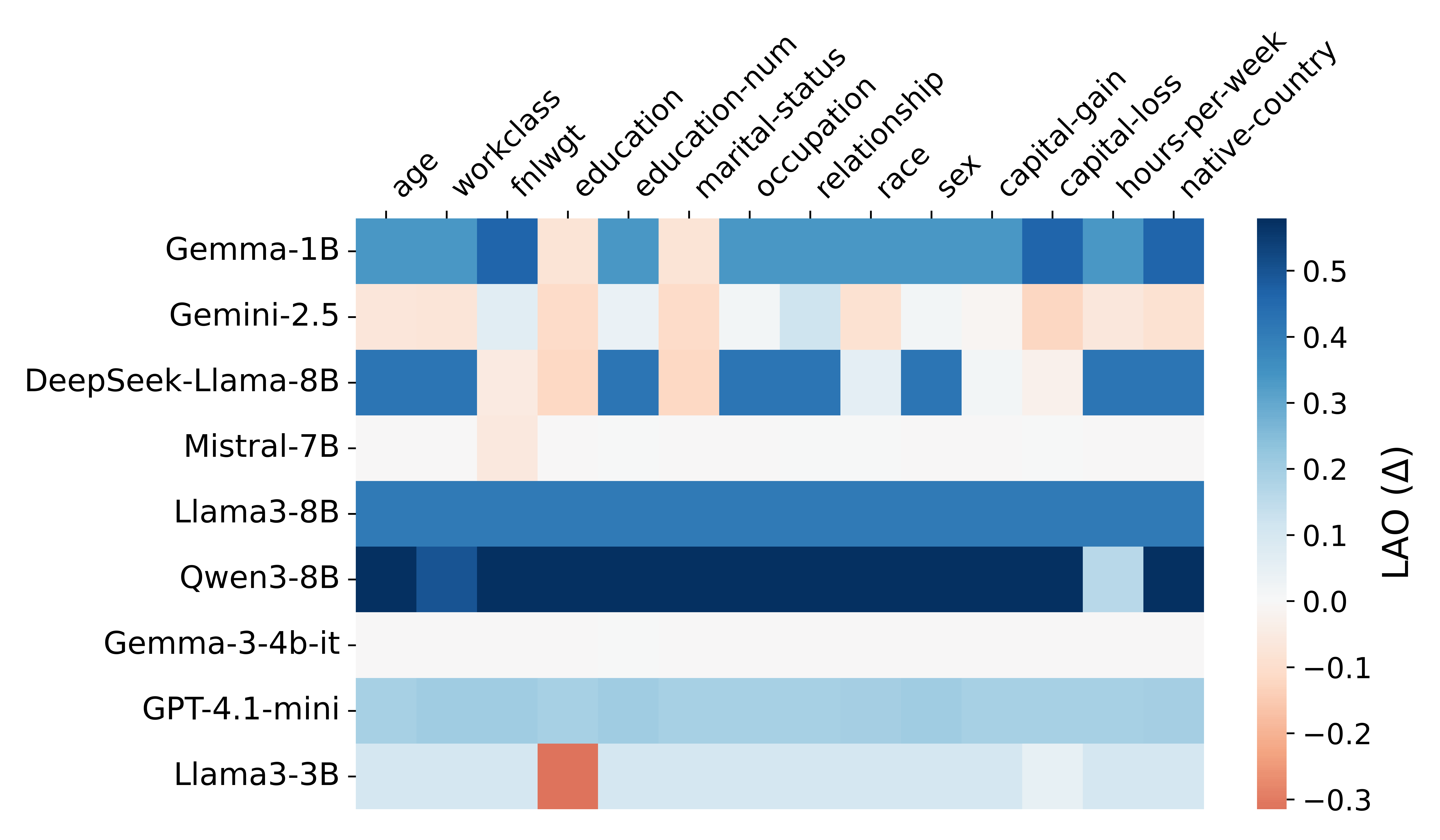}
        \caption{Adult Income}
    \end{subfigure}%
    \hfill
    \begin{subfigure}[t]{.5\textwidth}
        \centering
        \includegraphics[width=\textwidth]{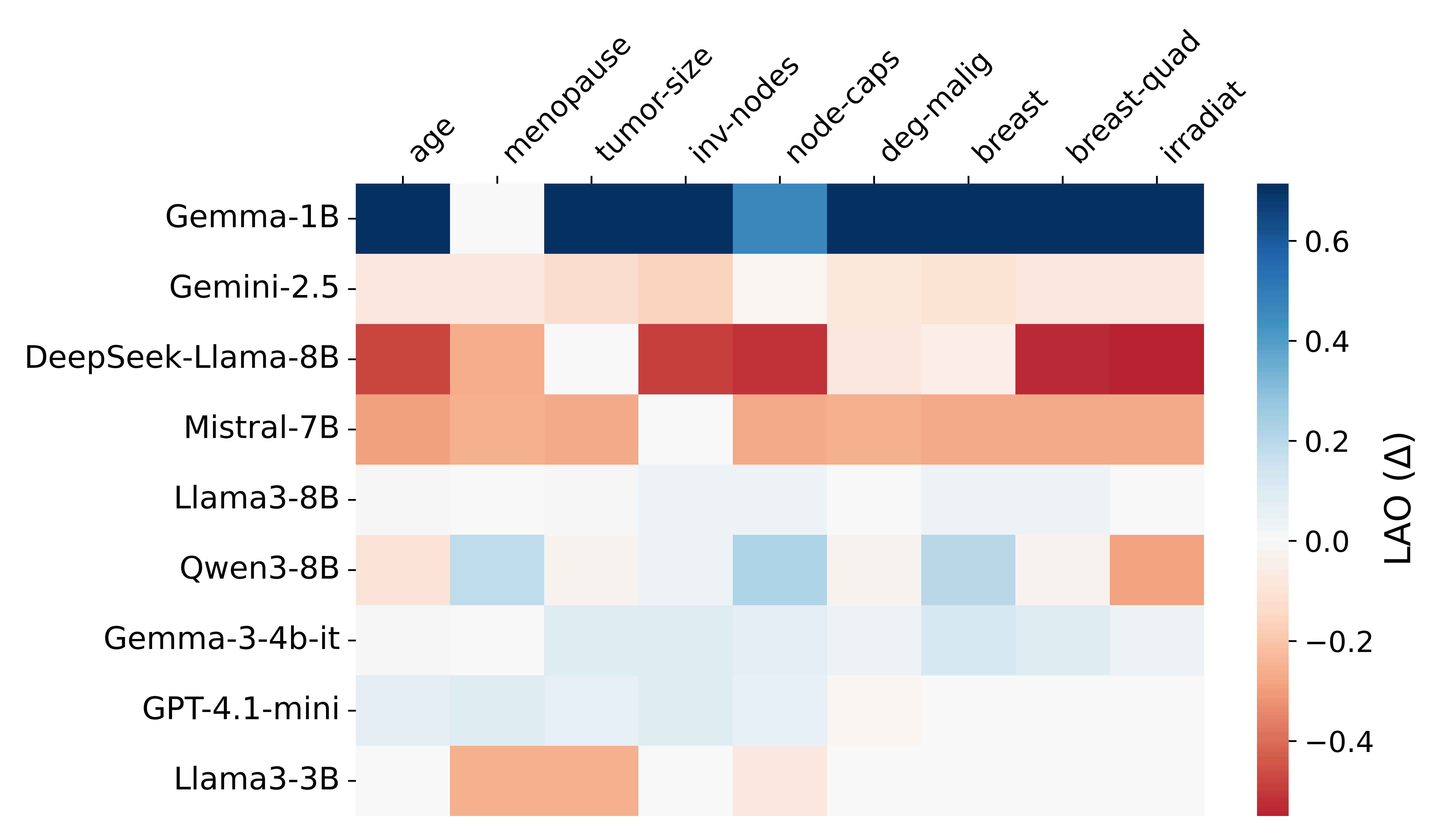}
        \caption{Breast Cancer}
    \end{subfigure}%
\caption{Heatmap of LAO performance ($\Delta_{\text{LAO}}$) for each feature (columns) and LLM (rows). Darker blue indicates a larger performance loss when the feature is removed (higher importance); red indicates a slight performance gain or negligible reliance. A few features dominate reliance for certain models (deep blue), while others spread reliance diffusely, consistent with their $\sigma_{\text{LAO}}$.}
\label{fig:lao_delta}
\end{figure*} 

\begin{figure*}[]
    \centering
    \includegraphics[width=\linewidth]{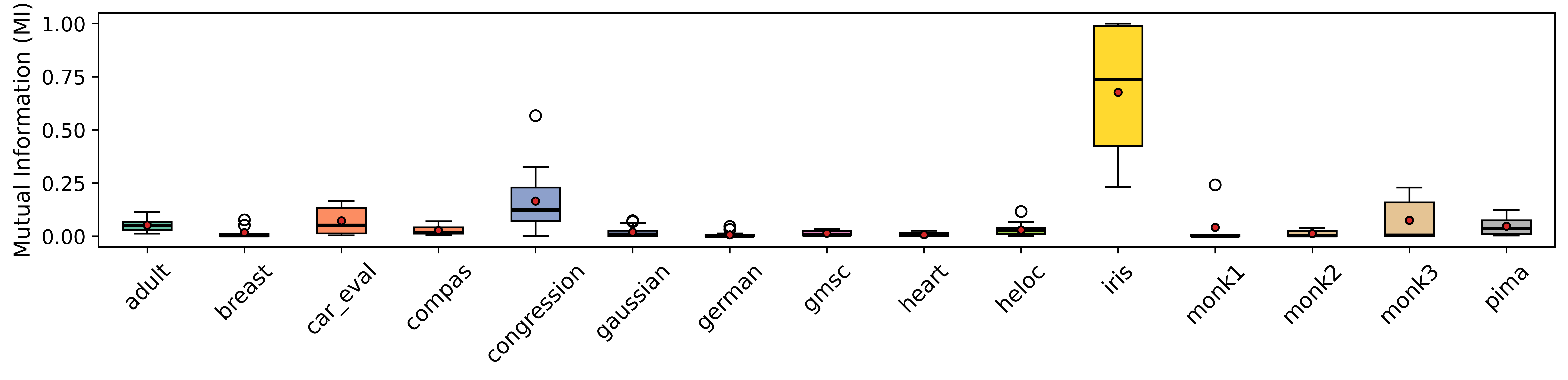}
    \caption{The boxplot summarizes the normalized distribution of Mutual NMI by dataset; Overall, NMI magnitudes are low, indicating that raw dataset co-occurrence cannot account for decision faithfulness.}
    \label{fig:mi_box}
\end{figure*}

\begin{figure*}[hp!]
\centering
    \begin{subfigure}[t]{.33\textwidth}
        \centering
        \includegraphics[width=\textwidth]{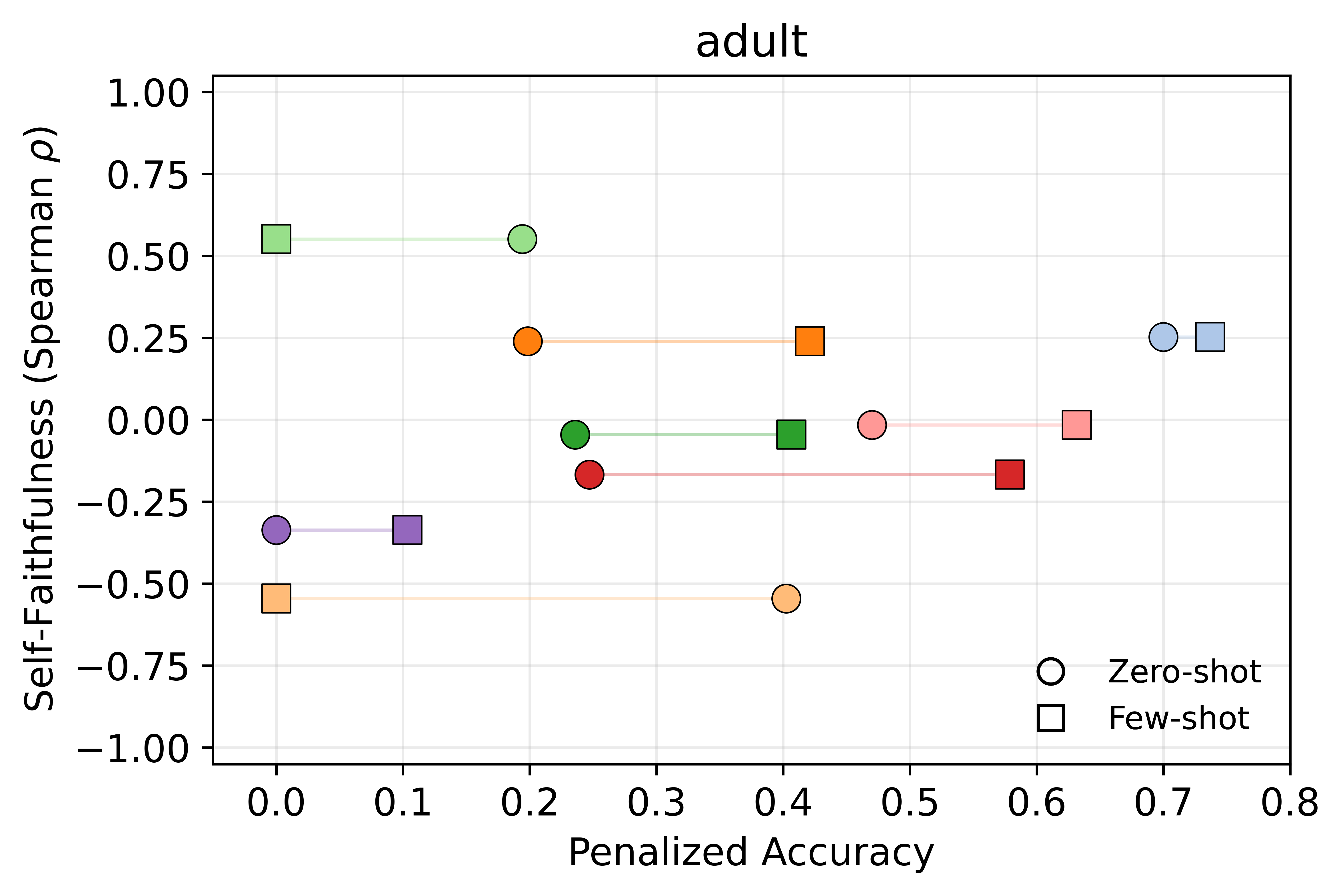}
        % \caption{Adult Income}
    \end{subfigure}%
    \hfill
    \begin{subfigure}[t]{.33\textwidth}
        \centering
        \includegraphics[width=\textwidth]{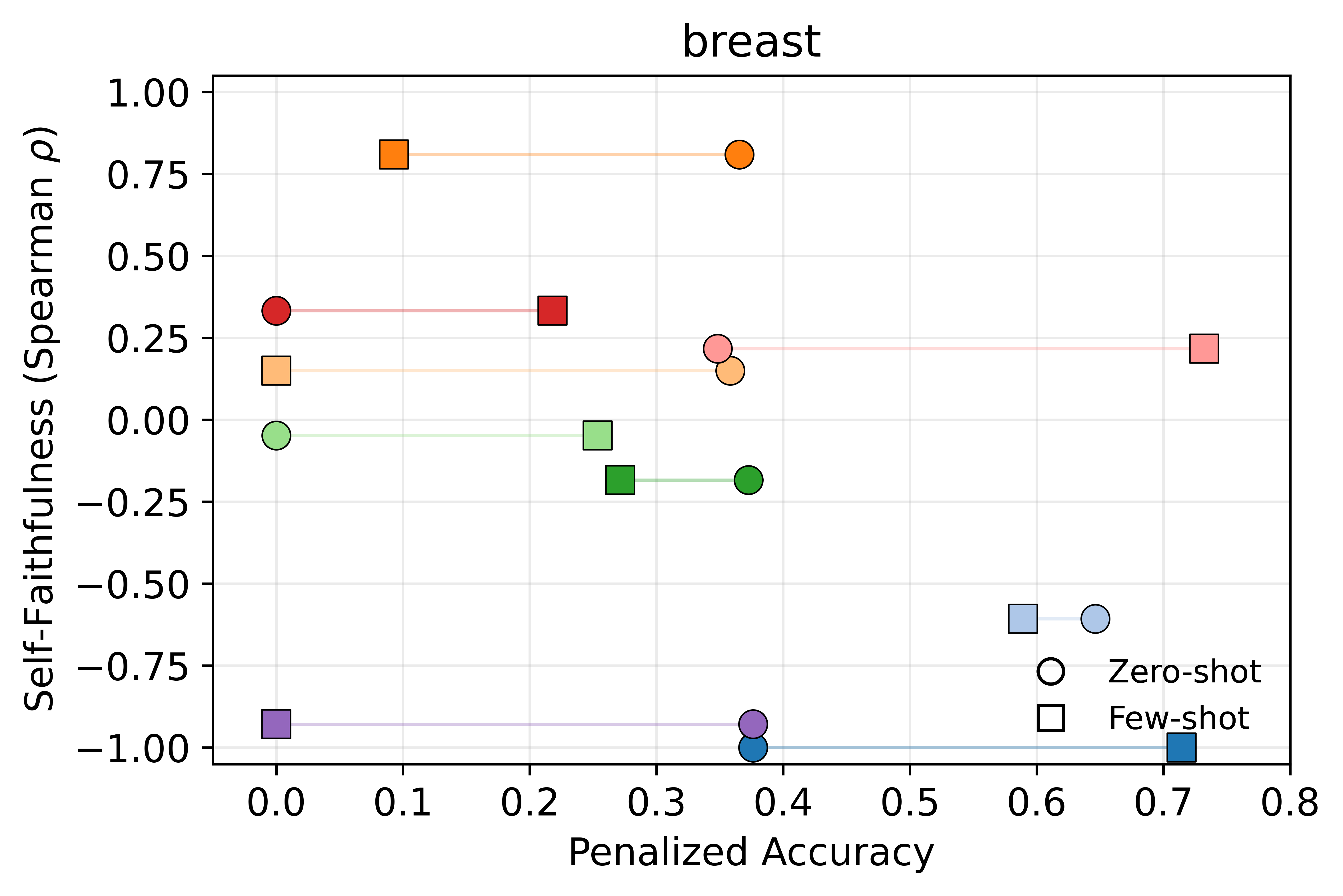}
        % \caption{Breast Cancer}
    \end{subfigure}%
    \hfill
    \begin{subfigure}[t]{.33\textwidth}
        \centering
        \includegraphics[width=\textwidth]{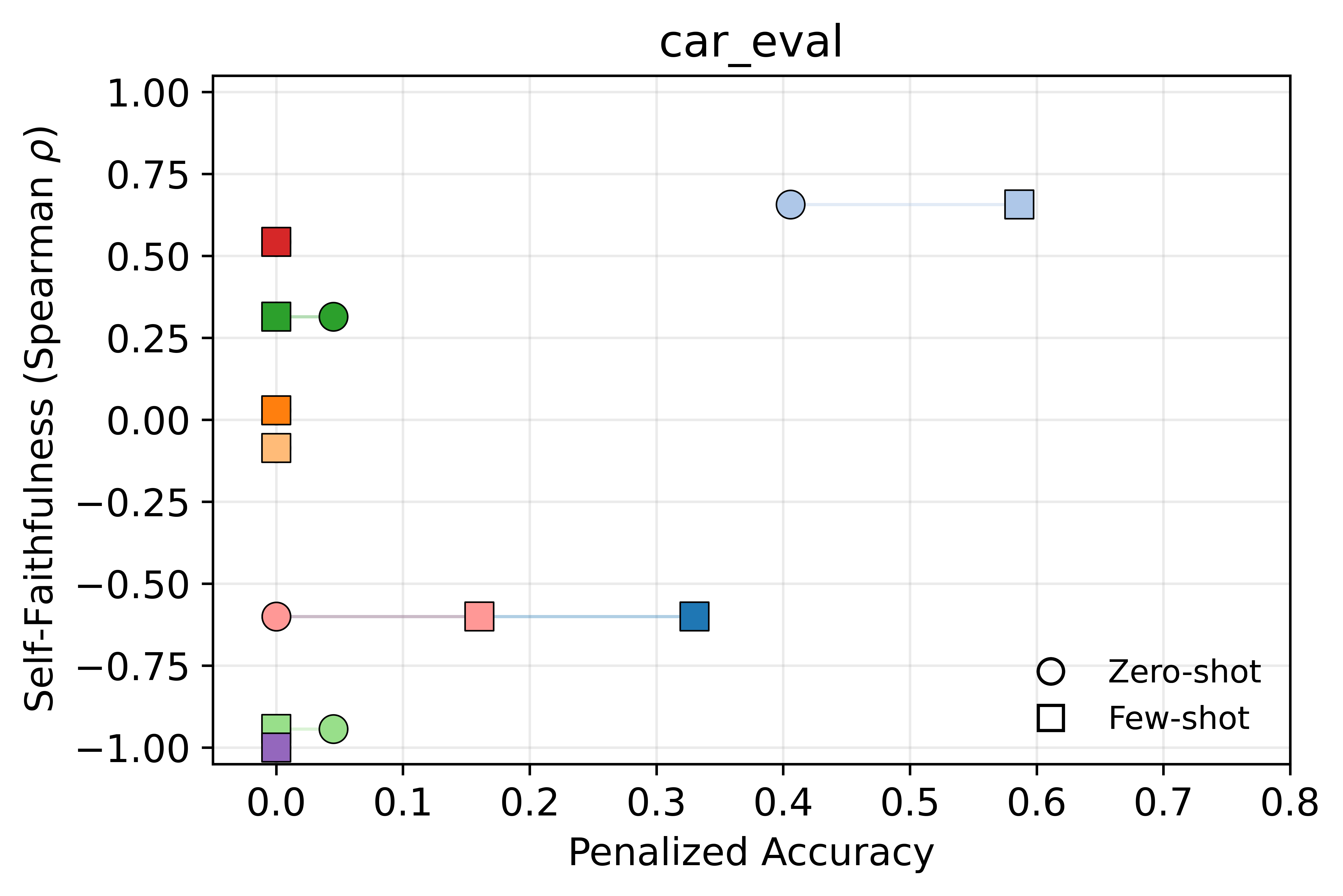}
        % \caption{Breast Cancer}
    \end{subfigure}%
    \vfill
    \begin{subfigure}[t]{.33\textwidth}
        \centering
        \includegraphics[width=\textwidth]{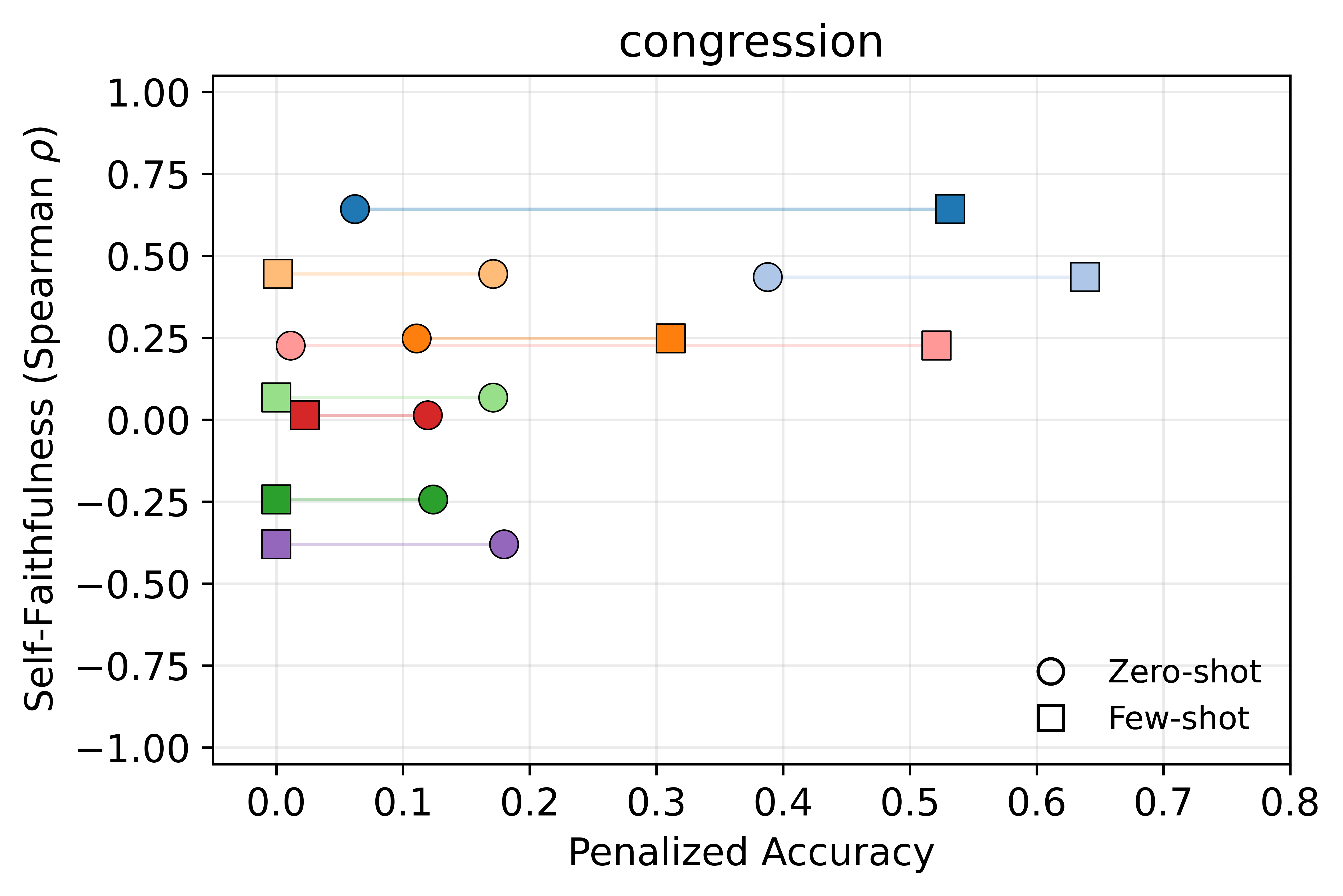}
        % \caption{Adult Income}
    \end{subfigure}%
    \hfill
    \begin{subfigure}[t]{.33\textwidth}
        \centering
        \includegraphics[width=\textwidth]{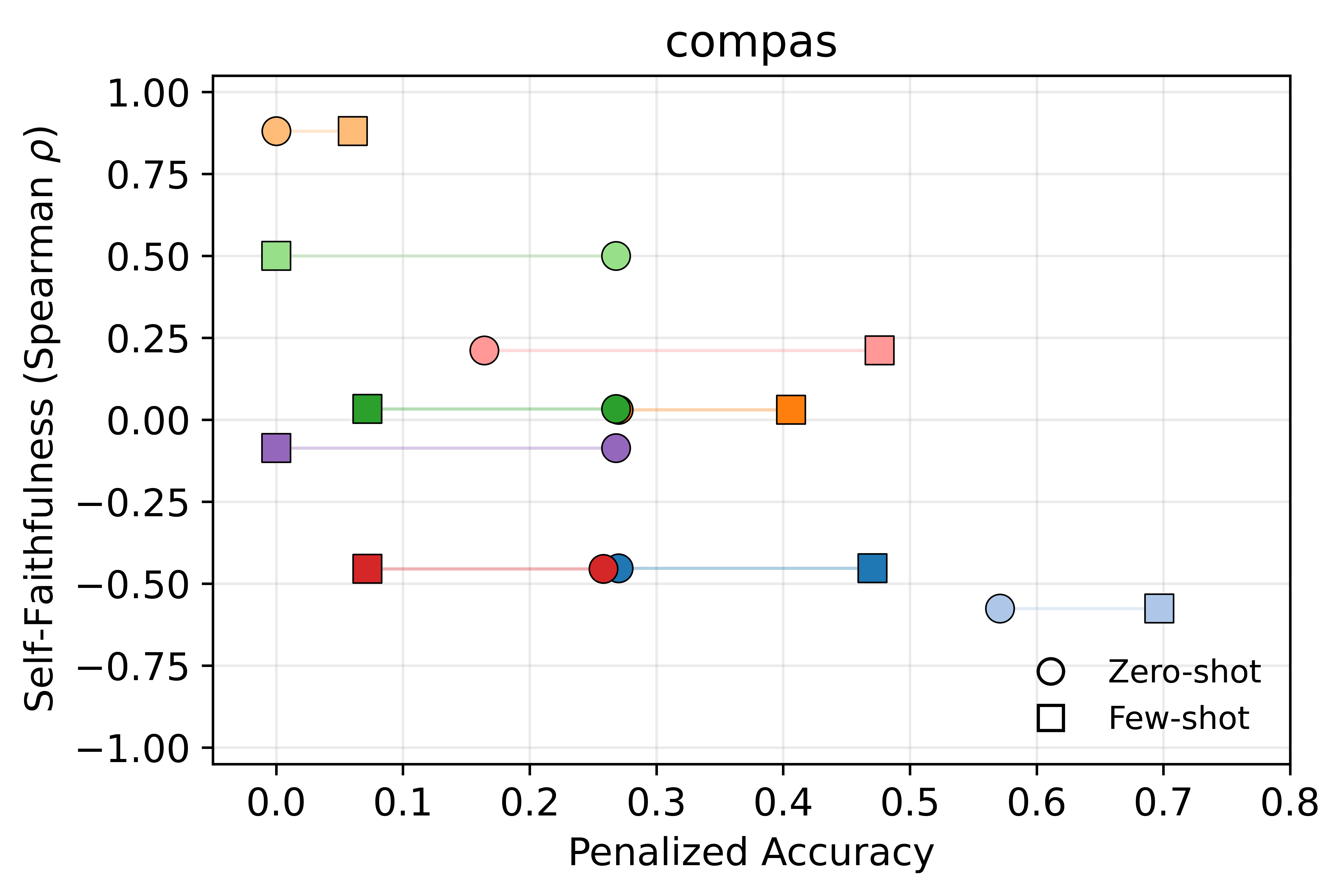}
        % \caption{Breast Cancer}
    \end{subfigure}%
    \hfill
    \begin{subfigure}[t]{.33\textwidth}
        \centering
        \includegraphics[width=\textwidth]{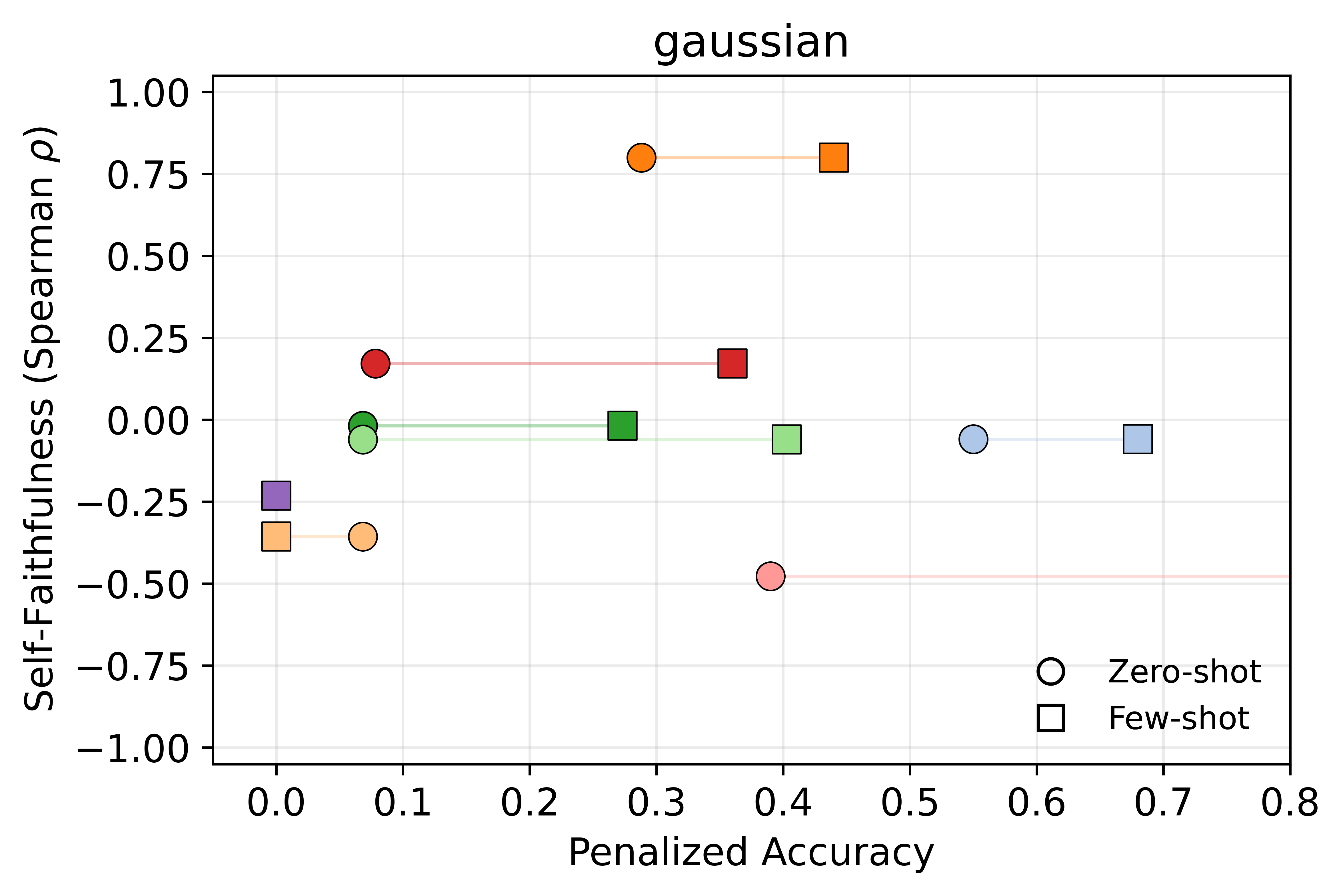}
        % \caption{Breast Cancer}
    \end{subfigure}%
\vfill
    \begin{subfigure}[t]{.33\textwidth}
        \centering
        \includegraphics[width=\textwidth]{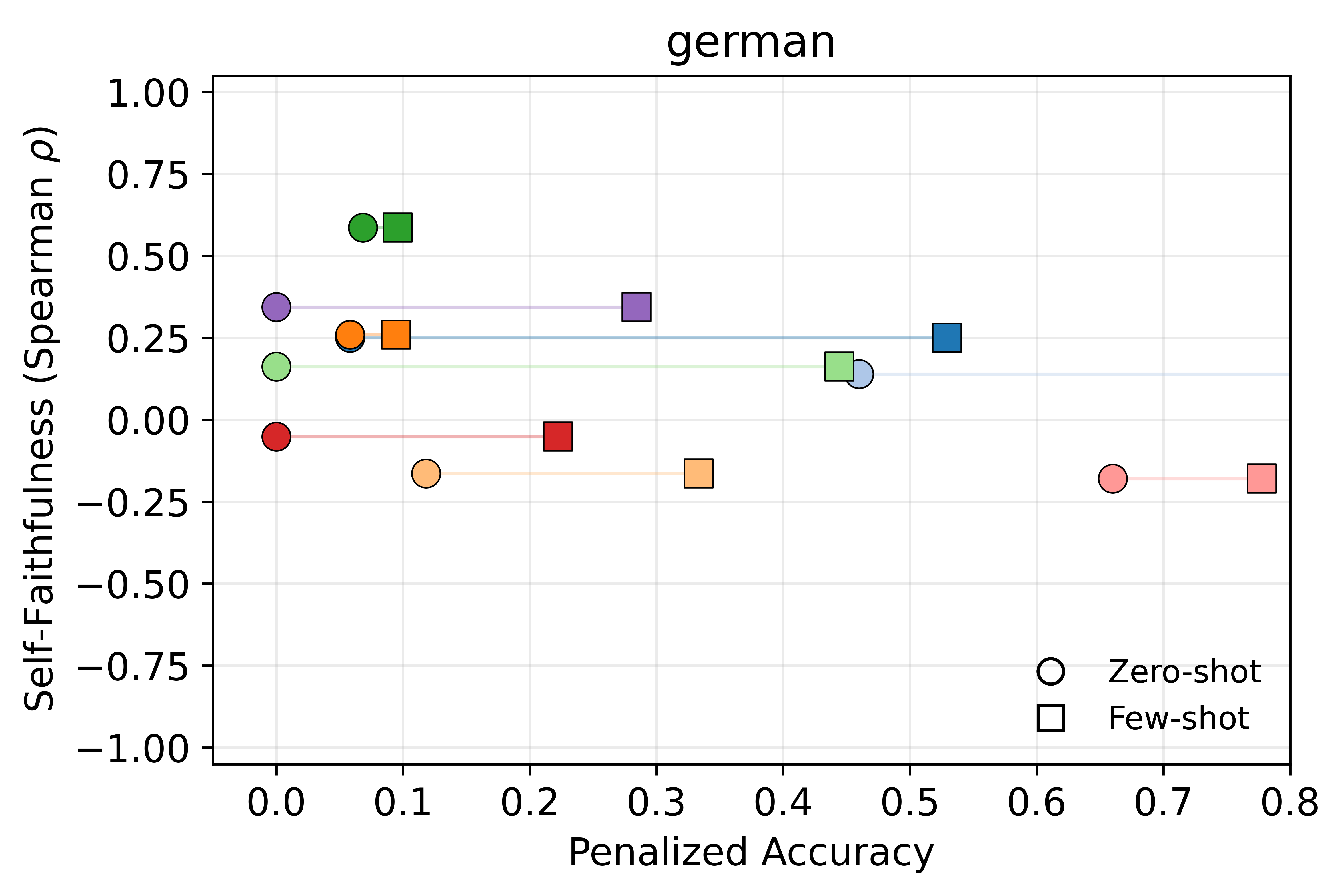}
        % \caption{Adult Income}
    \end{subfigure}%
    \hfill
    \begin{subfigure}[t]{.33\textwidth}
        \centering
        \includegraphics[width=\textwidth]{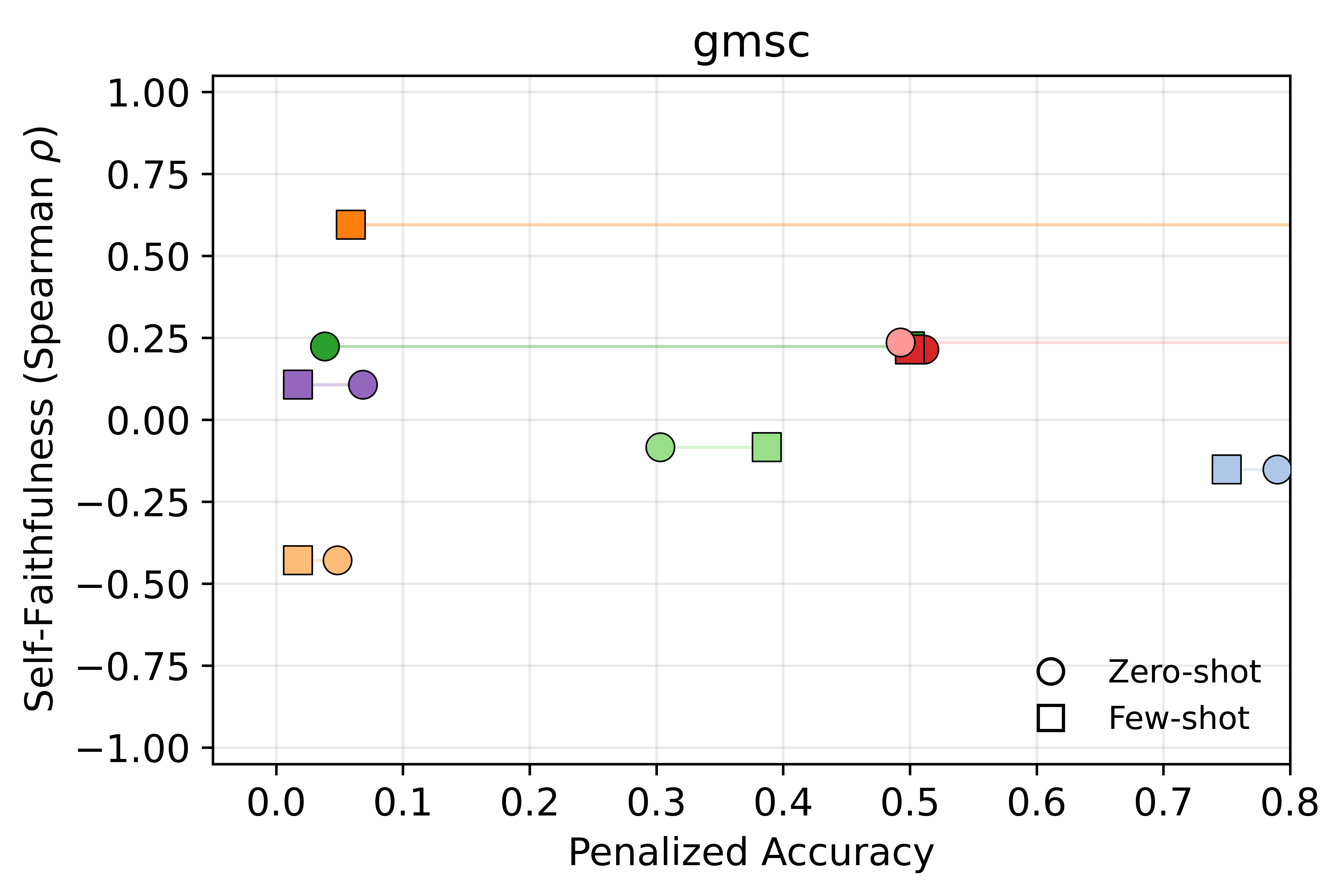}
        % \caption{Breast Cancer}
    \end{subfigure}%
    \hfill
    \begin{subfigure}[t]{.33\textwidth}
        \centering
        \includegraphics[width=\textwidth]{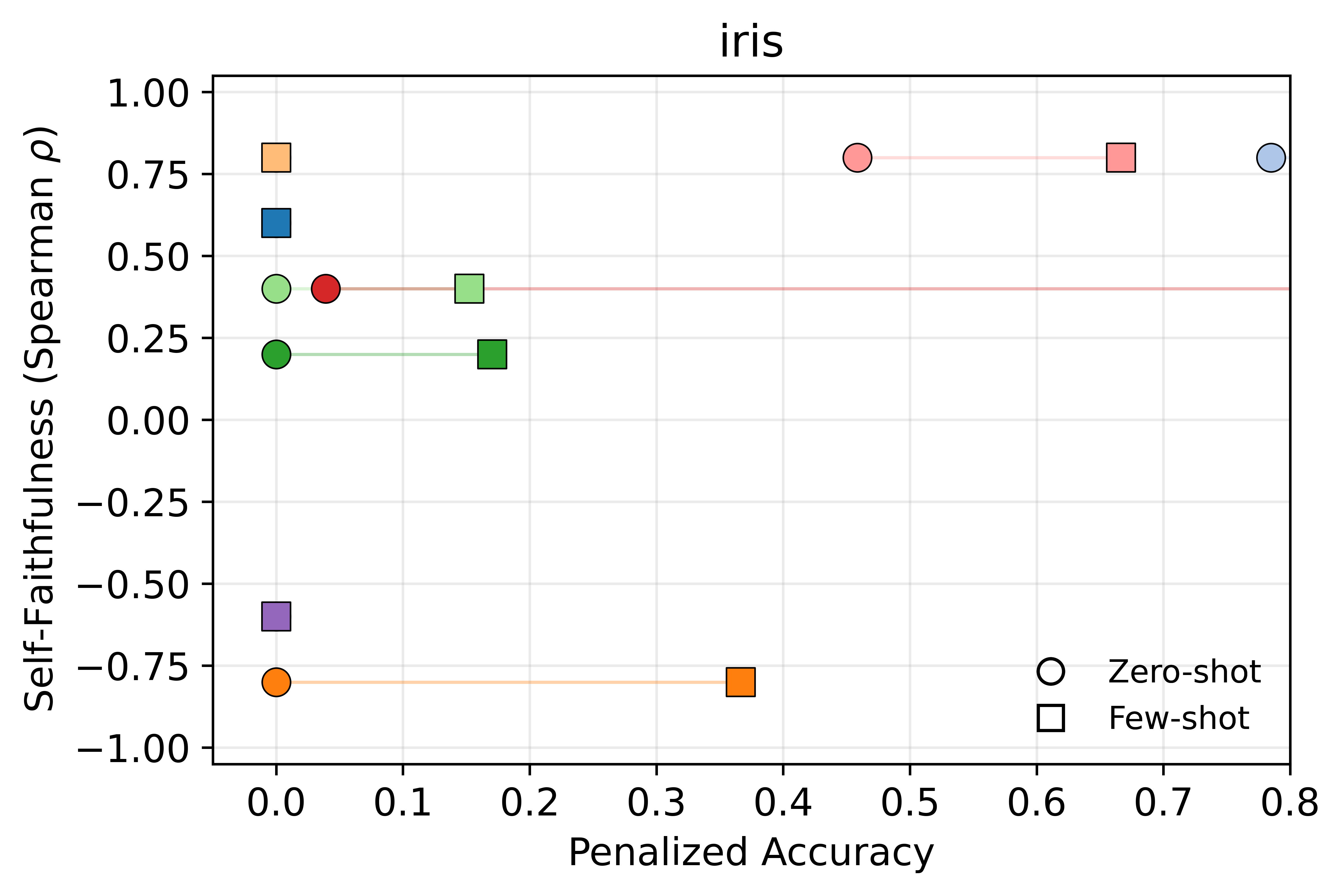}
        % \caption{Breast Cancer}
    \end{subfigure}%
\vfill
    \begin{subfigure}[t]{.33\textwidth}
        \centering
        \includegraphics[width=\textwidth]{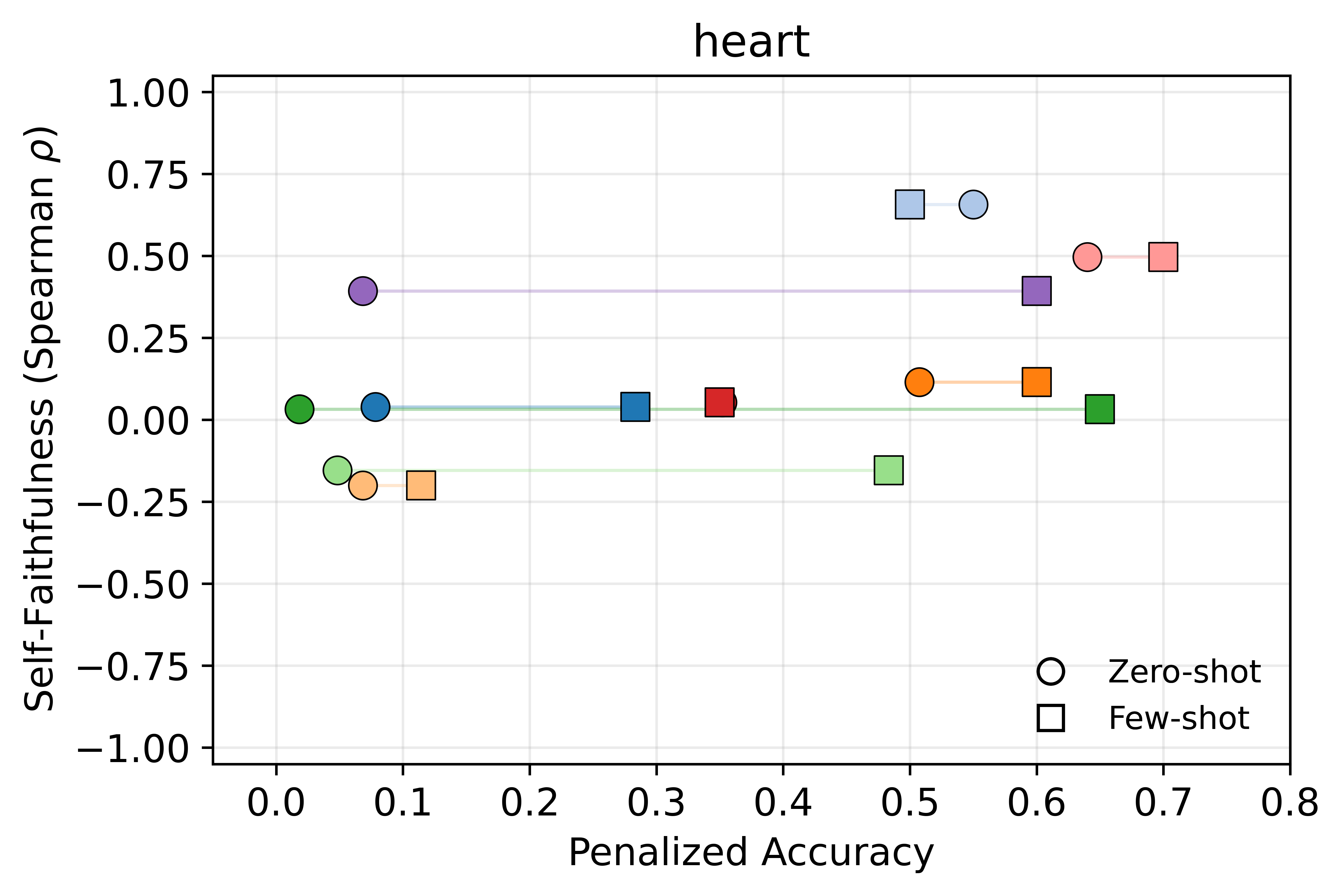}
        % \caption{Adult Income}
    \end{subfigure}%
    \hfill
    \begin{subfigure}[t]{.33\textwidth}
        \centering
        \includegraphics[width=\textwidth]{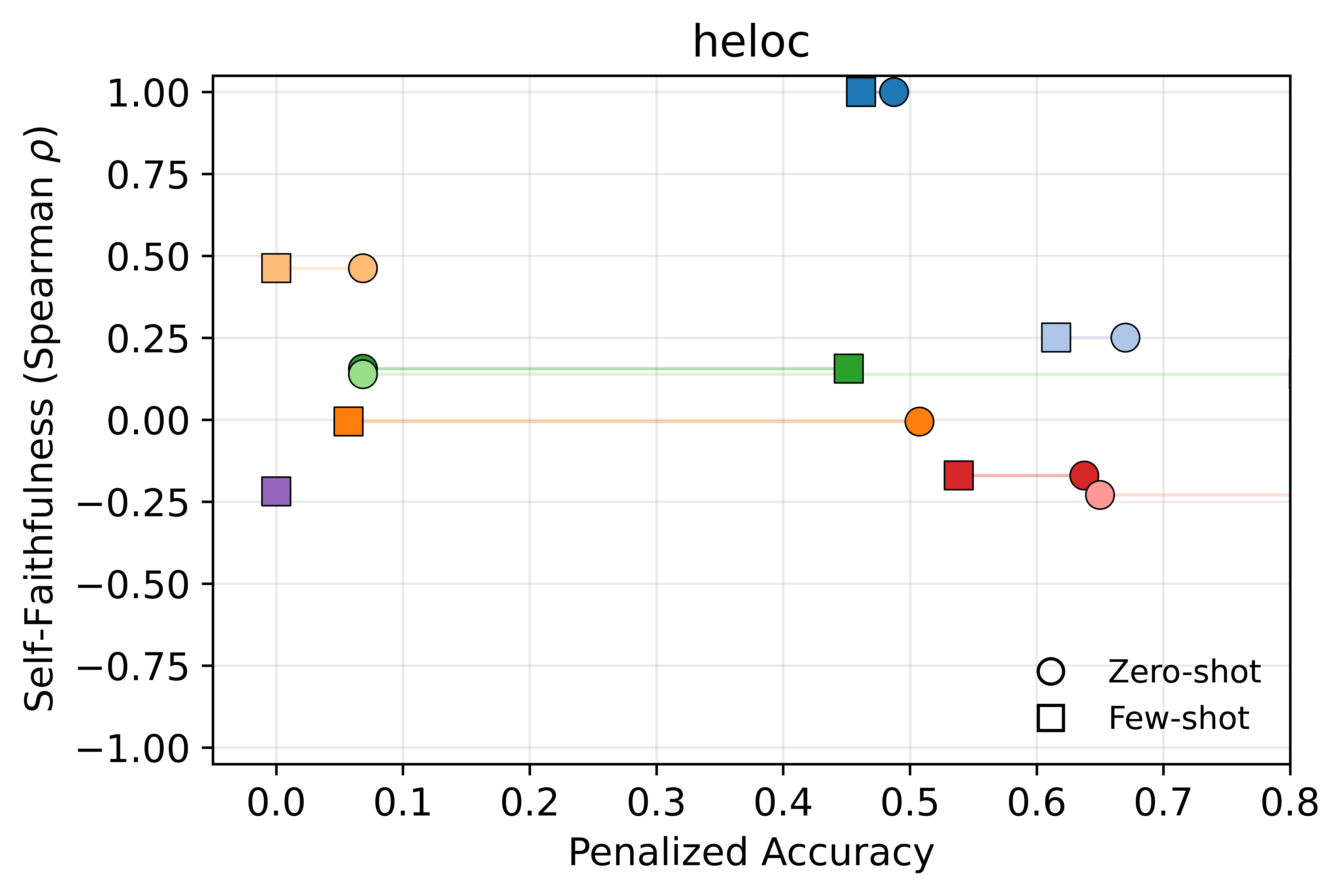}
        % \caption{Breast Cancer}
    \end{subfigure}%
    \hfill
    \begin{subfigure}[t]{.33\textwidth}
        \centering
        \includegraphics[width=\textwidth]{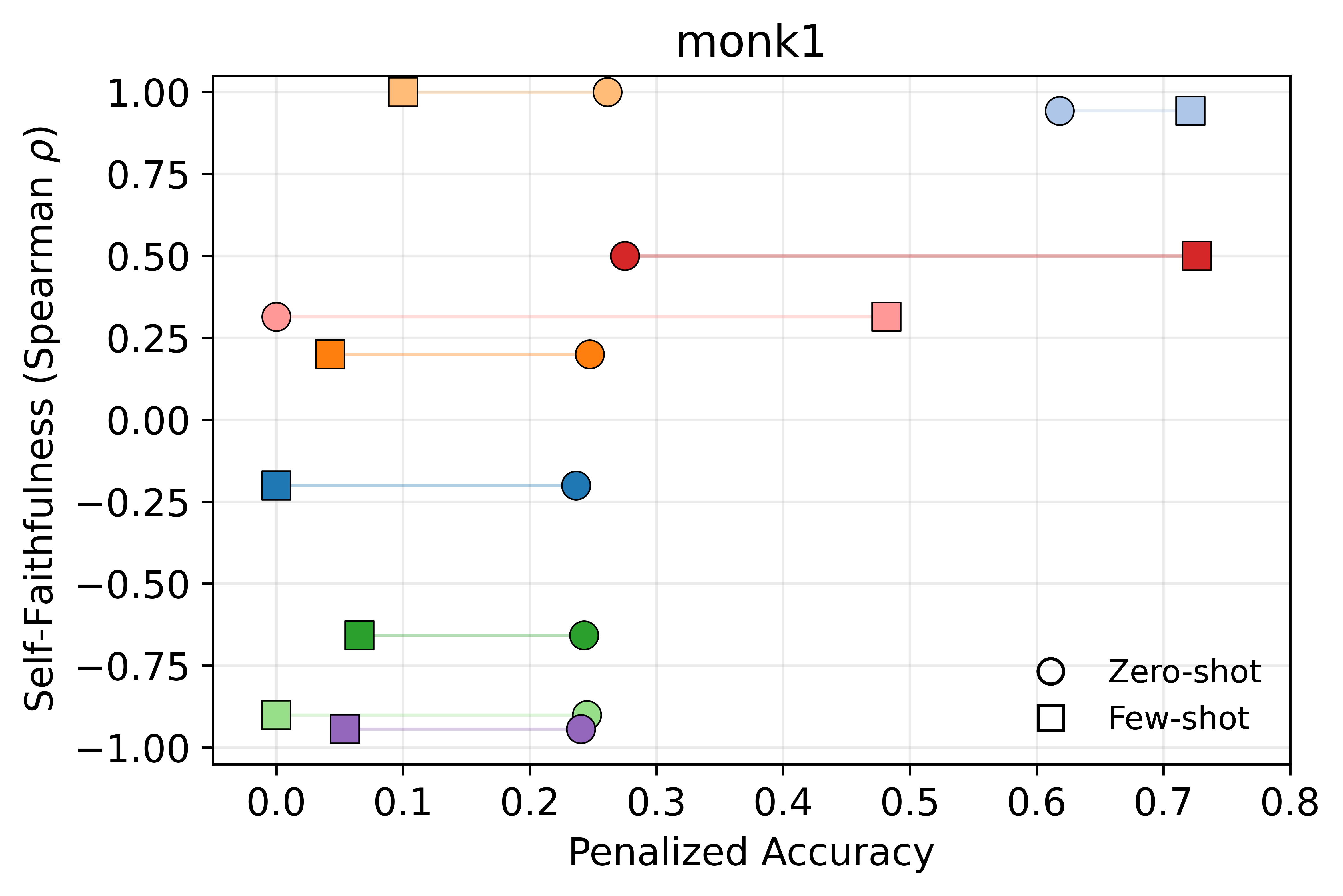}
        % \caption{Breast Cancer}
    \end{subfigure}%
\vfill
    \begin{subfigure}[t]{.33\textwidth}
        \centering
        \includegraphics[width=\textwidth]{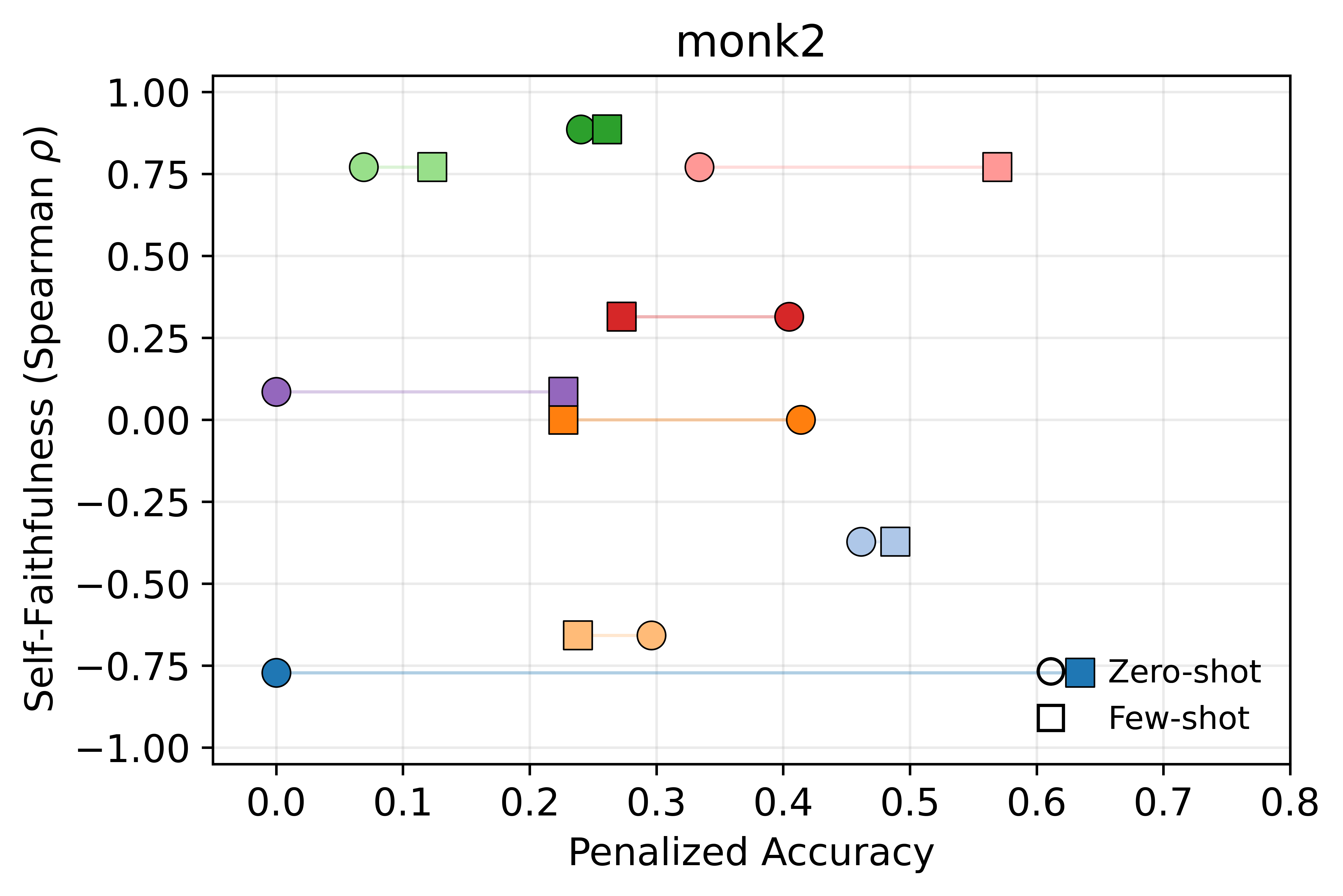}
        % \caption{Adult Income}
    \end{subfigure}%
    \hfill
    \begin{subfigure}[t]{.33\textwidth}
        \centering
        \includegraphics[width=\textwidth]{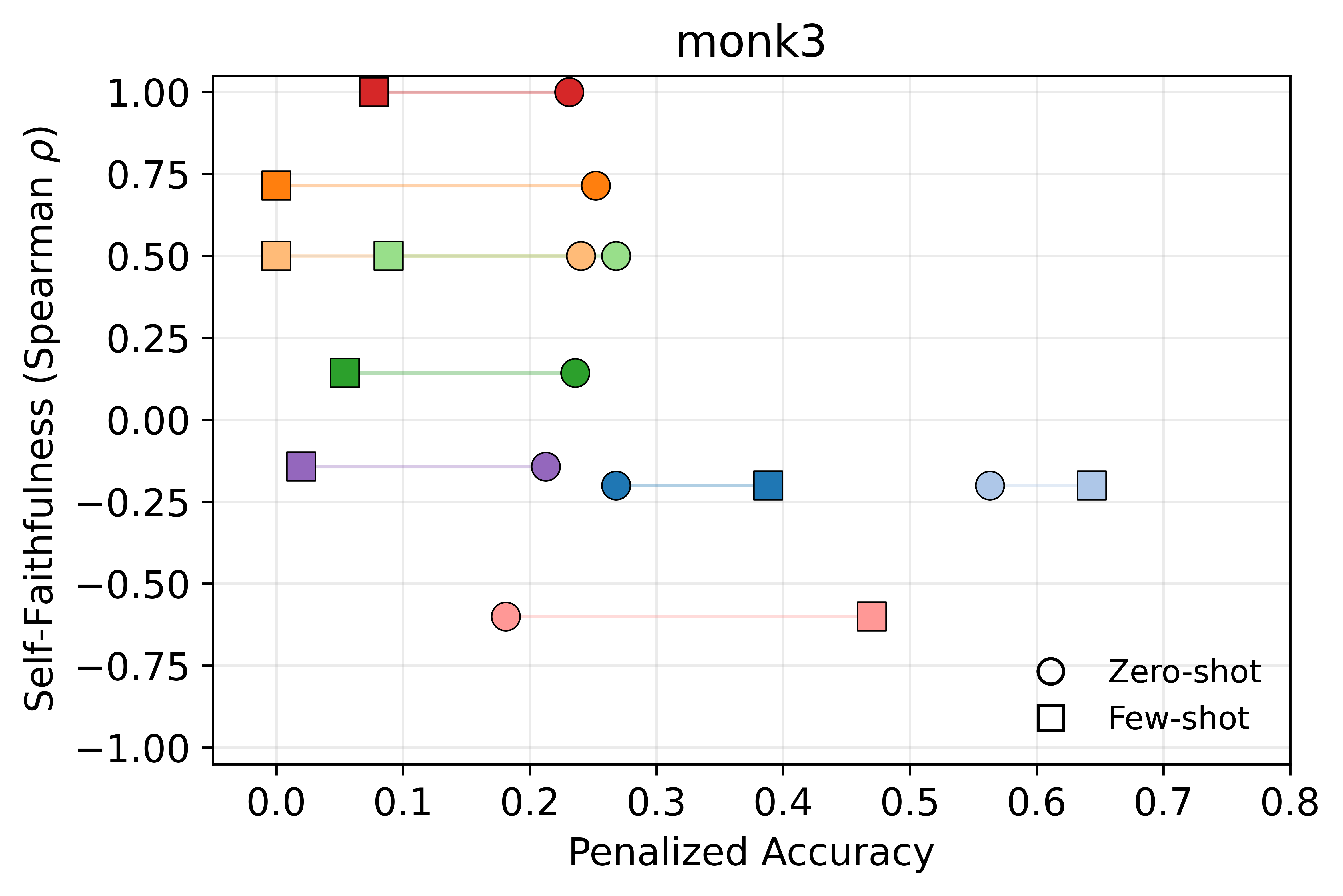}
        % \caption{Breast Cancer}
    \end{subfigure}%
    \hfill
    \begin{subfigure}[t]{.33\textwidth}
        \centering
        \includegraphics[width=\textwidth]{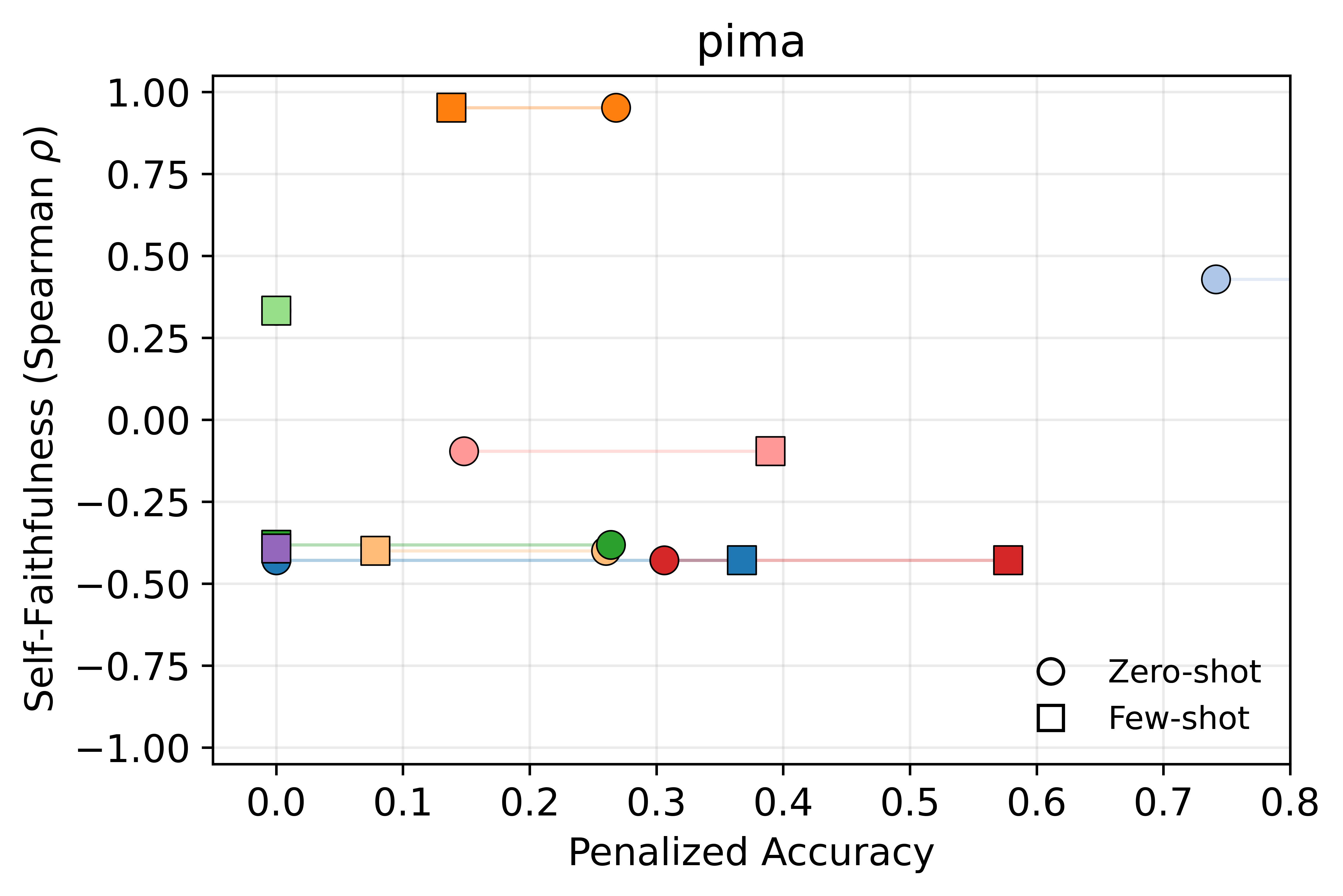}
        % \caption{Breast Cancer}
    \end{subfigure}%
    \vfill
    \begin{subfigure}[t]{\textwidth}
        \centering
        \includegraphics[width=\textwidth]{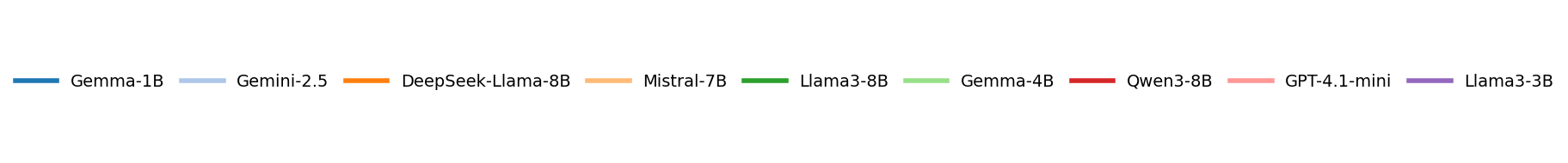}
        % \caption{Breast Cancer}
    \end{subfigure}%
\caption{Penalized Accuracy vs. Self-Decision Faithfulness Across All Datasets. 
Each subplot shows penalized accuracy (x-axis) against self-faith (y-axis), where self-faith defined by (Spearman’s $\rho$) between model-declared and behavioral (LAO) feature rankings. Zero-shot and few-shot results are shown with circles and squares, respectively. Models are colored constantly. Accuracy and faithfulness diverge for many model and dataset pairs (high-PenAcc with low $\rho$), i.e., accurate yet globally unfaithful.}
\label{fig:faithfulness}
\end{figure*}

\section{Results \& Discussion}
The following section presents our findings evaluating 9 LLMs with STaDS. 

\subsection{Comprehension Fidelity via $\Delta_{\text{acc}}$ \& \textsc{SelfAtt@\emph{k}}}

We first examine whether each model can handle \textbf{Comprehension Fidelity} task, namely \emph{reading the question, parsing the table, and adhering to the prescribed output format}. To this end, we visualize the distribution of \emph{format violations} across all 15 datasets in terms of the penalized accuracy difference ($\Delta_{\text{acc}}$) in Fig.~\ref{fig:acc_delta}, and report the Self-Attribution Recall (\textsc{SelfAtt@\emph{k}}) in Table~\ref{tab:au_metrics}.

\paragraph{Some models struggle to interpret and follow instructions.} As shown in Fig.~\ref{fig:acc_delta}, state-of-the-art instruction-tuned models, particularly \texttt{Gemini-2.5-Pro}, exhibit near-zero $\Delta_{\text{acc}}$ across most datasets, whereas smaller open-source models such as \texttt{Llama3-8B} experience performance drops of approximately 5–40\% due to straightforward formatting errors, e.g., exceeding expected output length and generating unknown labels. Interestingly, \texttt{GPT-4o} demonstrates that few-shot examples substantially enhance instruction-following capability, whereas smaller models (e.g., \texttt{DeepSeek-Llama-8B} and \texttt{Mistral-7B}) show little to no benefit, in some cases, even higher $\Delta_{\text{acc}}$ in few-shot settings, suggesting increased susceptibility to prompt variation rather than improved comprehension.

\paragraph{Even short responses pose challenges for strict instruction adherence.}
One might expect that long or verbose outputs are the primary cause of instruction violations, yet this assumption does not hold in our setting \cite{Tam2024speak}. Even when responses are expected to be short, such as in the self-attribution ranking prompt, which should contain only feature names, many models still fail to comply with the prescribed output format. For instance, models such as \texttt{DeepSeek-Llama-8B} and \texttt{Llama3-3B} consistently return exactly $k$ feature names for datasets like Adult Income (Table~\ref{tab:au_metrics}), demonstrating good length fidelity. In contrast, models like \texttt{Llama3-8B} (\textsc{SelfAtt@\emph{k}} = 0.73) and \texttt{Qwen3-8B} (0.44) often misidentify or omit relevant columns, indicating that instruction-following remains a persistent challenge independent of response length.

\subsection{Predictive Competence via PenAcc}

Following the instruction and table comprehension, this section mainly discusses whether models generate \textit{accurate} labels. This reveals if the model has failed to ground prior knowledge in the domain or apply it effectively to the task.

\paragraph{Mixed performance on short and long contexts.} While existing research commonly suggests that shorter contexts facilitate better model performance \cite{liu-etal-2024-lost}, our results indicate complexity beyond mere input length. For instance, \texttt{Gemini-2.5-Pro} achieves high accuracy (0.81) on the COMPAS dataset (39K tokens) yet shows significantly lower performance ($\sim$0.4) on the shorter Car Evaluation dataset. 
This discrepancy, presented in Table~\ref{tab:icu-main} and Fig.~\ref{fig:spider}, may arise from the multi-class nature of the Car Evaluation task, highlighting that \textit{context length alone does not fully explain model performance variations}. However, the multi-class nature from Iris does not account for this across all models. This finding will be further explained in Sec.~\ref{subsec:results-df}.

Furthermore, performance analyses (Fig.~\ref{fig:spider}, panels (a) and (b)) demonstrate that models consistently achieve better results on intermediate-sized datasets, such as Breast Cancer (17K tokens), Iris (7K tokens), and Heart Disease (12K tokens), compared to larger contexts like COMPAS and \textsc{Pima} (31K tokens). These observations suggest a complex interaction between context length, task complexity, and model capability. 

% underscoring the need for careful consideration beyond mere token count in evaluating model understanding.

\paragraph{Acting as a general professional across domains.}
PenAcc jointly captures a model’s ability to \emph{follow formatting rules}
and \emph{produce the correct label},
so it naturally reflects whether an LLM can ``act like a well-trained professional'' across heterogeneous structured tabular settings.
Fig.~\ref{fig:spider} overlays PenAcc in radar form for every model–dataset pair, while Table~\ref{tab:icu-main} reports the best zero/few-shot scores.

\texttt{Gemini-2.5-Pro} dominates, topping 10/15 datasets and exhibiting almost identical polygons in the zero/few-shot plots, evidence of strong domain generalisation. Open-source checkpoints tell a more fragmented story:
compact models such as \texttt{Llama3-3B} or \texttt{Gemma-1B} shine on small medical tables (e.g.\ Heart Disease) yet collapse on wide-schema sets like COMPAS;
even larger \texttt{Llama3-8B} and \texttt{Qwen3-8B} sometimes fail simply because they mis-parse column headers or drift from the required output format rather than because they ``do not know'' the task.

\paragraph{Zero-shot vs. few-shot: demonstrations matter.}
We also compare zero-shot and few-shot performance to ask how much ``being shown how to behave'' helps a model act like that professional. Zero-shot in STaDS is effectively pure retrieval and schema inference: \textit{the model must infer label semantics, column roles, and decision logic from the instruction and table alone}. Few-shot augments this with a handful of in-context demonstrations, i.e., explicit exemplars of how a domain expert would label similar rows. Across all 15 datasets, providing these demonstrations consistently improves the \emph{best achievable} PenAcc (see Table~\ref{tab:icu-main}), which is an upper-bound view (see detailed results in Appendix Tables \ref{tab:adult-income-detailed} - \ref{tab:icu-pa}). Averaged over all datasets, the best few-shot PenAcc exceeds the best zero-shot PenAcc by an absolute +0.15, corresponding to a 27\% relative improvement. This pattern holds even for datasets where zero-shot performance is already strong, such as Iris improved from 0.79 to 1.00, and Give Me Some Credit improved from 0.83 to 0.92. In other words, demonstrations do not just rescue weaker models, and they sharpen already competent ones.

\noindent \textbf{Answer to RQ1.}
Frontier LLMs exhibit \emph{partial} professional-level competence out of the box: the strongest models can often produce valid, well-formatted predictions across diverse decision settings, but this behavior is not yet consistent across domains or model families. Demonstrations remain crucial, indicating that current models still require task-specific guidance rather than universally understanding tabular decision rules in a zero-shot setting.

\subsection{Decision Faithfulness via Self-claimed, LAO, and Statistical Feature Importance Ranking}\label{subsec:results-df}
We then evaluate LAO Magnitude ($\sigma_{\text{LAO}}$), \textsc{Self-Faith} ($\rho$), and \textsc{SelfAtt@\,$k$} to assess decision faithfulness, results reported in Fig.~\ref{fig:lao_delta},  Fig.~\ref{fig:faithfulness}, Fig.~\ref{fig:mi_box}, and Table \ref{tab:rho_all_main}.

\paragraph{Models don't always act like what they claim.} 
Table~\ref{tab:au_metrics} reports these metrics on \textsc{Adult~Income}. Other datasets are provided in the Appendix Figs. \ref{fig:lao_delta_app_1} and \ref{fig:lao_delta_app_2}. \texttt{DeepSeek-Llama-8B} exhibits the largest LAO magnitude ($\rho = 0.24$) together with a large $\sigma_{\text{LAO}}$, signaling that its decisions hinge on a \emph{small}, clearly identifiable subset of features. This echoes \textit{complexity} scores used in recent XAI work~\citep{Nauta2021Survey}.
\texttt{Gemini-2.5-Pro} reveals an appealing trade-off: mid-level sparsity ($\Delta=0.07$) but a positive \textsc{Self-Faith} ($\rho = 0.25$), meaning its stated importance ordering aligns with observed behaviour. By contrast, \texttt{Mistral-7B} and \texttt{Llama-3-3B} score a markedly negative \textsc{Self-Faith} ($-0.54$ and $-0.34$), evidencing that the features it claims to rely on differ from those that truly drive its predictions.

Heatmaps in Fig.~\ref{fig:lao_delta} support these findings. For \textsc{AdultIncome}, \texttt{Gemini-2.5-Pro} and \texttt{DeepSeek-Llama-8B} predominantly depend on socio‑economic attributes such as \textit{education‑num} and \textit{occupation}, whereas \texttt{Qwen3‑8B} spreads its reliance across almost every feature, producing a harder‑to‑interpret signature. A similar pattern emerges on \textsc{BreastCancer} in Fig.~\ref{fig:lao_delta} (b). It is intuitive to plot faithfulness regimes (PenAcc $\times$ Self-Faith).
Fig.~\ref{fig:faithfulness} plots PenAcc against~$\rho$ and reveals the divergence between predictive performance and faithfulness: (i) \textbf{Accurate \& Faithful}: \texttt{Gemini-2.5-Pro} on Iris reaches $\text{PenAcc}\!\approx\!1.0$ and $\rho\!\approx\!0.9$, indicating genuine concept understanding; (ii) \textbf{Inaccurate \& Faithful}: \texttt{GPT-4.1-mini} on Iris reaches $\text{PenAcc}\!\approx\!0.6$ and $\rho\!\approx\!0.9$, indicating genuine concept understanding; (iii) \textbf{Accurate \& Unfaithful}: \texttt{GPT-4.1-mini} on Breast Cancer achieves high PenAcc but $\rho\approx 0.2$; and (vi) \textbf{Inaccurate \& Unfaithful}: Most small open-source models remain below $0.5$ PenAcc and $0.4$ $\rho$, failing both comprehension and faithful reasoning.

\paragraph{Understanding is NOT Statistical Dependency.}
Understanding whether LLMs uncover behavioral reliance or merely exploit statistical co-occurrence remains an open question. 
By comparing feature--label dependencies in the data, measured by \textbf{Normalized Mutual Information} (NMI) \footnote{Complemented by Cramér's V for categorical or multi-class targets and Pearson’s~$r$/Spearman's~$\rho$ for numerical targets.}, with models' self-claimed and behavioral attributions (via LAO), we observe that most tabular tasks exhibit weak intrinsic dependence 
(mean NMI~$<0.08$ for 12/15 datasets). 
This suggests that accurate predictions cannot be trivially explained by dataset regularities alone. 
High-NMI, rule-based tasks such as Iris or Congressional Voting naturally afford better in-context learning, and explains why models make accurate predictions in some cases (see Fig. \ref{fig:mi_box}). However, this trend does not generalize, datasets with similar dependence, such as Monk3, show no comparable gains. 
Across domains, the features most strongly correlated with labels 
(e.g., \texttt{education} and \texttt{capital-gain} in \textsc{Adult}, or \texttt{petal-length} and \texttt{petal-width} in Iris) 
correspond to intuitive decision factors, yet these associations reflect statistical covariation rather than genuine behavioral reliance. 
Hence, while LLMs can detect such correlations, their ``understanding’’ often remains pattern-based rather than mechanistic.

\paragraph{``Triangulated'' Faithfulness.}
Now that we collected self-stated feature importance ($\pi_{\text{self}}$), statistical dependence measured by NMI ($\pi_{\text{NMI}}$), and true behavioral reliance measured by LAO ranking ($\pi_{\text{LAO}}$) across datasets and models. By correlating any two of these rankings, we establish a ``triangulated'' view of faithfulness. We already introduced $\rho(\pi_{\text{self}}, \pi_{\text{LAO}})$ as a metric for decision faithfulness, while the other two correlations highlight the degree of \textbf{statistical} alignment between the models' stated importance and their actual reliance on features. Part of results are provided in Table~\ref{tab:rho_all_main} and complete results can be found in Appendix Table \ref{tab:rho_all}.
Results reveals that \textbf{LLMs tend to explain plausibly but act unfaithfully}. 
Across datasets, the correlation between self-declared feature importance and statistical dependence consistently exceeds that between behavioral attributions and true dependencies. This infers that models often emphasize features that are \emph{obvious from the data} rather than those truly driving their predictions. Only a few model--dataset pairs, such as \texttt{Gemma-4B} on \textsc{Adult}, \texttt{DeepSeek-Llama-8B} on \textsc{GMSC}, and \texttt{DeepSeek-Llama-8B} on Monk1, demonstrate significant positive decision faithfulness ($\rho(\pi_{\text{LAO}}, \pi_{\text{NMI}})>0.5$, $p<.05$). In short, current LLMs often tend to \emph{explain like statisticians}, tracking correlations, but rarely \emph{behave like reliable reasoners}.

\begin{table}[htbp]
\small
\centering
\caption{``Triangulated'' Faithfulness Across Models and Datasets.}
\label{tab:rho_all_main}
\begin{tabular}{@{}llrrr@{}}
\toprule
Dataset & Model & $\rho(\pi_{\text{self}}, \pi_{\text{LAO}})$ & $\rho(\pi_{\text{self}}, \pi_{\text{NMI}})$ & $\rho(\pi_{\text{LAO}}, \pi_{\text{NMI}})$ \\ \midrule
Adult Income 
 & Gemma-4B & 0.552$^\dagger$ [0.098] & 0.394 [0.260] & 0.547$^*$ [0.043] \\
 & Gemini-2.5-Pro & 0.253 [0.383] & 0.477$^\dagger$ [0.085] & 0.187 [0.523] \\
 & DeepSeek-Llama-8B & 0.240 [0.409] & 0.301 [0.296] & -0.187 [0.523] \\
 & GPT-4.1-mini & -0.015 [0.958] & 0.240 [0.409] & -0.618$^*$ [0.019] \\
Breast Cancer
 & DeepSeek-Llama-8B & 0.810$^*$ [0.015] & 0.586 [0.127] & 0.009 [0.982] \\
 & GPT-4.1-mini & 0.217 [0.576] & 0.775$^*$ [0.014] & 0.000 [1.000] \\
Car Evaluation
 & Gemini-2.5-Pro & 0.657 [0.156] & 0.886$^*$ [0.019] & 0.314 [0.544] \\
 & GPT-4.1-mini & -0.600 [0.208] & 0.657 [0.156] & -0.943$^*$ [0.005] \\
 & Qwen3-8B & 0.543 [0.266] & 0.257 [0.623] & 0.429 [0.397] \\
COMPAS
 & Mistral-7B & 0.881$^*$ [0.004] & 0.095 [0.823] & 0.285 [0.425] \\
 & Llama3-8B & 0.033 [0.932] & -0.417 [0.265] & 0.406 [0.244] \\
Iris
 & Gemini-2.5-Pro & 0.800 [0.200] & 1.000$^*$ [0.000] & 0.800 [0.200] \\
 & GPT-4.1-mini & 0.800 [0.200] & 1.000$^*$ [0.000] & 0.800 [0.200] \\
Monk1
 & DeepSeek-Llama-8B & 0.200 [0.704] & 0.676 [0.140] & 0.845$^*$ [0.034] \\
 & Mistral-7B & 1.000$^*$ [0.000] & 0.500 [0.667] & -0.068 [0.899] \\
Pima Diabetes
 & DeepSeek-Llama-8B & 0.952$^*$ [0.000] & -0.286 [0.493] & -0.286 [0.493] \\
 & GPT-4.1-mini & -0.095 [0.823] & 0.905$^*$ [0.002] & -0.333 [0.420] \\
 & Qwen3-8B & -0.429 [0.289] & 0.857$^*$ [0.007] & -0.286 [0.493] \\
\bottomrule
\end{tabular}

\vspace{0.5em}
\emph{Notes.} Stars: $^*$ $p{<}.05$, $^\dagger$ $p{<}.10$. Brackets show $p$-values.
\end{table}

% In contrast, many others exhibit ``plausible-but-not-behavioral’’ patterns, and some even show anti-causal reliance, 
% where models act counter to the data’s ground truth 
% (e.g., \texttt{GPT-4.1-mini} on \textsc{Adult} and Car Evaluation). 
% These findings highlight that high explanation plausibility or internal honesty does not guarantee causal grounding: 
% models can appear self-consistent yet still reason through spurious features. 
% Overall, current LLMs often \emph{explain like statisticians}---tracking correlations---but rarely \emph{behave like causal reasoners}. 

% Achieving robust causal faithfulness will require future methods that explicitly align behavioral reliance 
% (e.g., LAO-based feature effects) with ground-truth structure while penalizing anti-causal patterns, 
% especially in safety-critical decision domains.

\noindent
\textbf{Answer to RQ2.}
Current LLMs frequently state rationales diverging from behavioral reliance ($\rho(\pi_{\text{self}}, \pi_{\text{LAO}})$) even when PenAcc is high.
Correlations with dataset regularities ($\rho(\pi_{\text{self}}, \pi_{\text{NMI}})$) tend to exceed behavioral alignment ($\rho(\pi_{\text{NMI}}, \pi_{\text{LAO}})$), suggesting that models explain like statisticians (plausible correlations) more often than they act like reliable professionals (stable reliance on governing factors).

\subsection{Conclusion}
This work introduced the Structured Tabular Decision Simulations (STaDS) protocol, a principled framework for evaluating whether LLMs \textit{understand} decision tasks rather than merely reproduce surface-level answers. By operationalizing understanding as the ability to identify and rely on correct decision factors, STaDS jointly evaluates predictive competence and attributional faithfulness across diverse, domain-grounded settings.
Our large-scale analysis of 9 LLMs over 15 tabular domains reveals a persistent gap between accuracy and understanding. While frontier models often achieve high predictive performance, they frequently violate global faithfulness, producing correct answers for the wrong reasons or inconsistently relying on relevant factors. Only a small subset of model–dataset pairs exhibit both accuracy and stable reliance on domain-relevant factors, indicating that expert-like generalization remains limited even in frontier models.
These findings highlight the need to move beyond instance-level prediction and reasoning traces, and toward \textbf{global assessments of model understanding}. Future work should explore how fine-tuning, causal supervision, and interactive alignment with human experts can promote stable, faithful decision behavior across domains. STaDS offers a reproducible and extensible foundation for such efforts, bridging the evaluation of accuracy, interpretability, and explainability toward a more complete assessment of true understanding in LLMs.

\bibliography{main}
\bibliographystyle{tmlr}

\clearpage

\appendix
\section*{Appendix}
The appendix provides extended details of our experimental setup and additional results supporting the main text.

% ========= Implementation Details ===========

\section{Implementation Details.}
\label{sec:app-implementation-details}
\begin{figure}[h!]
    \centering
    \includegraphics[width=.85\linewidth]{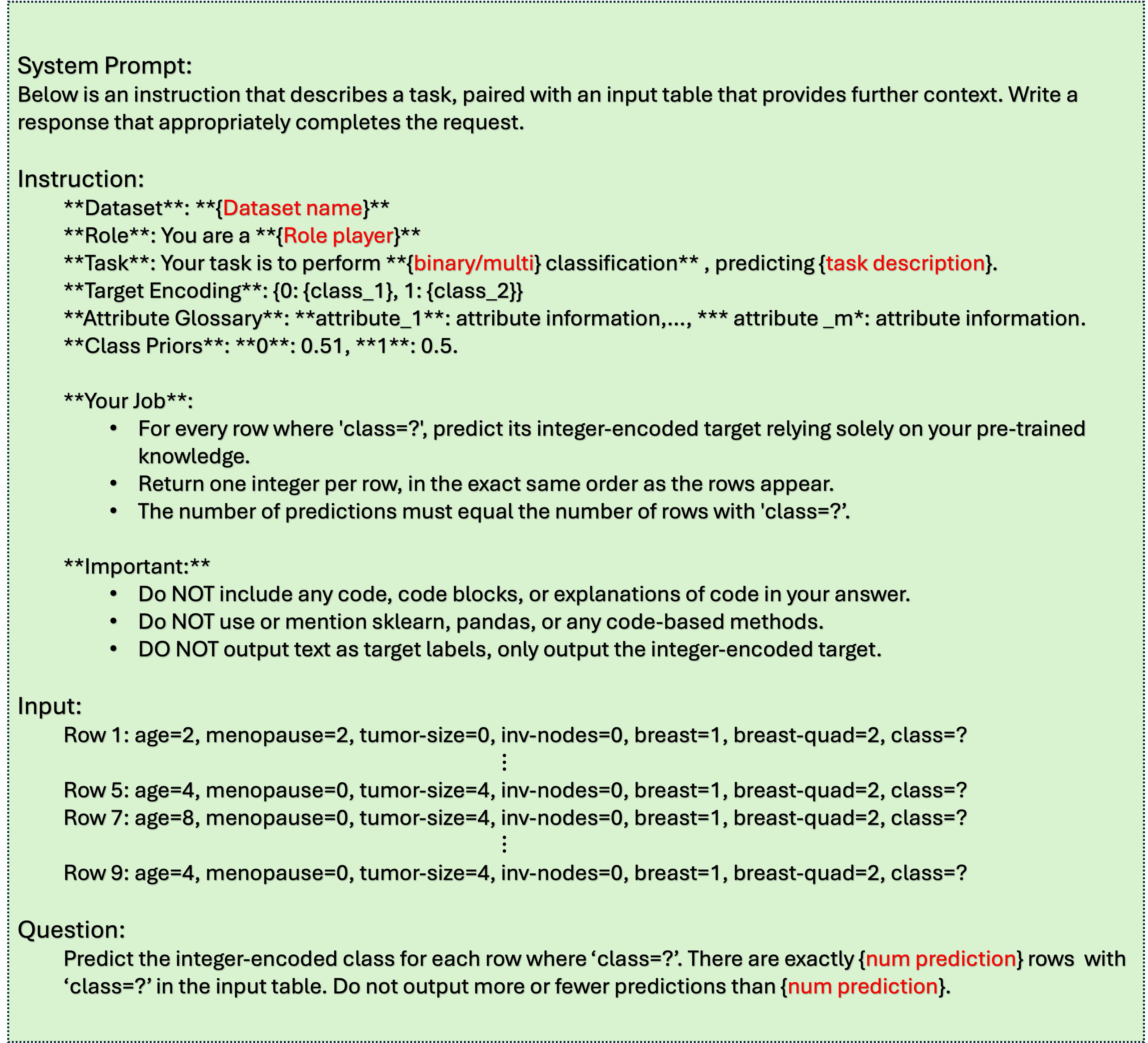}
    \caption{\textbf{Zero-shot prompt composition.} This setup is designed to evaluate: (i) the model's ability of instruction/table comprehension, and (ii) its intrinsic, pre-trained knowledge in specific role, serving as zero-shot baselines.}
    \label{fig:prompt_zero}
\end{figure}

\begin{figure}
    \centering
    \includegraphics[width=.85\linewidth]{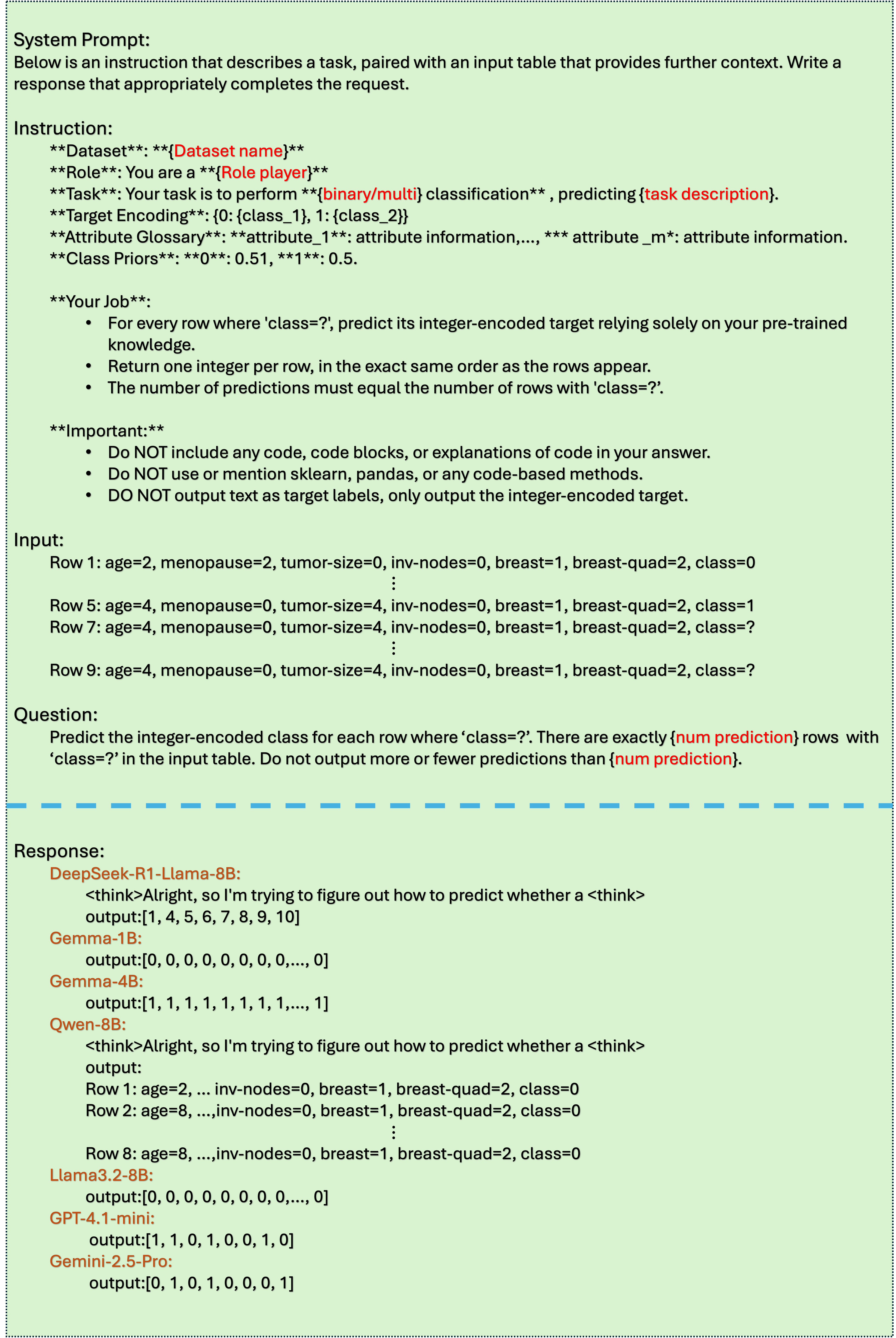}
    \caption{\textbf{Few-shot prompt composition.} The setup is designed to evaluate (i) the model’s capacity for in-context generalization beyond zero-shot baselines, and (ii) its ability to jointly parse instructions and structured tabular inputs. The bottom panel presents model outputs for the same prediction prompt on the example dataset (Breast Cancer) under the few-shot configuration.}
    \label{fig:pred_response}
\end{figure}

\begin{figure}
    \centering
    \includegraphics[width=.85\linewidth]{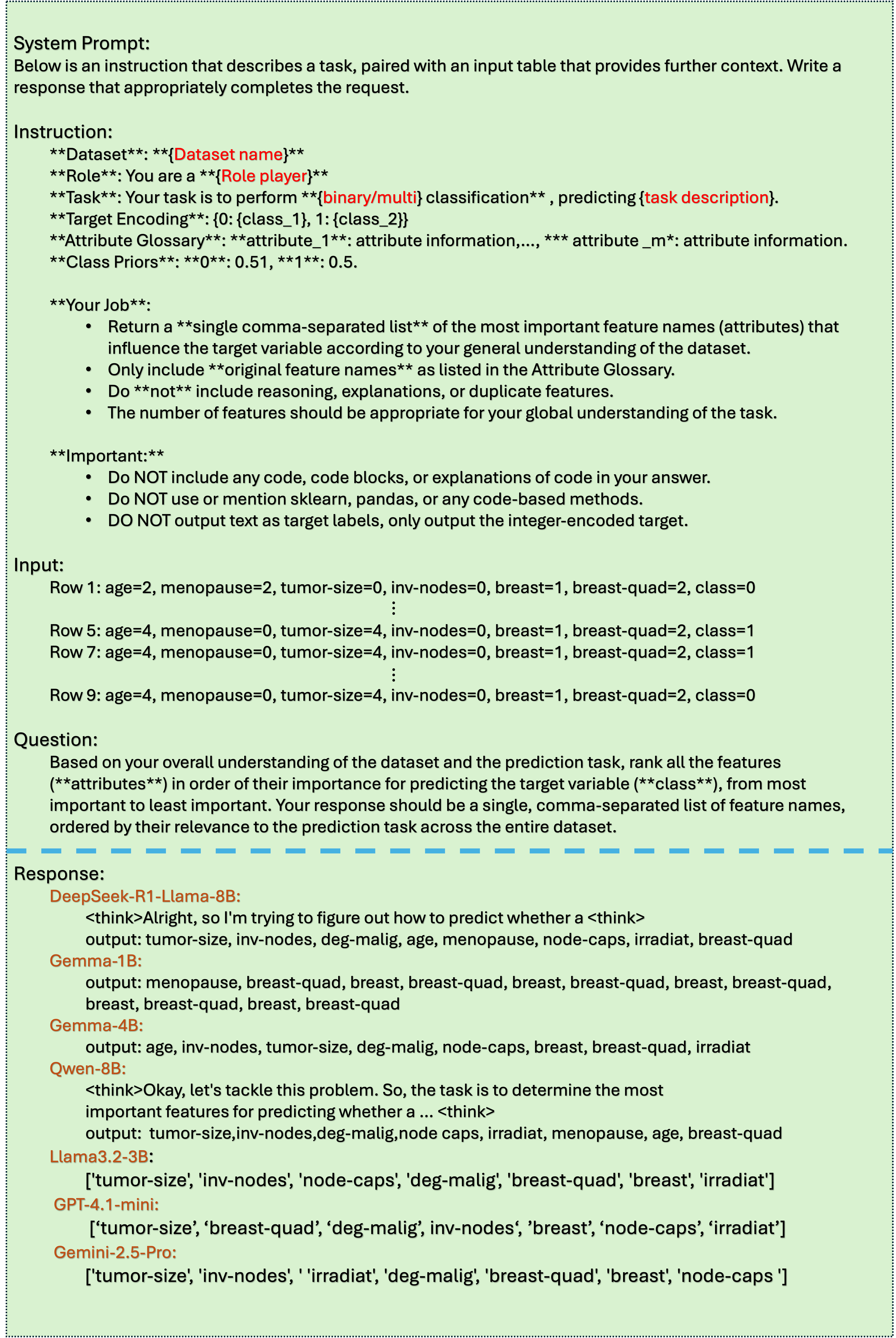}
    \caption{\textbf{Self-attribution prompt composition.} This setup is designed to produce a ranking of $m$ features, indicating which attributes it believes were most influential for its decision. The bottom panel presents model outputs for the same attribution prompt on the example dataset (Breast Cancer).}
    \label{fig:self-attribution}
\end{figure}

\begin{figure}
    \centering
    \includegraphics[width=.85\linewidth]{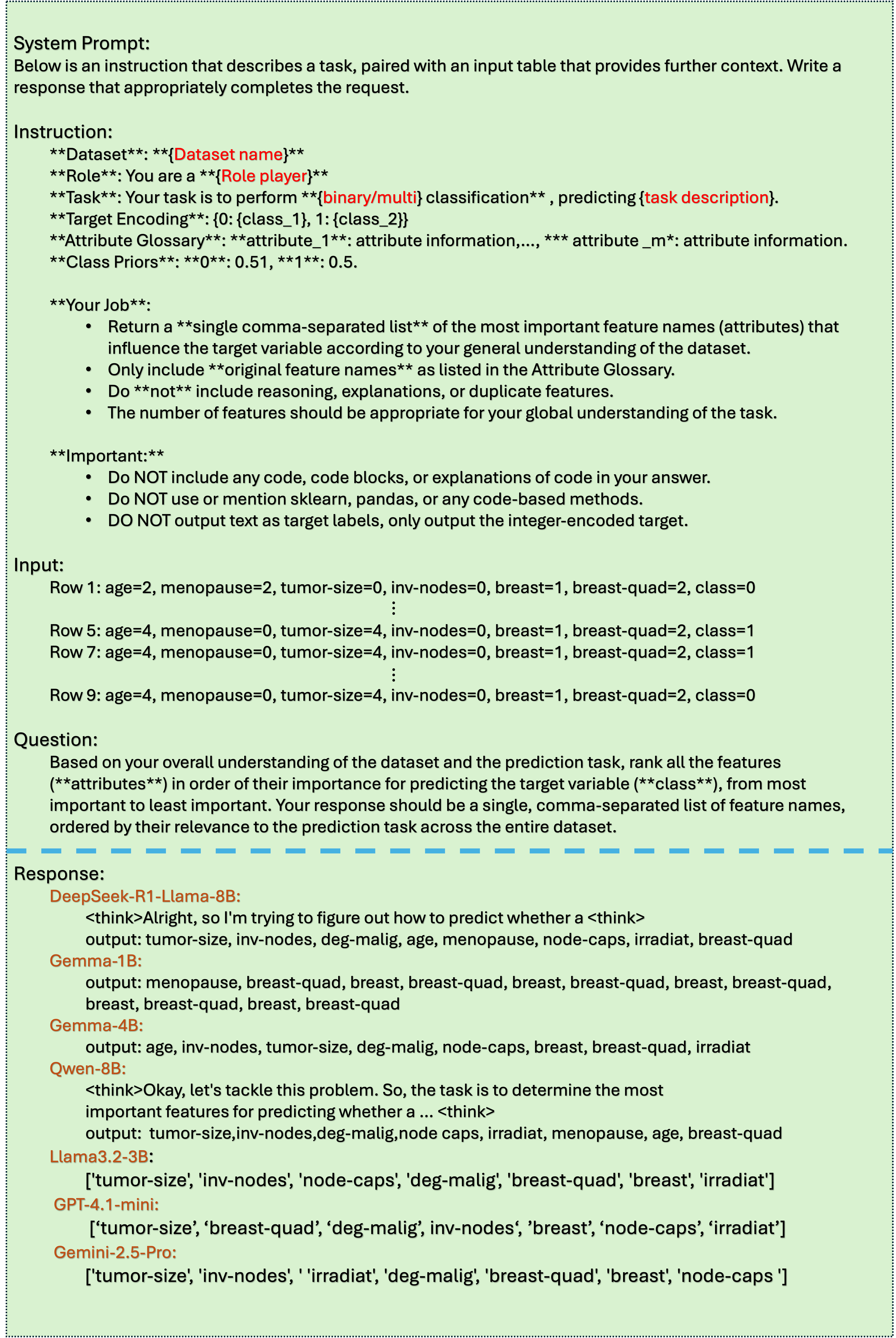}
    \caption{\textbf{LAO prompt composition.} This setup is designed to re-evaluate the model under identical prompting while ablating that feature from every row. The bottom panel presents model outputs for the same attribution prompt on the example dataset (Breast Cancer).}
    \label{fig:lao-prompt}
\end{figure}

\begin{figure}
    \centering
    \includegraphics[width=.85\linewidth]{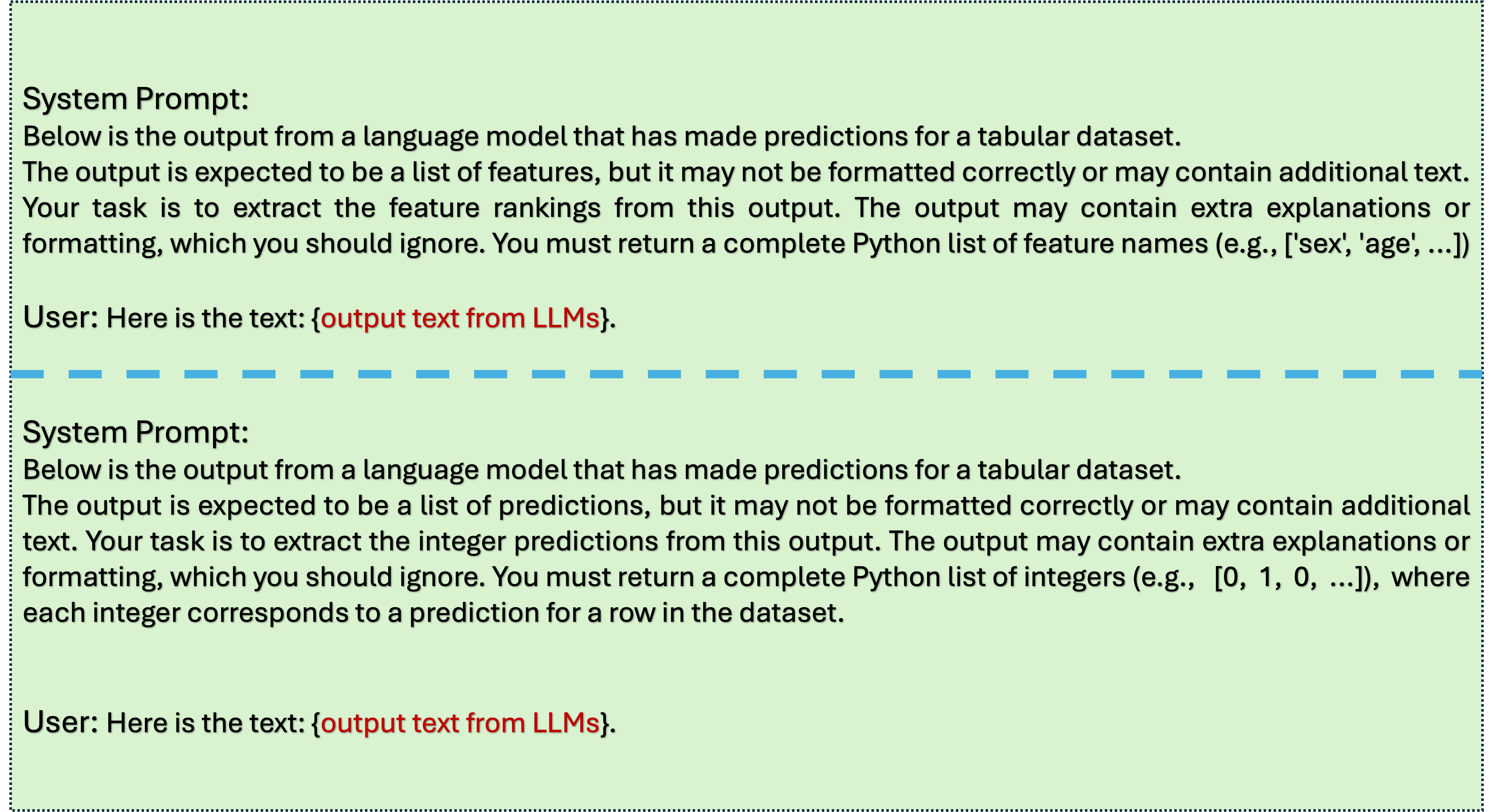}
    \caption{\textbf{Post-processing assistant prompts} for extracting structured outputs from LLM generations. \textit{Top panel}: Prompt for extracting a list of integer predictions from noisy or verbose LLM responses. \textit{Bottom panel}: Prompt for extracting ranked feature names from attribution outputs, isolating the comma-separated list from surrounding text. }
    \label{fig:prompt_post}
\end{figure}

We detail the full STaDS prompt template, decoding configuration, and evaluation setup to ensure reproducibility. Each prompt follows a deterministic structure composed of four blocks, as shown in Fig.~\ref{fig:prompt_zero} - \ref{fig:prompt_post}.

All open-source models (Llama-3, Mistral-7B, Gemma, Qwen3, DeepSeek-R1) were evaluated on an 8$\times$NVIDIA RTX 3090 cluster with temperature~=~0.2 and top-p~=~1.0. Commercial baselines (Gemini-2.5-Pro and GPT-4.1-mini) were accessed through their official APIs. Outputs were automatically cleaned using GPT-4-mini for consistent formatting. Official code will be published soon.

% ========= Extended Results ===========
\section{Extended Results}

\paragraph{Comprehension Fidelity \& Predictive Competence}
Table~\ref{tab:adult-income-detailed} -- \ref{tab:pima-detailed} list
detailed Acc, Macro‑F1, PenAcc, Len-F1, UnkLbl\%, and Set-Jacc for zero‑shot and few‑shot settings across all 15 datasets and 11 models. Table~\ref{tab:icu-pa} summarizes penalized accuracy of best performing models.

\begin{figure}[!t]
\centering
    \begin{subfigure}[t]{.5\textwidth}
        \centering
        \includegraphics[width=\textwidth]{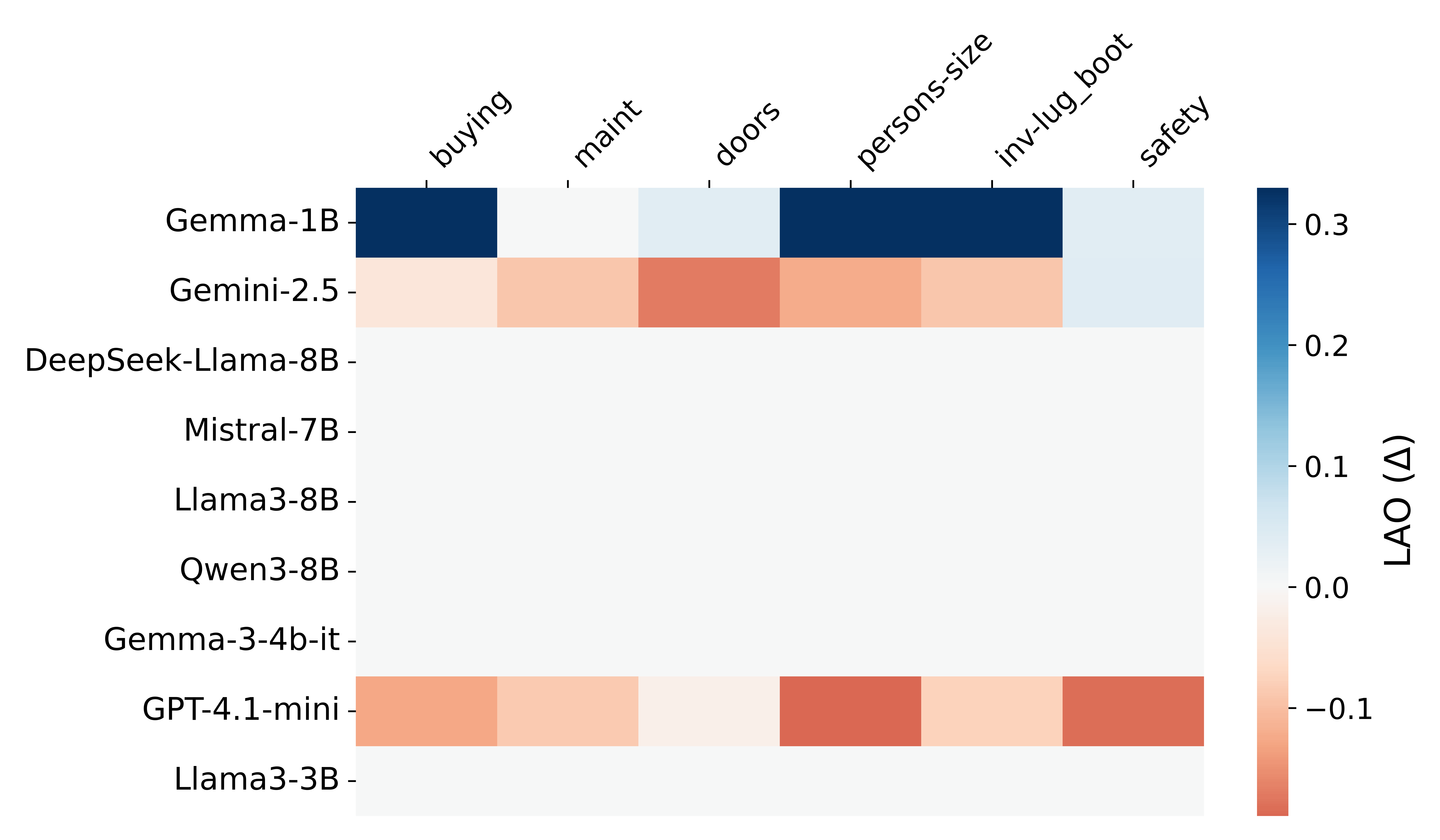}
    \end{subfigure}%
    \hfill
    \begin{subfigure}[t]{.5\textwidth}
        \centering
        \includegraphics[width=\textwidth]{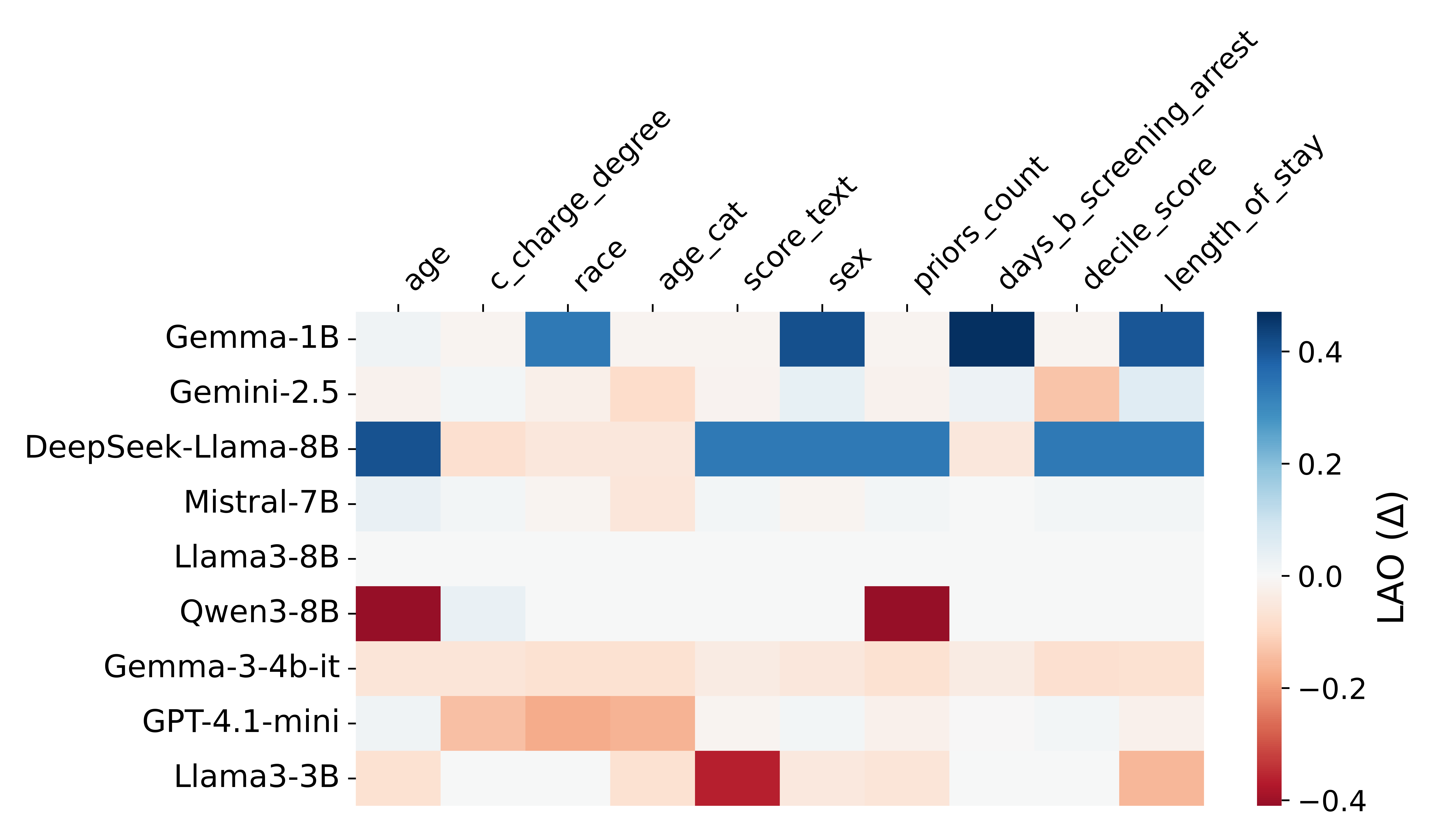}
    \end{subfigure}%
    \vfill
    \begin{subfigure}[t]{.5\textwidth}
        \centering
        \includegraphics[width=\textwidth]{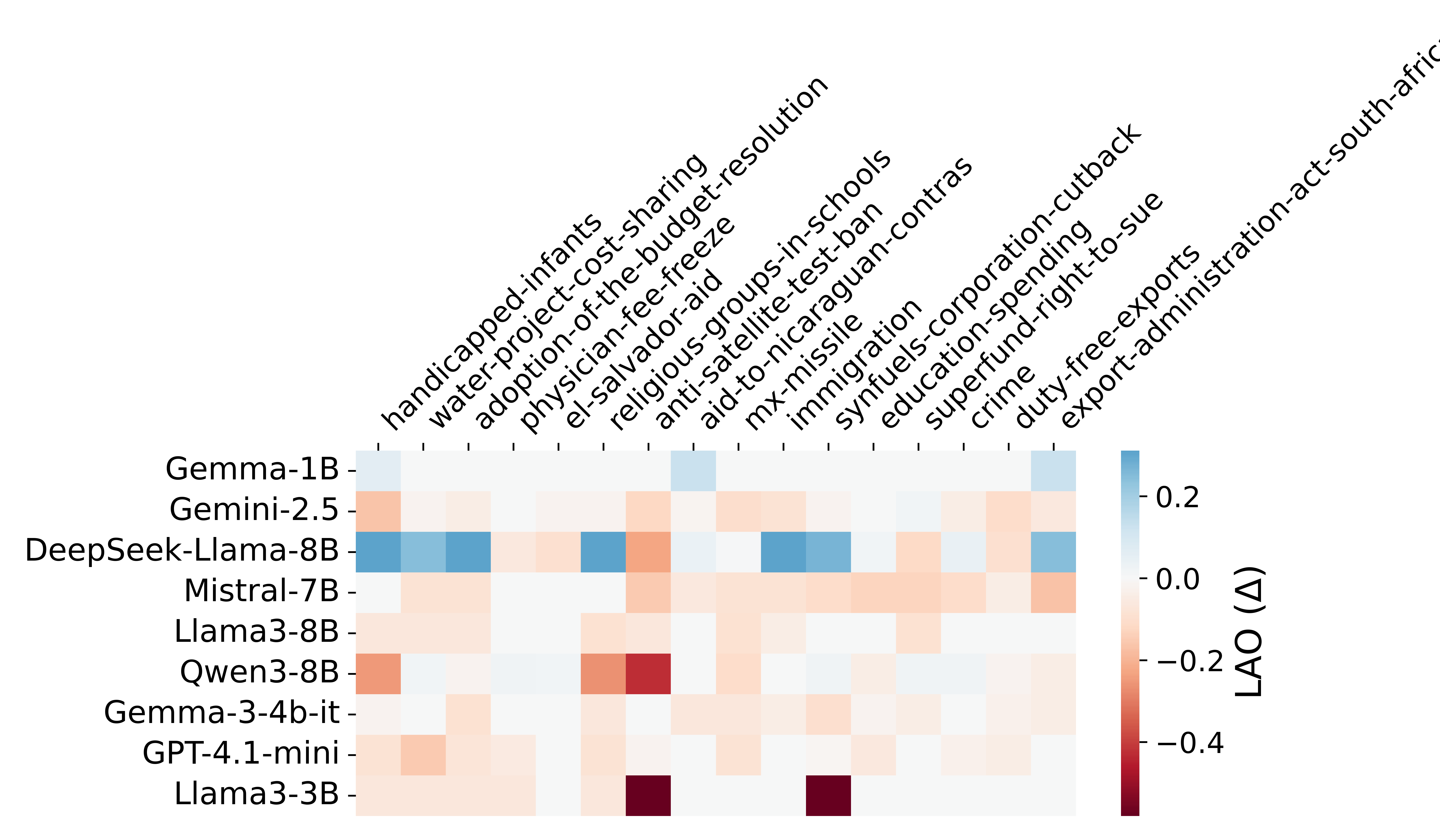}
    \end{subfigure}%
    \hfill
    \begin{subfigure}[t]{.5\textwidth}
        \centering
        \includegraphics[width=\textwidth]{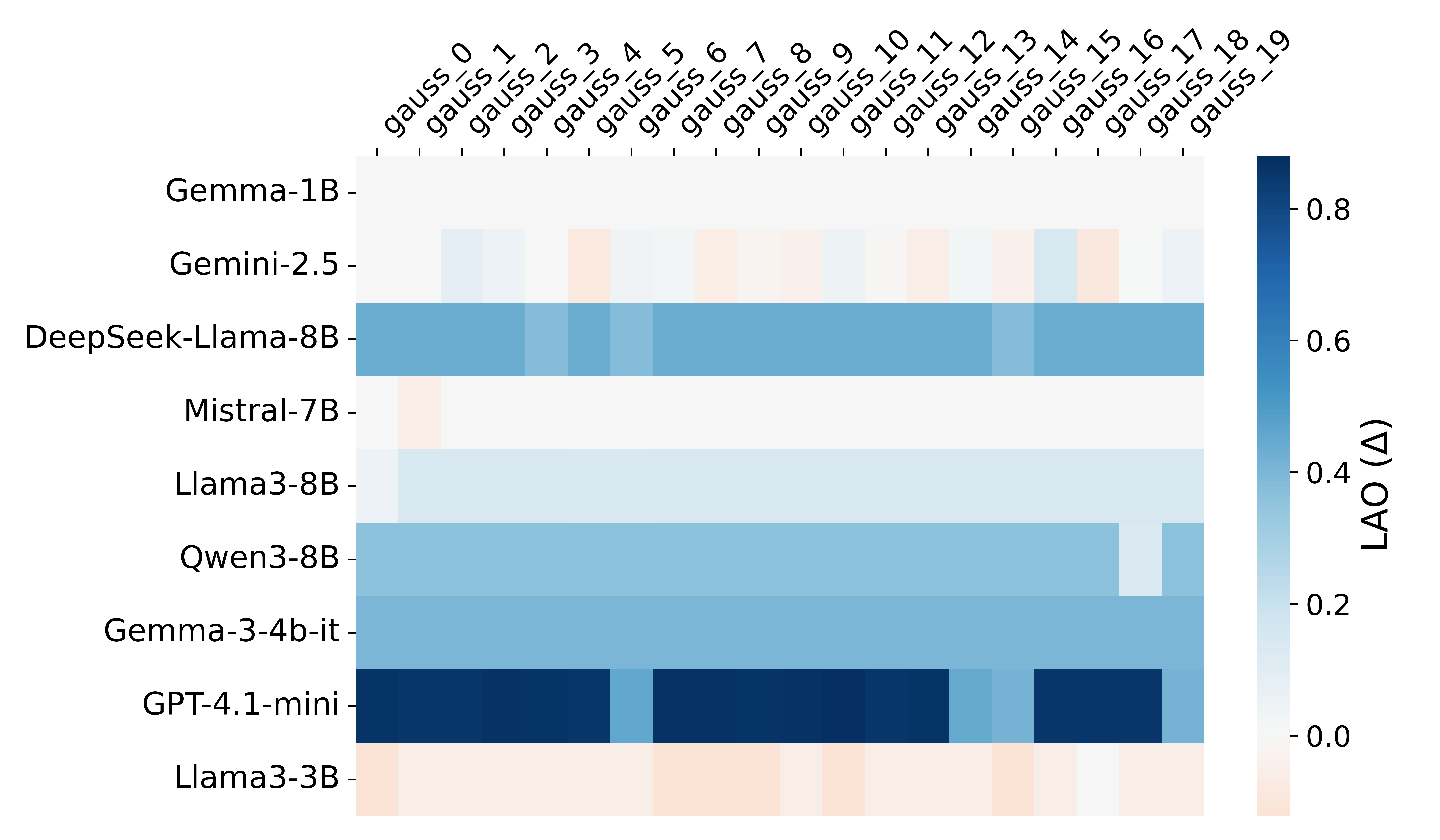}
    \end{subfigure}%
    \vfill
    \begin{subfigure}[t]{.5\textwidth}
        \centering
        \includegraphics[width=\textwidth]{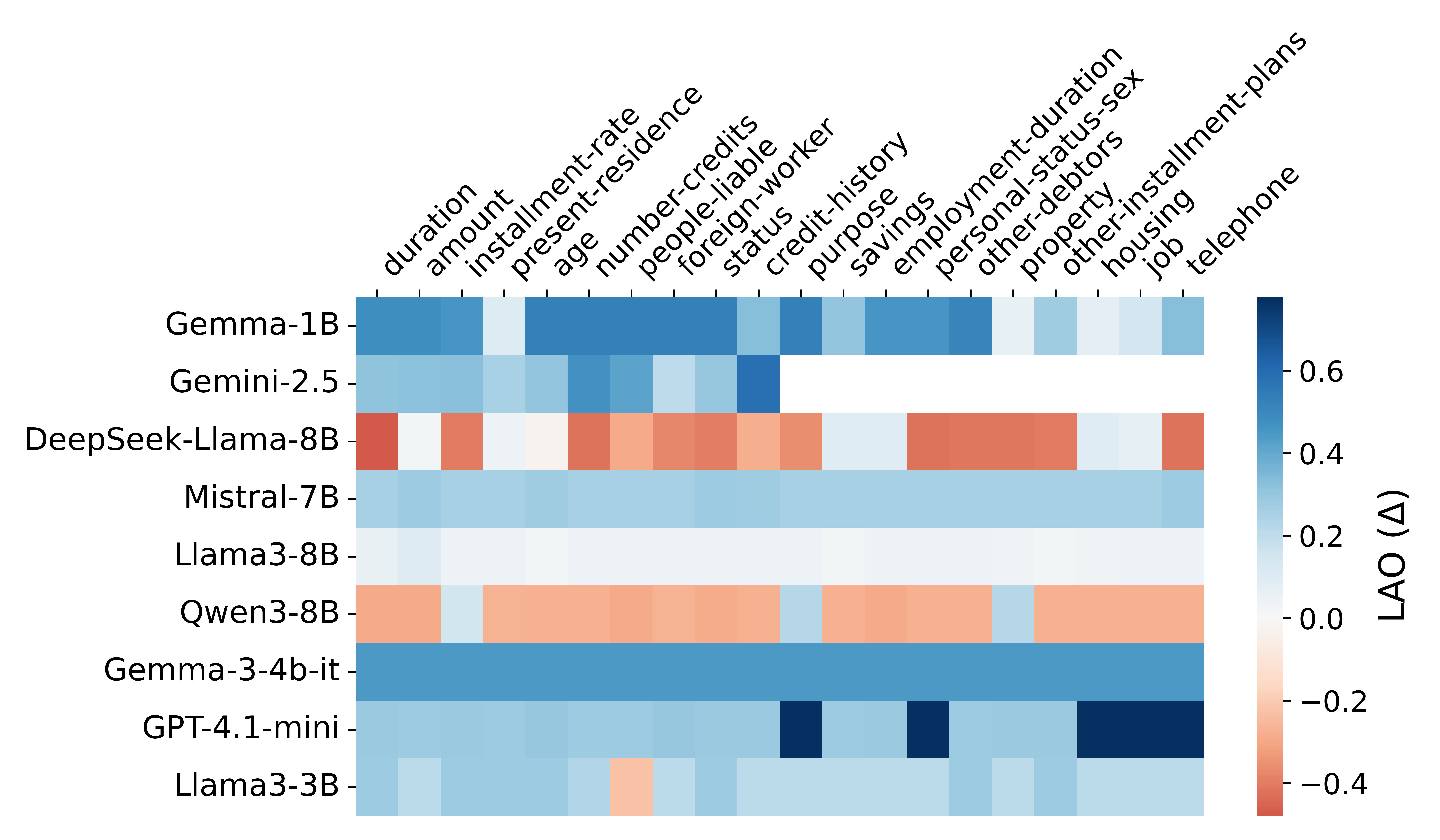}
    \end{subfigure}%
    \hfill
    \begin{subfigure}[t]{.5\textwidth}
        \centering
        \includegraphics[width=\textwidth]{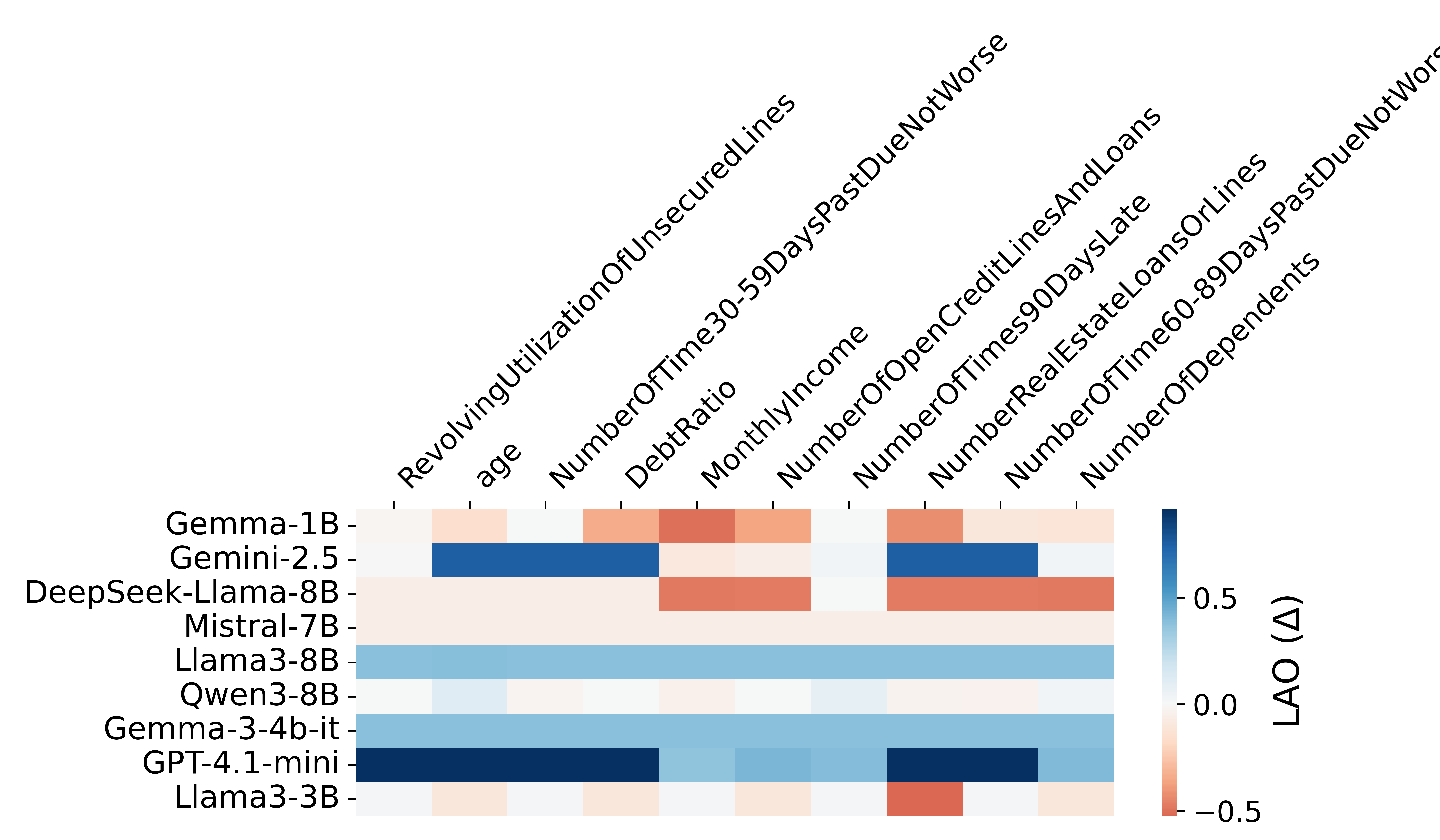}
    \end{subfigure}%
\caption{Heatmap of LAO performance ($\Delta_{\text{LAO}}$) for each feature (columns) and LLM (rows). Darker blue indicates a larger performance loss when the feature is removed (higher importance); red indicates a slight performance gain or negligible reliance.
}
\label{fig:lao_delta_app_1}
\end{figure}

\begin{figure}[!t]
\centering
    \begin{subfigure}[t]{.5\textwidth}
        \centering
        \includegraphics[width=\textwidth]{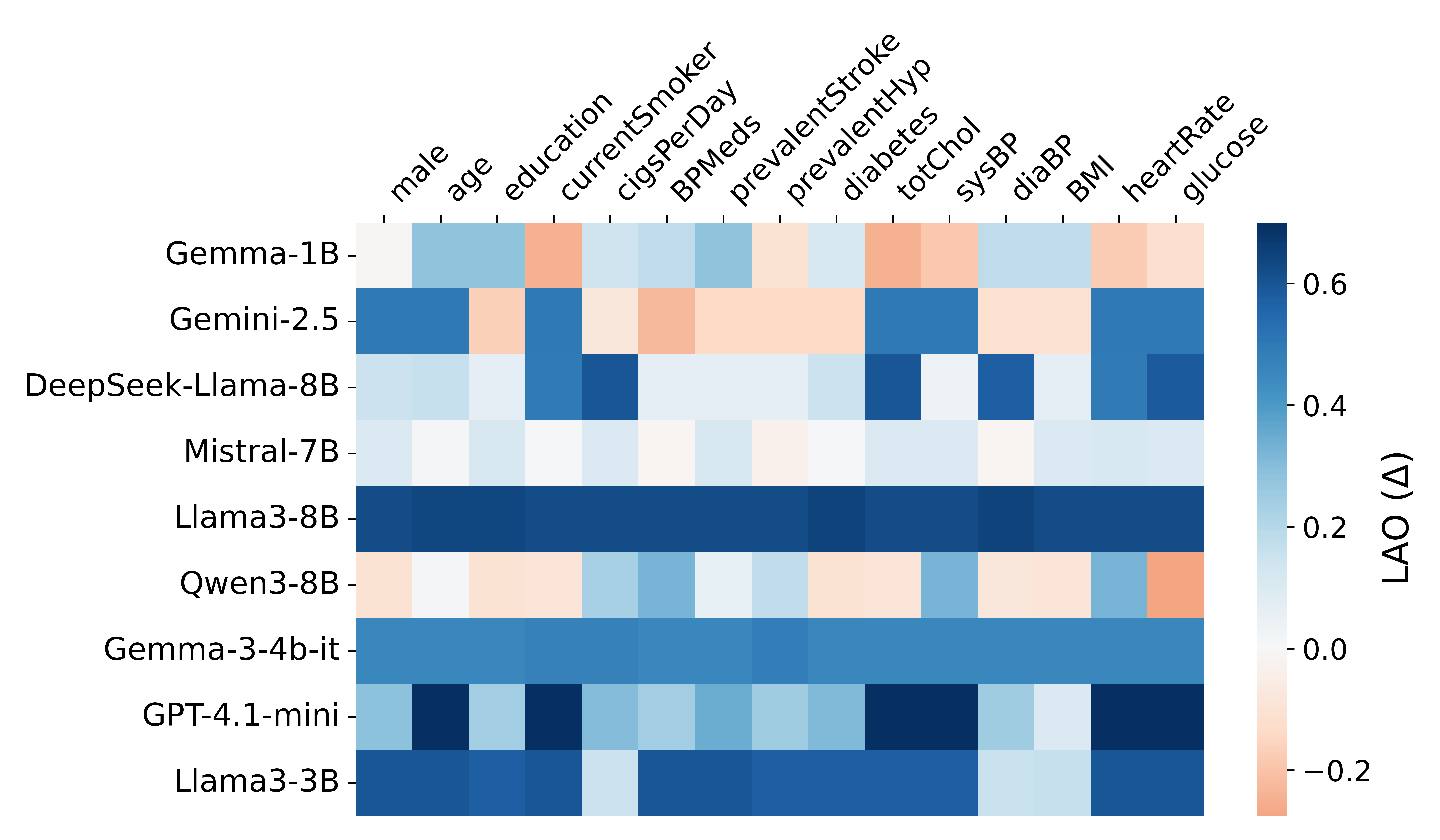}
    \end{subfigure}%
    \hfill
    \begin{subfigure}[t]{.5\textwidth}
        \centering
        \includegraphics[width=\textwidth]{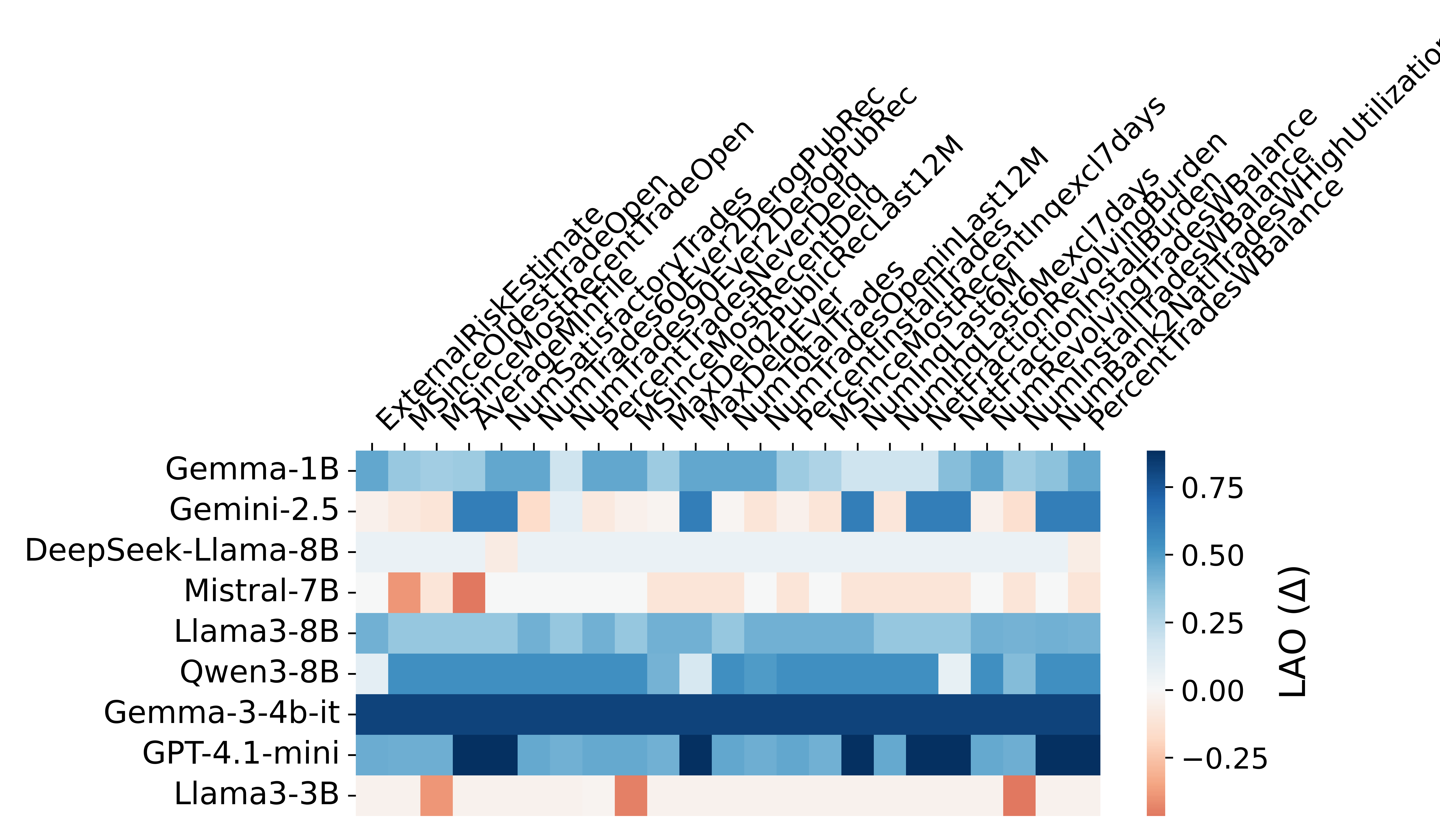}
    \end{subfigure}%
    \vfill
    \begin{subfigure}[t]{.5\textwidth}
        \centering
        \includegraphics[width=\textwidth]{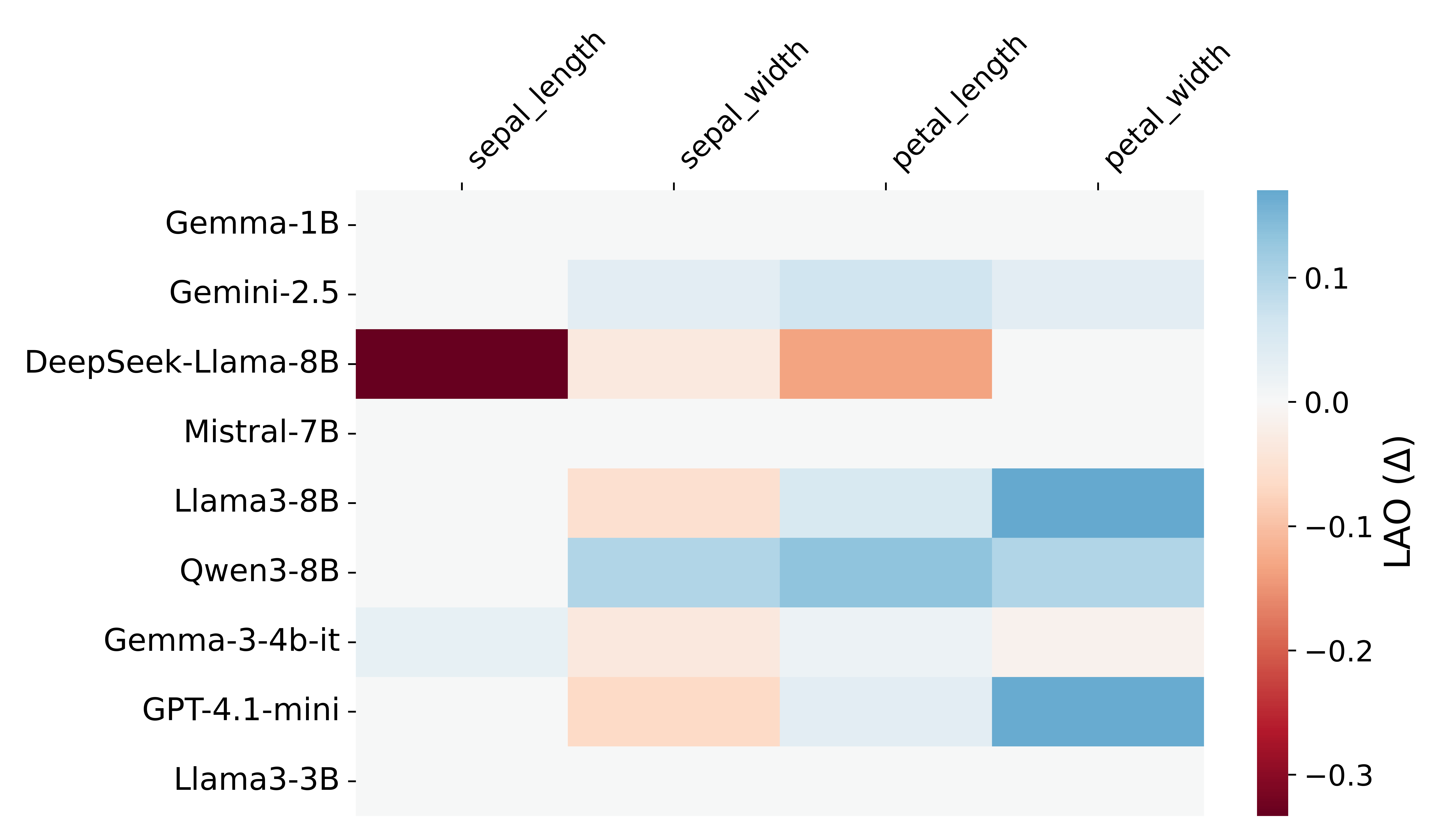}
    \end{subfigure}%
    \hfill
    \begin{subfigure}[t]{.5\textwidth}
        \centering
        \includegraphics[width=\textwidth]{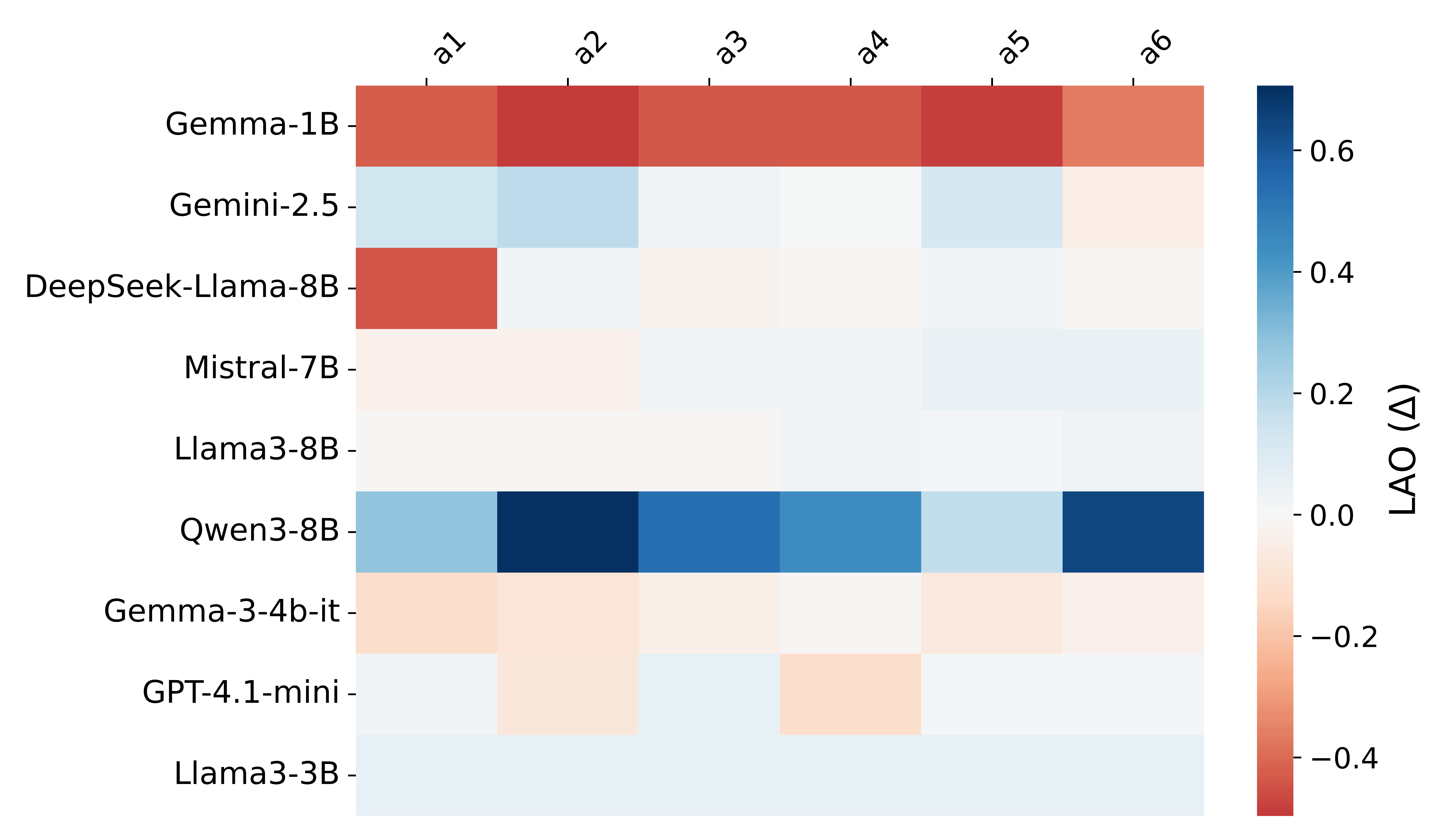}
    \end{subfigure}%
    \vfill
    \begin{subfigure}[t]{.5\textwidth}
        \centering
        \includegraphics[width=\textwidth]{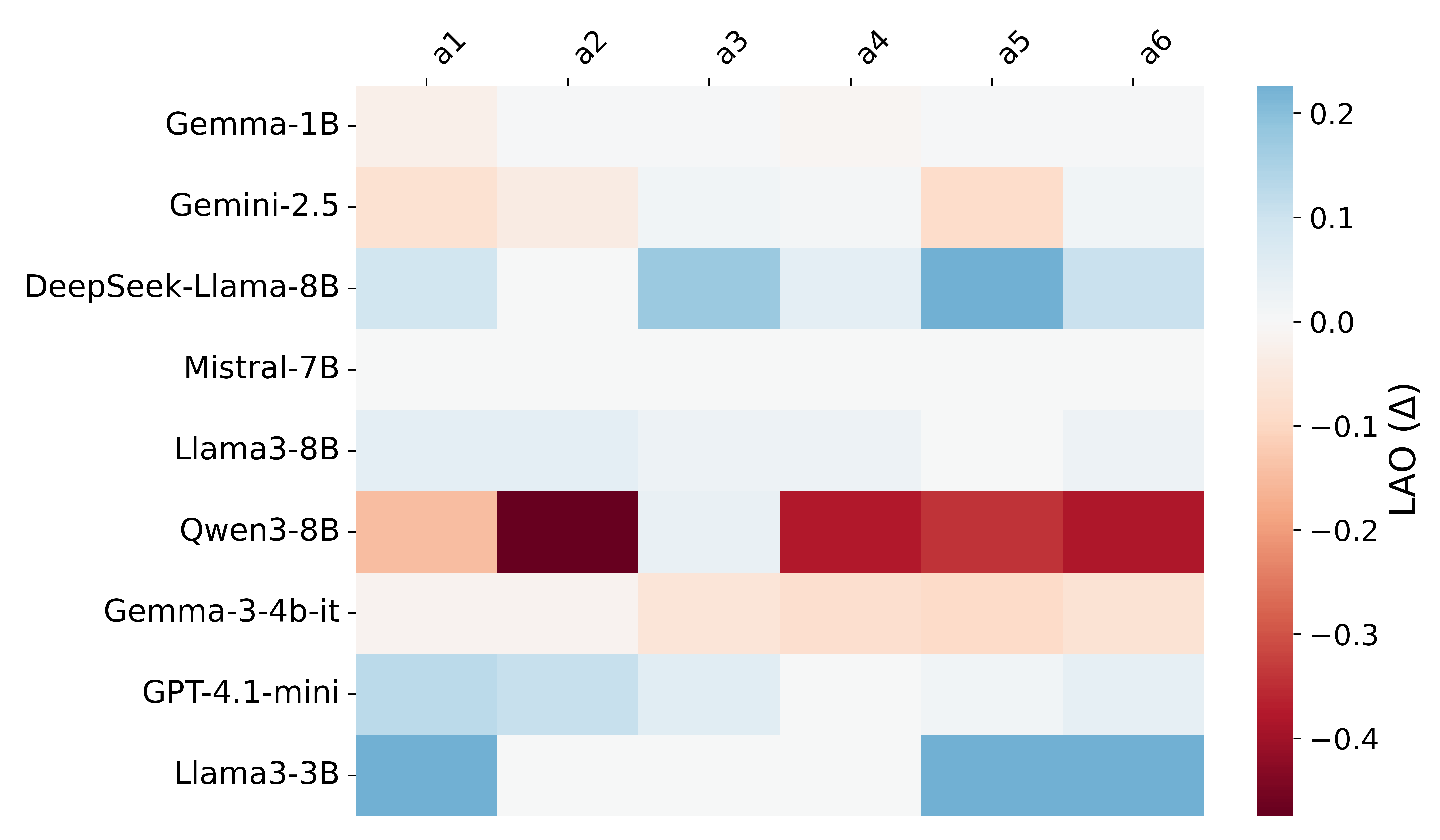}
    \end{subfigure}%
    \hfill
    \begin{subfigure}[t]{.5\textwidth}
        \centering
        \includegraphics[width=\textwidth]{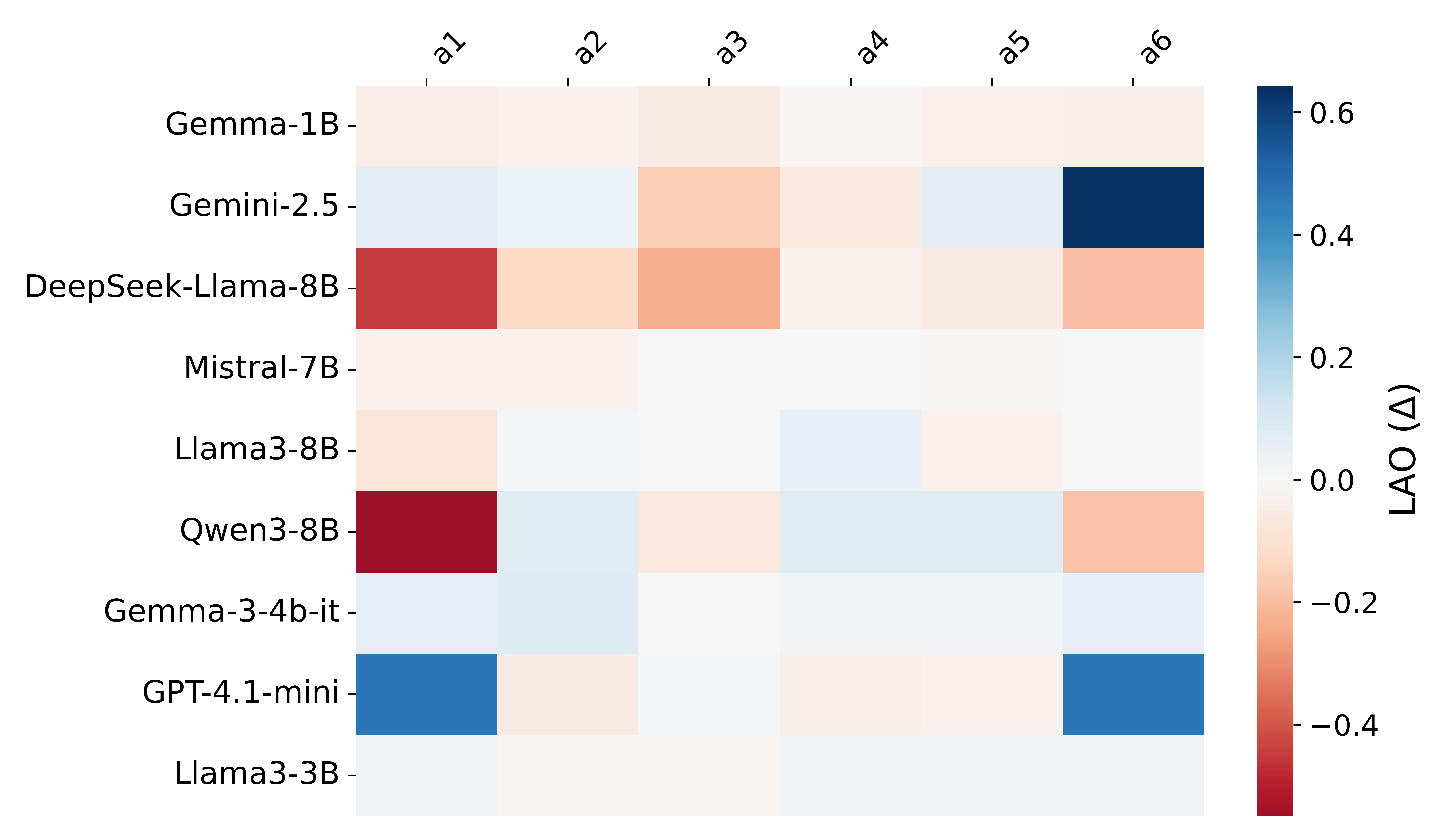}
    \end{subfigure}%
\caption{Heatmap of LAO performance ($\Delta_{\text{LAO}}$) for each feature (columns) and LLM (rows). Darker blue indicates a larger performance loss when the feature is removed (higher importance); red indicates a slight performance gain or negligible reliance.
}
\label{fig:lao_delta_app_2}
\end{figure}

\begin{figure}[!t]
\centering
    \begin{subfigure}[t]{.5\textwidth}
        \centering
        \includegraphics[width=\textwidth]{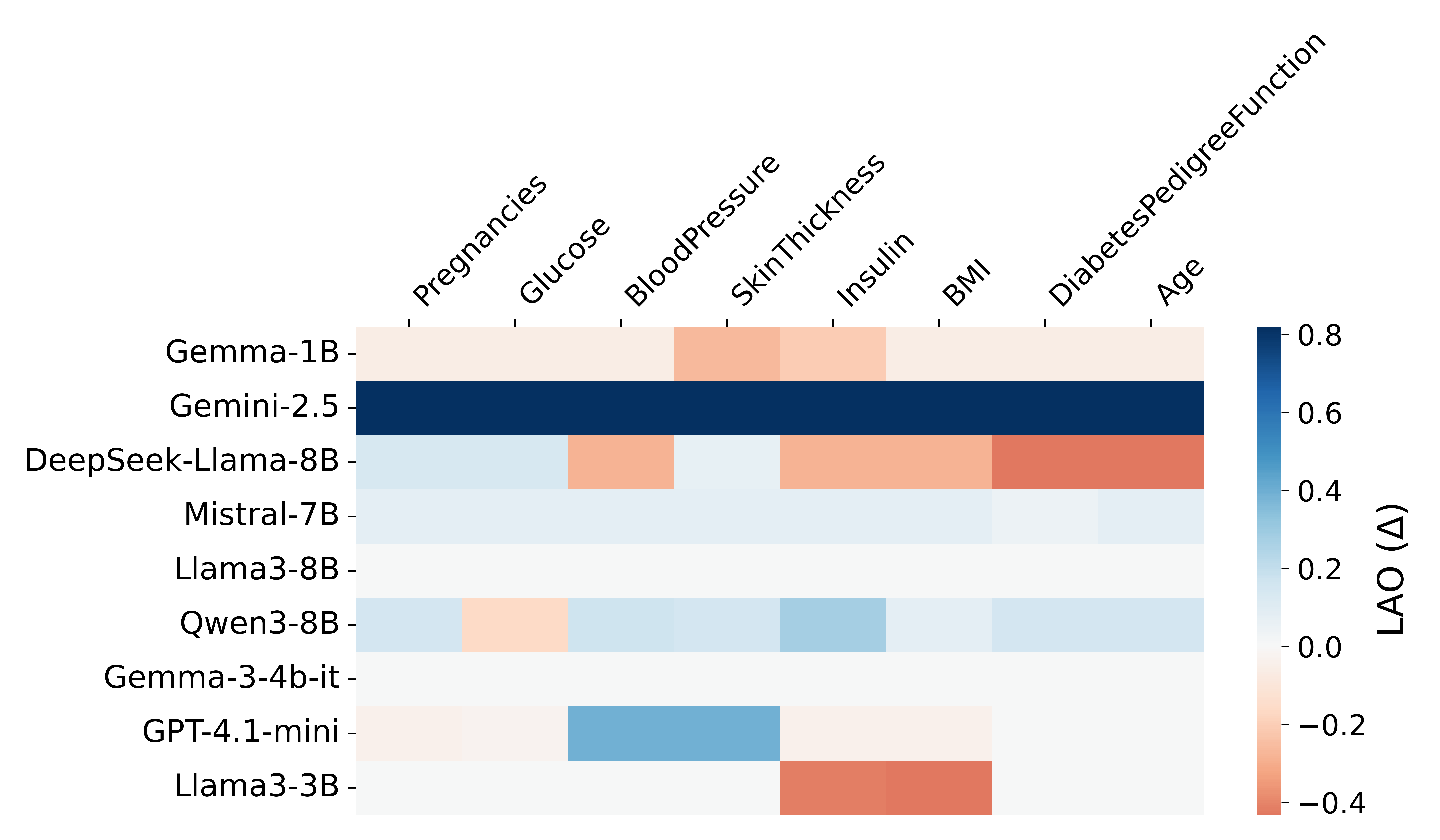}
    \end{subfigure}%
\caption{Heatmap of LAO performance ($\Delta_{\text{LAO}}$) for each feature (columns) of \texttt{Car Evaluation} and LLM (rows). Darker blue indicates a larger performance loss when the feature is removed (higher importance); red indicates a slight performance gain or negligible reliance.
}
\label{fig:pima_lao_delta}
\end{figure}

% \begin{table}[h]
% \centering
% \small
% \begin{tabular}{lccc}
% \toprule
% \textbf{Model} & \textbf{Params} & \textbf{Ctx.~Len.} & \textbf{Access}\\
% \midrule
% Llama3 8B Instruct           & 8B   &  8k & HF weights \\
% Llama3 3B                   & 3B   &  8k & HF weights \\
% Mistral-7B Instruct v0.3        & 7B   &  8k & HF weights \\
% DeepSeek-Llama-8B Distill-Llama 8B    & 8B   & 16k & HF weights \\
% Qwen3 8B                       & 8B   & 32k & HF weights \\
% Gemma-3 1B-it                   & 1B   &  4k & HF weights \\
% Gemma-3 4B-it                   & 4B   &  8k & HF weights \\
% Gemini-2.5 Pro                  & n/a   & 1 M & Google API \\
% GPT-4.1 Mini                    & n/a   & 128k& OpenAI API \\
% \bottomrule
% \end{tabular}
% \caption{Large language models evaluated.  
% ‘Params’ = reported parameter count;  
% ‘Ctx.~Len.’ = maximum context length tokens.}
% \label{tab:models}
% \end{table}

\begin{table}[t]
\small
\centering
\begin{tabular}{p{3cm}ccc}
\toprule
Model & $\sigma_{\text{LAO}}$ & \textsc{Self-Faith} & \textsc{SelfAtt@\,$k$} \\
\midrule
Gemma-1B & 0.17 & NaN & 0.00 \\
Gemini-2.5-Pro & 0.07 & 0.25 (0.38) & 1.00 \\
DeepSeek-Llama-8B & 0.24 & 0.24 (0.41) & 1.00 \\
Llama3-8B & 0.00 & $-0.05$ (0.89) & 0.73 \\
Qwen3-8B  & 0.11 & $-0.17$ (0.67) & 0.44 \\
GPT-4.1-mini & 0.01 & $-0.02$ (0.96) & 1.00 \\
Llama3-3B & 0.11 & $-0.34$ (0.24) & 1.00 \\
Mistral-7B & 0.01 & $-0.54$ (0.08) & 0.73 \\
\bottomrule
\end{tabular}
\caption{\label{tab:adult_au_metrics} Decision faithfulness  metrics between self‑attribution rank ($\pi_{\text{self}}$) and LAO-attribution rank ($\pi_{\text{LAO}}$); Adult Income Dataset ($m=14$). NaN indicates $\pi_{\text{self}}$ empty.}
\end{table}

%====================== Breast Cancer ======================
\begin{table}[t]
\small
\centering
\begin{tabular}{p{3cm}ccc}
\toprule
Model & $\sigma_{\text{LAO}}$ & \textsc{Self-Faith} & \textsc{SelfAtt@\,$k$} \\
\midrule
Gemma-1B      & 0.24 & $-1.00$ (0.00) & 0.33 \\
Gemini-2.5-Pro  & 0.04 & $-0.61$ (0.15) & 0.78 \\
DeepSeek-Llama-8B     & 0.23 & 0.81 (0.01)    & 0.89 \\
Llama3-8B    & 0.02 & $-0.18$ (0.64) & 1.00 \\
Qwen3-8B       & 0.16 & 0.33 (0.42)    & 0.89 \\
GPT-4.1-mini    & 0.04 & 0.22 (0.58)    & 1.00 \\
Llama3-3B    & 0.11 & $-0.93$ (0.00) & 0.78 \\
Mistral-7B      & 0.09 & 0.15 (0.70)    & 1.00 \\
\bottomrule
\end{tabular}
\caption{\label{tab:breast_metrics} Decision faithfulness  metrics between self‑attribution rank ($\pi_{\text{self}}$) and LAO-attribution rank ($\pi_{\text{LAO}}$); Breast Cancer Dataset ($m=14$). NaN indicates $\pi_{\text{self}}$ empty.}
\end{table}

%====================== Car Evaluation ======================
\begin{table}[t]
\small
\centering
\begin{tabular}{p{3cm}ccc}
\toprule
Model & $\sigma_{\text{LAO}}$ & \textsc{Self-Faith} & \textsc{SelfAtt@\,$k$} \\
\midrule
Gemma-1B      & 0.17 & $-0.60$ (0.21) & 1.00 \\
Gemini-2.5-Pro  & 0.07 & 0.66 (0.16)    & 1.00 \\
DeepSeek-Llama-8B     & 0.00 & 0.03 (0.96)    & 1.00 \\
Llama3-8B    & 0.00 & 0.31 (0.54)    & 1.00 \\
Qwen3-8B       & 0.00 & 0.54 (0.27)    & 1.00 \\
GPT-4.1-mini    & 0.07 & $-0.60$ (0.21) & 1.00 \\
Llama3-3B    & 0.00 & $-1.00$ (NaN)  & 0.33 \\
\bottomrule
\end{tabular}
\caption{\label{tab:car_metrics}
 Decision faithfulness  metrics between self‑attribution rank ($\pi_{\text{self}}$) and LAO-attribution rank ($\pi_{\text{LAO}}$); Car Evaluation Dataset ($m=14$). NaN indicates $\pi_{\text{self}}$ empty.}
\end{table}

%====================== COMPAS ======================
\begin{table}[t]
\small
\centering
\begin{tabular}{p{3cm}ccc}
\toprule
Model & $\sigma_{\text{LAO}}$ & \textsc{Self-Faith} & \textsc{SelfAtt@\,$k$} \\
\midrule
Gemini-2.5-Pro  & 0.06 & $-0.58$ (0.08) & 1.00 \\
Gemma-1B      & 0.21 & $-0.45$ (0.26) & 0.80 \\
DeepSeek-Llama-8B     & 0.21 & 0.03 (0.94)    & 1.00 \\
Llama3-8B    & 0.00 & 0.03 (0.93)    & 0.90 \\
Qwen3-8B       & 0.18 & $-0.45$ (0.19) & 1.00 \\
GPT-4.1-mini    & 0.08 & 0.21 (0.56)    & 1.00 \\
Llama3-3B    & 0.11 & $-0.09$ (0.87) & 0.60 \\
\bottomrule
\end{tabular}
\caption{\label{tab:compas_metrics} Decision faithfulness  metrics between self‑attribution rank ($\pi_{\text{self}}$) and LAO-attribution rank ($\pi_{\text{LAO}}$); COMPAS Dataset ($m=14$). NaN indicates $\pi_{\text{self}}$ empty.}
\end{table}

%====================== Congressional Voting ======================
\begin{table}[t]
\small
\centering
\begin{tabular}{p{3cm}ccc}
\toprule
Model & $\sigma_{\text{LAO}}$ & \textsc{Self-Faith} & \textsc{SelfAtt@\,$k$} \\
\midrule
Gemma-1B      & 0.04 & 0.64 (0.12)    & 0.44 \\
Gemini-2.5-Pro  & 0.05 & 0.44 (0.09)    & 1.00 \\
DeepSeek-Llama-8B     & 0.19 & 0.25 (0.39)    & 0.88 \\
Llama3-8B    & 0.04 & $-0.24$ (0.38) & 0.94 \\
Qwen3-8B       & 0.13 & 0.01 (0.96)    & 0.94 \\
GPT-4.1-mini    & 0.04 & 0.23 (0.40)    & 1.00 \\
Llama3-3B    & 0.19 & $-0.38$ (0.15) & 1.00 \\
\bottomrule
\end{tabular}
\caption{\label{tab:congress_metrics} Decision faithfulness  metrics between self‑attribution rank ($\pi_{\text{self}}$) and LAO-attribution rank ($\pi_{\text{LAO}}$); Congression Voting Record Dataset ($m=14$). NaN indicates $\pi_{\text{self}}$ empty.}
\end{table}

%====================== Iris ======================
\begin{table}[t]
\small
\centering
\begin{tabular}{p{3cm}ccc}
\toprule
Model & $\sigma_{\text{LAO}}$ & \textsc{Self-Faith} & \textsc{SelfAtt@\,$k$} \\
\midrule
Gemma-1B      & 0.00 & 0.60 (0.40)   & 1.00 \\
Gemini-2.5-Pro  & 0.03 & 0.80 (0.20)   & 1.00 \\
DeepSeek-Llama-8B     & 0.15 & $-0.80$ (0.20)& 1.00 \\
Llama3-8B    & 0.10 & 0.20 (0.80)   & 1.00 \\
Qwen3-8B       & 0.06 & 0.40 (0.60)   & 1.00 \\
GPT-4.1-mini    & 0.10 & 0.80 (0.20)   & 1.00 \\
Llama3-3B    & 0.00 & $-0.60$ (0.40)& 1.00 \\
\bottomrule
\end{tabular}
\caption{\label{tab:iris_metrics} Decision faithfulness  metrics between self‑attribution rank ($\pi_{\text{self}}$) and LAO-attribution rank ($\pi_{\text{LAO}}$); Iris Dataset ($m=14$). NaN indicates $\pi_{\text{self}}$ empty.}
\end{table}

%====================== Monks-1 ======================
\begin{table}[t]
\small
\centering
\begin{tabular}{p{3cm}ccc}
\toprule
Model & $\sigma_{\text{LAO}}$ & \textsc{Self-Faith} & \textsc{SelfAtt@\,$k$} \\
\midrule
Gemma-1B      & 0.05 & $-0.20$ (0.70) & 1.00 \\
Gemini-2.5-Pro  & 0.09 & 0.94 (0.00)    & 1.00 \\
DeepSeek-Llama-8B     & 0.18 & 0.20 (0.70)    & 1.00 \\
Llama3-8B    & 0.02 & $-0.66$ (0.16) & 1.00 \\
Qwen3-8B       & 0.21 & 0.50 (0.67)    & 0.50 \\
GPT-4.1-mini    & 0.07 & 0.31 (0.54)    & 1.00 \\
Llama3-3B    & 0.00 & $-0.94$ (0.00) & 1.00 \\
\bottomrule
\end{tabular}
\caption{\label{tab:monk1_metrics} Decision faithfulness  metrics between self‑attribution rank ($\pi_{\text{self}}$) and LAO-attribution rank ($\pi_{\text{LAO}}$); Monk-1 Dataset ($m=14$). NaN indicates $\pi_{\text{self}}$ empty.}
\end{table}

\begin{table*}[t]
\centering
\small
\begin{tabular}{lrrrrrr}
\toprule
\multicolumn{7}{c}{\textbf{Adult Income} — Zero-shot}\\
\midrule
Model & Acc & Macro-F1 & PenAcc & Len-F1 & UnkLbl\% & Set-Jacc \\
\midrule
Gemma-1B                 & 0.010 & 0.020 & 0.000 & 0.198 & 81.8 & 0.182 \\
Gemma-4B                 & 0.490 & 0.428 & 0.194 & 0.408 & 0.0  & 1.000 \\
Llama3-8B   & 0.530 & 0.530 & 0.236 & 0.412 & 0.0  & 1.000 \\
Mistral-7B      & 0.410 & 0.377 & 0.402 & 0.985 & 0.0  & 1.000 \\
GPT-4.1-mini                     & 0.470 & 0.320 & 0.470 & 1.000 & 0.0  & 1.000 \\
Llama3-3B                   & 0.010 & 0.020 & 0.000 & 0.020 & 0.0  & 0.500 \\
Gemini-2.5-Pro                 & 0.700 & 0.688 & 0.700 & 1.000 & 0.0  & 1.000 \\
Qwen3-8B                        & 0.380 & 0.433 & 0.247 & 0.734 & 0.0  & 1.000 \\
DeepSeek-Llama-8B & 0.500 & 0.333 & 0.198 & 0.397 & 0.0  & 0.500 \\
\midrule
\multicolumn{7}{c}{\textbf{Adult Income} — Few-shot}\\
\midrule
Model & Acc & Macro-F1 & PenAcc & Len-F1 & UnkLbl\% & Set-Jacc \\
\midrule
Gemma-4B                 & 0.053 & 0.083 & 0.000 & 0.100 & 0.0 & 0.500 \\
Gemma-1B                 & 0.474 & 0.321 & 0.461 & 0.974 & 0.0 & 1.000 \\
Qwen3-8B                        & 0.579 & 0.367 & 0.579 & 1.000 & 0.0 & 0.500 \\
Llama3-3B                   & 0.579 & 0.367 & 0.104 & 0.049 & 0.0 & 0.500 \\
Gemini-2.5-Pro                 & 0.737 & 0.708 & 0.737 & 1.000 & 0.0 & 1.000 \\
Llama3-8B   & 0.579 & 0.367 & 0.407 & 0.655 & 0.0 & 0.500 \\
GPT-4.1-mini                     & 0.632 & 0.614 & 0.632 & 1.000 & 0.0 & 1.000 \\
DeepSeek-Llama-8B & 0.421 & 0.296 & 0.421 & 1.000 & 0.0 & 0.500 \\
Mistral-7B      & 0.053 & 0.083 & 0.000 & 0.100 & 0.0 & 0.500 \\
\bottomrule
\end{tabular}
\caption{Adult Income: STaDS metrics per model for zero-shot and few-shot. Acc = accuracy; PenAcc = penalised accuracy; Len-F1 = length F1; UnkLbl\% = unknown label rate (\%).}
\label{tab:adult-income-detailed}
\end{table*}
\begin{table*}[t]
\centering
\small
\begin{tabular}{lrrrrrr}
\toprule
\multicolumn{7}{c}{\textbf{Breast} — Zero-shot}\\
\midrule
Model & Acc & Macro-F1 & PenAcc & Len-F1 & UnkLbl\% & Set-Jacc \\
\midrule
DeepSeek-Llama-8B     & 0.697 & 0.433 & 0.365 & 0.337 & 0.0 & 1.000 \\
Gemini-2.5-Pro                    & 0.650 & 0.575 & 0.646 & 0.993 & 0.0 & 1.000 \\
Gemma-1B                    & 0.708 & 0.414 & 0.376 & 0.337 & 0.0 & 0.500 \\
Gemma-4B                    & 0.000 & 0.000 & 0.000 & 0.007 & 1.0 & 0.000 \\
Mistral-7B         & 0.690 & 0.408 & 0.358 & 0.337 & 0.0 & 1.000 \\
GPT-4.1-mini                        & 0.729 & 0.524 & 0.348 & 0.238 & 0.0 & 1.000 \\
Llama3-8B      & 0.704 & 0.413 & 0.373 & 0.337 & 0.0 & 1.000 \\
Qwen3-8B                           & 0.224 & 0.296 & 0.000 & 0.478 & 0.0 & 1.000 \\
Llama3-3B                      & 0.708 & 0.414 & 0.376 & 0.337 & 0.0 & 0.500 \\

\midrule
\multicolumn{7}{c}{\textbf{Breast} — Few-shot}\\
\midrule

Model & Acc & Macro-F1 & PenAcc & Len-F1 & UnkLbl\% & Set-Jacc \\
\midrule
Mistral-7B         & 0.036 & 0.082 & 0.000 & 0.133 & 0.0 & 1.000 \\
GPT-4.1-mini                        & 0.732 & 0.525 & 0.732 & 1.000 & 0.0 & 1.000 \\
Qwen3-8B                           & 0.679 & 0.491 & 0.218 & 0.079 & 0.0 & 1.000 \\
Gemini-2.5-Pro                    & 0.589 & 0.548 & 0.589 & 1.000 & 0.0 & 1.000 \\
Llama3-8B      & 0.732 & 0.525 & 0.272 & 0.079 & 0.0 & 1.000 \\
DeepSeek-Llama-8B    & 0.554 & 0.495 & 0.093 & 0.079 & 0.0 & 1.000 \\
Gemma-4B                    & 0.714 & 0.470 & 0.254 & 0.079 & 0.0 & 1.000 \\
Gemma-1B                    & 0.714 & 0.417 & 0.714 & 1.000 & 0.0 & 0.500 \\
Llama3-3B                      & 0.107 & 0.120 & 0.000 & 0.303 & 0.0 & 0.500 \\

\bottomrule
\end{tabular}
\caption{Breast: STaDS metrics per model for zero-shot and few-shot. Acc = accuracy; PenAcc = penalised accuracy; Len-F1 = length F1; UnkLbl\% = unknown label rate (\%).}
\label{tab:breast-detailed}
\end{table*}

\begin{table*}[t]
\centering
\small
\begin{tabular}{lrrrrrr}
\toprule
\multicolumn{7}{c}{\textbf{Car evaluation} — Zero-shot}\\
\midrule
Model & Acc & Macro-F1 & PenAcc & Len-F1 & UnkLbl\% & Set-Jacc \\
\midrule
Gemini-2.5-Pro                    & 0.419 & 0.243 & 0.406 & 0.973 & 0.0 & 0.500 \\
Gemma-4B                    & 0.326 & 0.123 & 0.045 & 0.439 & 0.0 & 0.250 \\
Gemma-1B                    & 0.005 & 0.008 & 0.000 & 0.036 & 0.0 & 1.000 \\
GPT-4.1-mini                        & 0.326 & 0.129 & 0.000 & 0.316 & 0.0 & 0.750 \\
DeepSeek-Llama-8B    & 0.263 & 0.219 & 0.000 & 0.439 & 0.0 & 1.000 \\
Mistral-7B         & 0.279 & 0.215 & 0.000 & 0.439 & 0.0 & 0.750 \\
Llama3-3B                      & 0.000 & 0.000 & 0.000 & 0.000 & 0.0 & 0.000 \\
Qwen3-8B                           & 0.021 & 0.030 & 0.000 & 0.410 & 0.0 & 0.500 \\
Llama3-8B      & 0.326 & 0.123 & 0.045 & 0.439 & 0.0 & 0.250 \\
\midrule
\multicolumn{7}{c}{\textbf{Car evaluation} — Few-shot}\\
\midrule
Model & Acc & Macro-F1 & PenAcc & Len-F1 & UnkLbl\% & Set-Jacc \\
\midrule
Mistral-7B         & 0.333 & 0.125 & 0.000 & 0.104 & 0.0 & 0.250 \\
Gemma-1B                    & 0.333 & 0.126 & 0.330 & 0.993 & 0.0 & 0.250 \\
Gemma-4B                    & 0.293 & 0.113 & 0.000 & 0.104 & 0.0 & 0.500 \\
DeepSeek-Llama-8B    & 0.333 & 0.125 & 0.000 & 0.104 & 0.0 & 0.250 \\
Llama3-8B      & 0.013 & 0.019 & 0.000 & 0.026 & 0.0 & 0.250 \\
Gemini-2.5-Pro                    & 0.600 & 0.616 & 0.586 & 0.973 & 0.0 & 1.000 \\
Qwen3-8B                           & 0.307 & 0.117 & 0.000 & 0.104 & 0.0 & 0.250 \\
Llama3-3B                      & 0.000 & 0.000 & 0.000 & 0.000 & 0.0 & 0.000 \\
GPT-4.1-mini                        & 0.173 & 0.079 & 0.160 & 0.974 & 0.0 & 0.750 \\
\bottomrule
\end{tabular}
\caption{Car evaluation: STaDS metrics per model for zero-shot and few-shot. Acc = accuracy; PenAcc = penalised accuracy; Len-F1 = length F1; UnkLbl\% = unknown label rate (\%).}
\label{tab:car-detailed}
\end{table*}

\begin{table*}[t]
\centering
\small
\begin{tabular}{lrrrrrr}
\toprule
\multicolumn{7}{c}{\textbf{COMPAS} — Zero-shot}\\
\midrule
Model & Acc & Macro-F1 & PenAcc & Len-F1 & UnkLbl\% & Set-Jacc \\
\midrule
Gemma-1B                    & 0.502 & 0.338 & 0.270 & 0.536 & 0.0 & 1.000 \\
Gemma-4B                    & 0.500 & 0.495 & 0.268 & 0.536 & 0.0 & 1.000 \\
Qwen3-8B                           & 0.490 & 0.329 & 0.258 & 0.536 & 0.0 & 1.000 \\
Gemini-2.5-Pro                    & 0.816 & 0.810 & 0.571 & 0.510 & 0.0 & 1.000 \\
GPT-4.1-mini                        & 0.468 & 0.319 & 0.164 & 0.392 & 0.0 & 1.000 \\
Llama3-8B      & 0.500 & 0.333 & 0.268 & 0.536 & 0.0 & 0.500 \\
Mistral-7B         & 0.006 & 0.012 & 0.000 & 0.016 & 0.0 & 1.000 \\
Llama3-3B                      & 0.500 & 0.333 & 0.268 & 0.536 & 0.0 & 0.500 \\
DeepSeek-Llama-8B    & 0.502 & 0.338 & 0.270 & 0.536 & 0.0 & 1.000 \\

\midrule
\multicolumn{7}{c}{\textbf{COMPAS} — Few-shot}\\
\midrule
Model & Acc & Macro-F1 & PenAcc & Len-F1 & UnkLbl\% & Set-Jacc \\
\midrule
Llama3-8B      & 0.511 & 0.338 & 0.072 & 0.121 & 0.0 & 0.500 \\
Mistral-7B         & 0.500 & 0.333 & 0.061 & 0.121 & 0.0 & 1.000 \\
Gemma-4B                    & 0.011 & 0.022 & 0.000 & 0.022 & 0.0 & 0.500 \\
GPT-4.1-mini                        & 0.500 & 0.355 & 0.476 & 0.952 & 0.0 & 1.000 \\
Gemma-1B                    & 0.500 & 0.333 & 0.471 & 0.941 & 0.0 & 1.000 \\
Qwen3-8B                           & 0.511 & 0.338 & 0.072 & 0.121 & 0.0 & 0.500 \\
Llama3-3B                      & 0.011 & 0.022 & 0.000 & 0.108 & 0.0 & 1.000 \\
Gemini-2.5-Pro                    & 0.716 & 0.714 & 0.697 & 0.962 & 0.0 & 1.000 \\
DeepSeek-Llama-8B    & 0.409 & 0.321 & 0.406 & 0.994 & 0.0 & 1.000 \\

\bottomrule
\end{tabular}
\caption{COMPAS: STaDS metrics per model for zero-shot and few-shot. Acc = accuracy; PenAcc = penalised accuracy; Len-F1 = length F1; UnkLbl\% = unknown label rate (\%).}
\label{tab:adult-detailed}
\end{table*}

\begin{table*}[t]
\centering
\small
\begin{tabular}{lrrrrrr}
\toprule
\multicolumn{7}{c}{\textbf{Congression Vote} — Zero-shot}\\
\midrule
Model & Acc & Macro-F1 & PenAcc & Len-F1 & UnkLbl\% & Set-Jacc \\
\midrule
Mistral-7B         & 0.526 & 0.524 & 0.171 & 0.291 & 0.0 & 1.000 \\
Gemini-2.5-Pro                    & 0.397 & 0.395 & 0.388 & 0.982 & 0.0 & 1.000 \\
Llama3-3B                      & 0.534 & 0.348 & 0.180 & 0.291 & 0.0 & 0.500 \\
Llama3-8B      & 0.478 & 0.478 & 0.124 & 0.291 & 0.0 & 1.000 \\
Qwen3-8B                           & 0.474 & 0.360 & 0.119 & 0.291 & 0.0 & 1.000 \\
GPT-4.1-mini                        & 0.409 & 0.401 & 0.011 & 0.204 & 0.0 & 1.000 \\
DeepSeek-Llama-8B    & 0.466 & 0.325 & 0.111 & 0.291 & 0.0 & 1.000 \\
Gemma-1B                    & 0.263 & 0.274 & 0.062 & 0.598 & 0.0 & 0.500 \\
Gemma-4B                    & 0.526 & 0.525 & 0.171 & 0.291 & 0.0 & 1.000 \\

\midrule
\multicolumn{7}{c}{\textbf{Congression Vote} — Few-shot}\\
\midrule
Model & Acc & Macro-F1 & PenAcc & Len-F1 & UnkLbl\% & Set-Jacc \\
\midrule
Llama3-3B                      & 0.064 & 0.086 & 0.000 & 0.351 & 0.0 & 0.500 \\
GPT-4.1-mini                        & 0.532 & 0.451 & 0.521 & 0.978 & 0.0 & 1.000 \\
Gemini-2.5-Pro                    & 0.638 & 0.636 & 0.638 & 1.000 & 0.0 & 1.000 \\
Qwen3-8B                           & 0.489 & 0.478 & 0.023 & 0.067 & 0.0 & 1.000 \\
DeepSeek-Llama-8B    & 0.489 & 0.390 & 0.311 & 0.644 & 0.0 & 1.000 \\
Mistral-7B         & 0.468 & 0.467 & 0.001 & 0.067 & 0.0 & 1.000 \\
Llama3-8B      & 0.404 & 0.402 & 0.000 & 0.067 & 0.0 & 1.000 \\
Gemma-4B                    & 0.447 & 0.447 & 0.000 & 0.067 & 0.0 & 1.000 \\
Gemma-1B                    & 0.532 & 0.347 & 0.532 & 1.000 & 0.0 & 0.500 \\

\bottomrule
\end{tabular}
\caption{Congression Vote: STaDS metrics per model for zero-shot and few-shot. Acc = accuracy; PenAcc = penalised accuracy; Len-F1 = length F1; UnkLbl\% = unknown label rate (\%).}
\label{tab:adult-detailed}
\end{table*}

\begin{table*}[t]
\centering
\small
\begin{tabular}{lrrrrrr}
\toprule
\multicolumn{7}{c}{\textbf{Synthetic Gaussian} — Zero-shot}\\
\midrule
Model & Acc & Macro-F1 & PenAcc & Len-F1 & UnkLbl\% & Set-Jacc \\
\midrule
Gemma-1B                    & 0.500 & 0.500 & 0.068 & 0.137 & 0.0 & 1.000 \\
Gemma-4B                    & 0.500 & 0.487 & 0.068 & 0.137 & 0.0 & 1.000 \\
GPT-4.1-mini                        & 0.390 & 0.281 & 0.390 & 1.000 & 0.0 & 1.000 \\
Qwen3-8B                           & 0.510 & 0.355 & 0.078 & 0.137 & 0.0 & 1.000 \\
Llama3-8B      & 0.500 & 0.500 & 0.068 & 0.137 & 0.0 & 1.000 \\
Gemini-2.5-Pro                    & 0.550 & 0.448 & 0.550 & 1.000 & 0.0 & 1.000 \\
DeepSeek-Llama-8B    & 0.350 & 0.398 & 0.288 & 0.876 & 0.0 & 1.000 \\
Mistral-7B         & 0.500 & 0.500 & 0.068 & 0.137 & 0.0 & 1.000 \\
Llama3-3B                      & 0.000 & 0.000 & 0.000 & 0.000 & 0.0 & 0.000 \\

\midrule
\multicolumn{7}{c}{\textbf{Synthetic Gaussian} — Few-shot}\\
\midrule
Model & Acc & Macro-F1 & PenAcc & Len-F1 & UnkLbl\% & Set-Jacc \\
\midrule
Llama3-8B      & 0.440 & 0.306 & 0.273 & 0.667 & 0.0 & 1.000 \\
Gemma-4B                    & 0.440 & 0.306 & 0.403 & 0.926 & 0.0 & 0.500 \\
Gemma-1B                    & 0.400 & 0.384 & 0.000 & 0.036 & 0.0 & 1.000 \\
Mistral-7B         & 0.000 & 0.000 & 0.000 & 0.958 & 95.7 & 0.042 \\
GPT-4.1-mini                        & 0.880 & 0.873 & 0.880 & 1.000 & 0.0 & 1.000 \\
Gemini-2.5-Pro                    & 0.680 & 0.603 & 0.680 & 1.000 & 0.0 & 1.000 \\
Qwen3-8B                           & 0.360 & 0.359 & 0.360 & 1.000 & 0.0 & 1.000 \\
Llama3-3B                      & 0.000 & 0.000 & 0.000 & 0.000 & 0.0 & 0.000 \\
DeepSeek-Llama-8B    & 0.440 & 0.306 & 0.440 & 1.000 & 0.0 & 0.500 \\

\bottomrule
\end{tabular}
\caption{Synthetic Gaussian: STaDS metrics per model for zero-shot and few-shot. Acc = accuracy; PenAcc = penalised accuracy; Len-F1 = length F1; UnkLbl\% = unknown label rate (\%).}
\label{tab:adult-detailed}
\end{table*}

\begin{table*}[t]
\centering
\small
\begin{tabular}{lrrrrrr}
\toprule
\multicolumn{7}{c}{\textbf{German Credit Risk} — Zero-shot}\\
\midrule
Model & Acc & Macro-F1 & PenAcc & Len-F1 & UnkLbl\% & Set-Jacc \\
\midrule
GPT-4.1-mini                   & 0.660 & 0.616 & 0.660 & 1.000 & 0.00 & 1.000 \\
Mistral-7B    & 0.550 & 0.436 & 0.118 & 0.137 & 0.00 & 1.000 \\
Llama3-8B & 0.500 & 0.500 & 0.068 & 0.137 & 0.00 & 1.000 \\
DeepSeek-Llama-8B & 0.490 & 0.329 & 0.058 & 0.137 & 0.00 & 1.000 \\
Gemma-1B               & 0.490 & 0.490 & 0.058 & 0.137 & 0.00 & 1.000 \\
Gemini-2.5-Pro               & 0.460 & 0.457 & 0.460 & 1.000 & 0.00 & 1.000 \\
Qwen3-8B                      & 0.020 & 0.038 & 0.000 & 0.387 & 0.00 & 1.000 \\
Llama3-3B                 & 0.000 & 0.000 & 0.000 & 0.000 & 0.00 & 0.000 \\
Gemma-4B               & 0.000 & 0.000 & 0.000 & 0.020 & 1.00 & 0.000 \\
\midrule
\multicolumn{7}{c}{\textbf{German Credit Risk} — Few-shot}\\
\midrule
Model & Acc & Macro-F1 & PenAcc & Len-F1 & UnkLbl\% & Set-Jacc \\
\midrule
Mistral-7B         & 0.333 & 0.325 & 0.333 & 1.000 & 0.0 & 1.000 \\
DeepSeek-Llama-8B    & 0.444 & 0.308 & 0.094 & 0.300 & 0.0 & 1.000 \\
Llama3-8B      & 0.222 & 0.222 & 0.096 & 0.947 & 20.0 & 0.333 \\
Llama3-3B                      & 0.778 & 0.438 & 0.284 & 0.013 & 0.0 & 0.500 \\
GPT-4.1-mini                        & 0.778 & 0.679 & 0.778 & 1.000 & 0.0 & 1.000 \\
Gemini-2.5-Pro                    & 0.889 & 0.862 & 0.889 & 1.000 & 0.0 & 1.000 \\
Gemma-4B                    & 0.444 & 0.444 & 0.444 & 1.000 & 0.0 & 1.000 \\
Qwen3-8B                           & 0.222 & 0.182 & 0.222 & 1.000 & 0.0 & 0.500 \\
Gemma-1B                    & 0.556 & 0.357 & 0.529 & 0.947 & 0.0 & 1.000 \\

\bottomrule
\end{tabular}
\caption{German Credit Risk: STaDS metrics per model for zero-shot and few-shot. Acc = accuracy; PenAcc = penalised accuracy; Len-F1 = length F1; UnkLbl\% = unknown label rate (\%).}
\label{tab:adult-detailed}
\end{table*}

\begin{table*}[t]
\centering
\small
\begin{tabular}{lrrrrrr}
\toprule
\multicolumn{7}{c}{\textbf{Give Me Some Credit} — Zero-shot}\\
\midrule
Model & Acc & Macro-F1 & PenAcc & Len-F1 & UnkLbl\% & Set-Jacc \\
\midrule
GPT-4.1-mini                        & 0.550 & 0.533 & 0.492 & 0.885 & 0.0 & 1.000 \\
Llama3-8B      & 0.470 & 0.320 & 0.038 & 0.137 & 0.0 & 1.000 \\
DeepSeek-Llama-8B    & 0.830 & 0.832 & 0.825 & 0.990 & 0.0 & 1.000 \\
Mistral-7B         & 0.480 & 0.324 & 0.048 & 0.137 & 0.0 & 1.000 \\
Llama3-3B                      & 0.500 & 0.333 & 0.068 & 0.137 & 0.0 & 0.500 \\
Gemini-2.5-Pro                    & 0.790 & 0.785 & 0.790 & 1.000 & 0.0 & 1.000 \\
Qwen3-8B                           & 0.570 & 0.646 & 0.511 & 0.883 & 0.0 & 1.000 \\
Gemma-4B                    & 0.440 & 0.436 & 0.303 & 0.726 & 0.0 & 1.000 \\
Gemma-1B                    & 0.480 & 0.327 & 0.477 & 0.995 & 0.0 & 1.000 \\

\midrule
\multicolumn{7}{c}{\textbf{Give Me Some Credit} — Few-shot}\\
\midrule
Model & Acc & Macro-F1 & PenAcc & Len-F1 & UnkLbl\% & Set-Jacc \\
\midrule
Qwen3-8B                           & 0.500 & 0.500 & 0.500 & 1.000 & 0.0 & 1.000 \\
Llama3-3B                      & 0.500 & 0.333 & 0.017 & 0.035 & 0.0 & 0.500 \\
DeepSeek-Llama-8B    & 0.542 & 0.420 & 0.059 & 0.035 & 0.0 & 1.000 \\
Gemini-2.5-Pro                    & 0.750 & 0.743 & 0.750 & 1.000 & 0.0 & 1.000 \\
Llama3-8B      & 0.500 & 0.333 & 0.500 & 1.000 & 0.0 & 0.500 \\
Mistral-7B         & 0.500 & 0.333 & 0.017 & 0.035 & 0.0 & 0.500 \\
GPT-4.1-mini                        & 0.917 & 0.916 & 0.917 & 1.000 & 0.0 & 1.000 \\
Gemma-1B                    & 0.417 & 0.294 & 0.000 & 0.035 & 0.0 & 1.000 \\
Gemma-4B                    & 0.500 & 0.438 & 0.387 & 0.774 & 0.0 & 1.000 \\

\bottomrule
\end{tabular}
\caption{Give Me Some Credit: STaDS metrics per model for zero-shot and few-shot. Acc = accuracy; PenAcc = penalised accuracy; Len-F1 = length F1; UnkLbl\% = unknown label rate (\%).}
\label{tab:adult-detailed}
\end{table*}

\begin{table*}[t]
\centering
\small
\begin{tabular}{lrrrrrr}
\toprule
\multicolumn{7}{c}{\textbf{Heart Disease} — Zero-shot}\\
\midrule
Model & Acc & Macro-F1 & PenAcc & Len-F1 & UnkLbl\% & Set-Jacc \\
\midrule
Qwen3-8B                           & 0.420 & 0.484 & 0.352 & 0.864 & 0.0 & 1.000 \\
GPT-4.1-mini                        & 0.640 & 0.614 & 0.640 & 1.000 & 0.0 & 1.000 \\
Llama3-8B      & 0.450 & 0.449 & 0.018 & 0.137 & 0.0 & 1.000 \\
Gemma-1B                    & 0.510 & 0.510 & 0.078 & 0.137 & 0.0 & 1.000 \\
Gemma-4B                    & 0.480 & 0.448 & 0.048 & 0.137 & 0.0 & 1.000 \\
Mistral-7B         & 0.500 & 0.487 & 0.068 & 0.137 & 0.0 & 1.000 \\
DeepSeek-Llama-8B    & 0.510 & 0.372 & 0.507 & 0.995 & 0.0 & 1.000 \\
Llama3-3B                      & 0.500 & 0.333 & 0.068 & 0.137 & 0.0 & 0.500 \\
Gemini-2.5-Pro                    & 0.550 & 0.544 & 0.550 & 1.000 & 0.0 & 1.000 \\

\midrule
\multicolumn{7}{c}{\textbf{Heart Disease} — Few-shot}\\
\midrule
Model & Acc & Macro-F1 & PenAcc & Len-F1 & UnkLbl\% & Set-Jacc \\
\midrule
Gemma-4B                    & 0.650 & 0.642 & 0.483 & 0.667 & 0.0 & 1.000 \\
Gemma-1B                    & 0.450 & 0.429 & 0.283 & 0.667 & 0.0 & 1.000 \\
GPT-4.1-mini                        & 0.700 & 0.697 & 0.700 & 1.000 & 0.0 & 1.000 \\
DeepSeek-Llama-8B    & 0.600 & 0.375 & 0.600 & 1.000 & 0.0 & 0.500 \\
Llama3-8B      & 0.650 & 0.642 & 0.650 & 1.000 & 0.0 & 1.000 \\
Llama3-3B                      & 0.600 & 0.375 & 0.600 & 1.000 & 0.0 & 0.500 \\
Gemini-2.5-Pro                    & 0.500 & 0.495 & 0.500 & 1.000 & 0.0 & 1.000 \\
Qwen3-8B                           & 0.350 & 0.307 & 0.350 & 1.000 & 0.0 & 1.000 \\
Mistral-7B         & 0.600 & 0.375 & 0.114 & 0.029 & 0.0 & 0.500 \\

\bottomrule
\end{tabular}
\caption{Heart Disease: STaDS metrics per model for zero-shot and few-shot. Acc = accuracy; PenAcc = penalised accuracy; Len-F1 = length F1; UnkLbl\% = unknown label rate (\%).}
\label{tab:adult-detailed}
\end{table*}

\begin{table*}[t]
\centering
\small
\begin{tabular}{lrrrrrr}
\toprule
\multicolumn{7}{c}{\textbf{HELOC} — Zero-shot}\\
\midrule
Model & Acc & Macro-F1 & PenAcc & Len-F1 & UnkLbl\% & Set-Jacc \\
\midrule
Llama3-3B                  & 0.000 & 0.000 & 0.000 & 0.000 & 0.0 & 0.000 \\
Llama3-8B  & 0.500 & 0.333 & 0.068 & 0.137 & 0.0 & 0.500 \\
GPT-4.1-mini                    & 0.650 & 0.601 & 0.650 & 1.000 & 0.0 & 1.000 \\
DeepSeek-Llama-8B & 0.510 & 0.372 & 0.507 & 0.995 & 0.0 & 1.000 \\
Mistral-7B     & 0.500 & 0.500 & 0.068 & 0.137 & 0.0 & 1.000 \\
Qwen3-8B                       & 0.640 & 0.596 & 0.637 & 0.995 & 0.0 & 1.000 \\
Gemma-4B                & 0.500 & 0.500 & 0.068 & 0.137 & 0.0 & 1.000 \\
Gemma-1B                & 0.490 & 0.331 & 0.487 & 0.995 & 0.0 & 1.000 \\
Gemini-2.5-Pro                & 0.670 & 0.670 & 0.670 & 1.000 & 0.0 & 1.000 \\

\midrule
\multicolumn{7}{c}{\textbf{HELOC} — Few-shot}\\
\midrule
Model & Acc & Macro-F1 & PenAcc & Len-F1 & UnkLbl\% & Set-Jacc \\
\midrule
Llama3-3B                  & 0.038 & 0.077 & 0.000 & 0.074 & 0.0 & 0.500 \\
Gemma-1B                & 0.462 & 0.316 & 0.462 & 1.000 & 0.0 & 0.500 \\
Gemma-4B                & 0.885 & 0.883 & 0.811 & 0.852 & 0.0 & 1.000 \\
Qwen3-8B                       & 0.538 & 0.513 & 0.538 & 1.000 & 0.0 & 1.000 \\
Llama3-8B  & 0.462 & 0.324 & 0.452 & 0.980 & 0.0 & 0.500 \\
Gemini-2.5-Pro                & 0.615 & 0.613 & 0.615 & 1.000 & 0.0 & 1.000 \\
GPT-4.1-mini                    & 0.885 & 0.880 & 0.885 & 1.000 & 0.0 & 1.000 \\
Mistral-7B     & 0.038 & 0.077 & 0.000 & 0.143 & 0.0 & 1.000 \\

\bottomrule
\end{tabular}
\caption{HELOC: STaDS metrics per model for zero-shot and few-shot. Acc = accuracy; PenAcc = penalised accuracy; Len-F1 = length F1; UnkLbl\% = unknown label rate (\%).}
\label{tab:heloc-detailed}
\end{table*}

\begin{table*}[t]
\centering
\small
\begin{tabular}{lrrrrrr}
\toprule
\multicolumn{7}{c}{\textbf{Iris} — Zero-shot}\\
\midrule
Model & Acc & Macro-F1 & PenAcc & Len-F1 & UnkLbl\% & Set-Jacc \\
\midrule
Llama3-3B                  & 0.007 & 0.013 & 0.000 & 0.039 & 0.0 & 1.000 \\
GPT-4.1-mini                    & 0.487 & 0.516 & 0.458 & 0.944 & 0.0 & 1.000 \\
Llama3-8B  & 0.373 & 0.373 & 0.000 & 0.198 & 0.0 & 1.000 \\
Qwen3-8B                       & 0.440 & 0.402 & 0.039 & 0.198 & 0.0 & 1.000 \\
Mistral-7B     & 0.320 & 0.266 & 0.000 & 0.198 & 0.0 & 1.000 \\
Gemma-1B                & 0.007 & 0.013 & 0.000 & 0.997 & 98.0 & 0.020 \\
Gemma-4B                & 0.360 & 0.360 & 0.000 & 0.198 & 0.0 & 1.000 \\
Gemini-2.5-Pro                & 0.787 & 0.787 & 0.785 & 0.997 & 0.0 & 1.000 \\
DeepSeek-Llama-8B & 0.380 & 0.376 & 0.000 & 0.198 & 0.0 & 1.000 \\

\midrule
\multicolumn{7}{c}{\textbf{Iris} — Few-shot}\\
\midrule
Model & Acc & Macro-F1 & PenAcc & Len-F1 & UnkLbl\% & Set-Jacc \\
\midrule
Llama3-3B                  & 0.333 & 0.167 & 0.000 & 0.043 & 0.0 & 0.333 \\
Gemma-4B                & 0.333 & 0.334 & 0.152 & 0.638 & 0.0 & 1.000 \\
Gemma-1B                & 0.033 & 0.061 & 0.000 & 0.984 & 90.3 & 0.097 \\
Gemini-2.5-Pro                & 1.000 & 1.000 & 1.000 & 1.000 & 0.0 & 1.000 \\
GPT-4.1-mini                    & 0.667 & 0.662 & 0.667 & 1.000 & 0.0 & 1.000 \\
Qwen3-8B                       & 0.933 & 0.933 & 0.933 & 1.000 & 0.0 & 1.000 \\
DeepSeek-Llama-8B & 0.367 & 0.330 & 0.367 & 1.000 & 0.0 & 1.000 \\
Llama3-8B  & 0.300 & 0.295 & 0.170 & 0.741 & 0.0 & 1.000 \\
Mistral-7B     & 0.367 & 0.355 & 0.000 & 0.043 & 0.0 & 1.000 \\

\bottomrule
\end{tabular}
\caption{Iris: STaDS metrics per model for zero-shot and few-shot. Acc = accuracy; PenAcc = penalised accuracy; Len-F1 = length F1; UnkLbl\% = unknown label rate (\%).}
\label{tab:iris-detailed}
\end{table*}

\begin{table*}[t]
\centering
\small
\begin{tabular}{lrrrrrr}
\toprule
\multicolumn{7}{c}{\textbf{Monk 1} — Zero-shot}\\
\midrule
Model & Acc & Macro-F1 & PenAcc & Len-F1 & UnkLbl\% & Set-Jacc \\
\midrule
Qwen3-8B                       & 0.535 & 0.427 & 0.275 & 0.481 & 0.0 & 1.000 \\
Llama3-8B  & 0.502 & 0.502 & 0.243 & 0.481 & 0.0 & 1.000 \\
Gemma-4B                & 0.505 & 0.505 & 0.245 & 0.481 & 0.0 & 1.000 \\
Mistral-7B     & 0.521 & 0.507 & 0.261 & 0.481 & 0.0 & 1.000 \\
Gemma-1B                & 0.498 & 0.334 & 0.237 & 0.481 & 30.0 & 0.333 \\
Gemini-2.5-Pro                & 0.620 & 0.613 & 0.618 & 0.995 & 0.0 & 1.000 \\
GPT-4.1-mini                    & 0.118 & 0.193 & 0.000 & 0.348 & 0.0 & 1.000 \\
DeepSeek-Llama-8B & 0.507 & 0.352 & 0.247 & 0.481 & 0.0 & 1.000 \\
Llama3-3B                  & 0.500 & 0.333 & 0.240 & 0.481 & 0.0 & 0.500 \\

\midrule
\multicolumn{7}{c}{\textbf{Monk 1} — Few-shot}\\
\midrule
Model & Acc & Macro-F1 & PenAcc & Len-F1 & UnkLbl\% & Set-Jacc \\
\midrule
Gemma-1B               & 0.034 & 0.059 & 0.000 & 0.168 & 0.0 & 1.000 \\
Gemma-4B               & 0.011 & 0.023 & 0.000 & 0.023 & 0.0 & 0.500 \\
GPT-4.1-mini                   & 0.506 & 0.495 & 0.482 & 0.952 & 0.0 & 1.000 \\
Llama3-8B & 0.506 & 0.449 & 0.066 & 0.120 & 0.0 & 1.000 \\
Mistral-7B    & 0.540 & 0.512 & 0.100 & 0.120 & 0.0 & 1.000 \\
Gemini-2.5-Pro               & 0.724 & 0.724 & 0.721 & 0.994 & 0.0 & 1.000 \\
Llama3-3B                 & 0.494 & 0.331 & 0.054 & 0.120 & 0.0 & 1.000 \\
DeepSeek-Llama-8B & 0.483 & 0.476 & 0.043 & 0.120 & 0.0 & 1.000 \\
Qwen3-8B                      & 0.759 & 0.758 & 0.726 & 0.935 & 0.0 & 1.000 \\

\bottomrule
\end{tabular}
\caption{Monk 1: STaDS metrics per model for zero-shot and few-shot. Acc = accuracy; PenAcc = penalised accuracy; Len-F1 = length F1; UnkLbl\% = unknown label rate (\%).}
\label{tab:monk1-detailed}
\end{table*}

\begin{table*}[t]
\centering
\small
\begin{tabular}{lrrrrrr}
\toprule
\multicolumn{7}{c}{\textbf{Monk 2} — Zero-shot}\\
\midrule
Model & Acc & Macro-F1 & PenAcc & Len-F1 & UnkLbl\% & Set-Jacc \\
\midrule
Gemma-1B               & 0.174 & 0.187 & 0.000 & 0.481 & 40.0 & 0.200 \\
Gemma-4B               & 0.329 & 0.247 & 0.069 & 0.481 & 0.000 & 0.500 \\
Llama3-8B & 0.500 & 0.485 & 0.240 & 0.481 & 0.000 & 1.000 \\
DeepSeek-Llama-8B & 0.674 & 0.409 & 0.414 & 0.481 & 0.000 & 1.000 \\
Qwen3-8B                      & 0.664 & 0.476 & 0.405 & 0.481 & 0.000 & 1.000 \\
Mistral-7B    & 0.556 & 0.498 & 0.296 & 0.481 & 0.000 & 1.000 \\
Llama3-3B                 & 0.021 & 0.030 & 0.000 & 0.045 & 0.000 & 0.500 \\
GPT-4.1-mini                   & 0.660 & 0.397 & 0.334 & 0.348 & 0.000 & 1.000 \\
Gemini-2.5-Pro               & 0.484 & 0.475 & 0.461 & 0.955 & 0.000 & 1.000 \\

\midrule
\multicolumn{7}{c}{\textbf{Monk 2} — Few-shot}\\
\midrule
Model & Acc & Macro-F1 & PenAcc & Len-F1 & UnkLbl\% & Set-Jacc \\
\midrule
Gemma-1B               & 0.667 & 0.400 & 0.634 & 0.935 & 0.00  & 0.500 \\
Gemma-4B               & 0.563 & 0.422 & 0.123 & 0.120 & 0.00  & 1.000 \\
Llama3-8B & 0.701 & 0.525 & 0.261 & 0.120 & 0.00  & 1.000 \\
DeepSeek-Llama-8B & 0.667 & 0.552 & 0.227 & 0.120 & 0.00  & 1.000 \\
Qwen3-8B                      & 0.713 & 0.601 & 0.273 & 0.120 & 0.00  & 1.000 \\
Mistral-7B    & 0.678 & 0.436 & 0.238 & 0.120 & 0.00  & 1.000 \\
Llama3-3B                 & 0.667 & 0.400 & 0.227 & 0.120 & 0.00  & 0.500 \\
GPT-4.1-mini                   & 0.575 & 0.450 & 0.569 & 0.988 & 0.00  & 1.000 \\
Gemini-2.5-Pro               & 0.494 & 0.456 & 0.489 & 0.989 & 0.00  & 1.000 \\

\bottomrule
\end{tabular}
\caption{Monk 2: STaDS metrics per model for zero-shot and few-shot. Acc = accuracy; PenAcc = penalised accuracy; Len-F1 = length F1; UnkLbl\% = unknown label rate (\%).}
\label{tab:monk2-detailed}
\end{table*}

\begin{table*}[t]
\centering
\small
\begin{tabular}{lrrrrrr}
\toprule
\multicolumn{7}{c}{\textbf{Monk 3} — Zero-shot}\\
\midrule
Model & Acc & Macro-F1 & PenAcc & Len-F1 & UnkLbl\% & Set-Jacc \\
\midrule
DeepSeek-Llama-8B & 0.512 & 0.511 & 0.252 & 0.481 & 0.00 & 1.000 \\
Llama3-3B                 & 0.472 & 0.321 & 0.213 & 0.481 & 0.00 & 0.500 \\
Gemini-2.5-Pro               & 0.579 & 0.596 & 0.563 & 0.969 & 0.00 & 1.000 \\
GPT-4.1-mini                   & 0.507 & 0.423 & 0.181 & 0.348 & 0.00 & 1.000 \\
Llama3-8B & 0.495 & 0.495 & 0.236 & 0.481 & 0.00 & 1.000 \\
Gemma-4B               & 0.528 & 0.345 & 0.268 & 0.481 & 0.00 & 0.500 \\
Gemma-1B               & 0.528 & 0.345 & 0.268 & 0.481 & 0.00 & 0.500 \\
Mistral-7B    & 0.500 & 0.474 & 0.240 & 0.481 & 0.00 & 1.000 \\
Qwen3-8B                      & 0.491 & 0.362 & 0.231 & 0.481 & 0.00 & 1.000 \\

\midrule
\multicolumn{7}{c}{\textbf{Monk 3} — Few-shot}\\
\midrule
Model & Acc & Macro-F1 & PenAcc & Len-F1 & UnkLbl\% & Set-Jacc \\
\midrule
Llama3-8B & 0.494 & 0.494 & 0.054 & 0.120 & 0.00 & 1.000 \\
Gemini-2.5-Pro               & 0.644 & 0.642 & 0.644 & 1.000 & 0.00 & 1.000 \\
Qwen3-8B                      & 0.517 & 0.516 & 0.077 & 0.120 & 0.00 & 1.000 \\
Llama3-3B                 & 0.460 & 0.315 & 0.020 & 0.120 & 0.00 & 1.000 \\
Mistral-7B    & 0.437 & 0.335 & 0.000 & 0.120 & 0.00 & 1.000 \\
DeepSeek-Llama-8B & 0.437 & 0.432 & 0.000 & 0.120 & 0.00 & 1.000 \\
Gemma-1B               & 0.425 & 0.351 & 0.388 & 0.926 & 0.00 & 1.000 \\
Gemma-4B               & 0.529 & 0.346 & 0.089 & 0.120 & 0.00 & 0.500 \\
GPT-4.1-mini                   & 0.494 & 0.492 & 0.470 & 0.952 & 0.00 & 1.000 \\

\bottomrule
\end{tabular}
\caption{Monk 3: STaDS metrics per model for zero-shot and few-shot. Acc = accuracy; PenAcc = penalised accuracy; Len-F1 = length F1; UnkLbl\% = unknown label rate (\%).}
\label{tab:monk3-detailed}
\end{table*}

\begin{table*}[t]
\centering
\small
\begin{tabular}{lrrrrrr}
\toprule
\multicolumn{7}{c}{\textbf{Pima Indians Diabetes} — Zero-shot}\\
\midrule
Model & Acc & Macro-F1 & PenAcc & Len-F1 & UnkLbl\% & Set-Jacc \\
\midrule
Gemini-2.5-Pro               & 0.758 & 0.784 & 0.741 & 0.967 & 0.00 & 1.000 \\
Qwen3-8B                      & 0.538 & 0.531 & 0.306 & 0.536 & 0.00 & 1.000 \\
Llama3-8B & 0.496 & 0.332 & 0.264 & 0.536 & 0.00 & 1.000 \\
Mistral-7B    & 0.492 & 0.330 & 0.260 & 0.536 & 0.00 & 1.000 \\
DeepSeek-Llama-8B & 0.500 & 0.333 & 0.268 & 0.536 & 0.00 & 0.500 \\
GPT-4.1-mini                   & 0.452 & 0.311 & 0.148 & 0.392 & 0.00 & 1.000 \\
Gemma-1B               & 0.196 & 0.282 & 0.000 & 0.331 & 0.00 & 1.000 \\
Gemma-4B               & 0.020 & 0.038 & 0.000 & 0.039 & 0.00 & 0.500 \\
Llama3-3B                 & 0.002 & 0.004 & 0.000 & 0.008 & 0.00 & 1.000 \\

\midrule
\multicolumn{7}{c}{\textbf{Pima Indians Diabetes} — Few-shot}\\
\midrule
Model & Acc & Macro-F1 & PenAcc & Len-F1 & UnkLbl\% & Set-Jacc \\
\midrule
Gemini-2.5-Pro               & 0.820 & 0.814 & 0.820 & 1.000 & 0.00 & 1.000 \\
Qwen3-8B                      & 0.580 & 0.408 & 0.577 & 0.995 & 0.00 & 1.000 \\
DeepSeek-Llama-8B & 0.570 & 0.363 & 0.138 & 0.137 & 0.00 & 0.500 \\
Mistral-7B    & 0.510 & 0.508 & 0.078 & 0.137 & 0.00 & 1.000 \\
Gemma-4B               & 0.420 & 0.296 & 0.000 & 0.137 & 0.00 & 1.000 \\
Llama3-8B & 0.430 & 0.301 & 0.000 & 0.137 & 0.00 & 0.500 \\
GPT-4.1-mini                   & 0.390 & 0.281 & 0.390 & 1.000 & 0.00 & 1.000 \\
Gemma-1B               & 0.370 & 0.272 & 0.367 & 0.995 & 0.00 & 1.000 \\
Llama3-3B                 & 0.000 & 0.000 & 0.000 & 0.000 & 0.00 & 0.000 \\

\bottomrule
\end{tabular}
\caption{Pima Indians Diabetes: STaDS metrics per model for zero-shot and few-shot. Acc = accuracy; PenAcc = penalised accuracy; Len-F1 = length F1; UnkLbl\% = unknown label rate (\%).}
\label{tab:pima-detailed}
\end{table*}

\begin{table*}[t]
\centering
\small
\begin{tabular}{l|cc|c|cc|c}
\toprule
Dataset & \multicolumn{2}{c|}{Zero-shot (by PA)} & Best Z & \multicolumn{2}{c|}{Few-shot (by PA)} & Best F \\
& P-Acc & (Acc) & Model & P-Acc & (Acc) & Model \\
\midrule
Adult Income & 0.700 & 0.700 & Gemini-2.5-Pro & 0.737 & 0.737 & Gemini-2.5-Pro \\
Breast Cancer & 0.646 & 0.650 & Gemini-2.5-Pro & 0.732 & 0.732 & GPT-4.1-mini \\
Car Evaluation & 0.406 & 0.419 & Gemini-2.5-Pro & 0.586 & 0.600 & Gemini-2.5-Pro \\
COMPAS & 0.571 & 0.816 & Gemini-2.5-Pro & 0.697 & 0.716 & Gemini-2.5-Pro \\
Congression Voting& 0.388 & 0.397 & Gemini-2.5-Pro & 0.638 & 0.638 & Gemini-2.5-Pro \\
Gaussian Synthetic& 0.550 & 0.550 & Gemini-2.5-Pro & 0.880 & 0.880 & GPT-4.1-mini \\
German Credit & 0.118 & 0.550 & Mistral-7B-Instruct-v0.3 & 0.889 & 0.889 & Gemini-2.5-Pro \\
Give Me Some Credit & 0.825 & 0.830 & DeepSeek-Llama-8B & 0.917 & 0.917 & GPT-4.1-mini \\
Heart Disease & 0.640 & 0.640 & GPT-4.1-mini & 0.700 & 0.700 & GPT-4.1-mini \\
HELOC & 0.670 & 0.670 & Gemini-2.5-Pro & 0.885 & 0.885 & GPT-4.1-mini \\
Iris & 0.785 & 0.787 & Gemini-2.5-Pro & 1.000 & 1.000 & Gemini-2.5-Pro \\
Monk1 & 0.618 & 0.620 & Gemini-2.5-Pro & 0.726 & 0.759 & Qwen3-8B  \\
Monk2 & 0.461 & 0.484 & Gemini-2.5-Pro & 0.634 & 0.667 & Gemma-1B \\
Monk3 & 0.563 & 0.579 & Gemini-2.5-Pro & 0.644 & 0.644 & Gemini-2.5-Pro \\
Pima Diabetes & 0.741 & 0.758 & Gemini-2.5-Pro & 0.820 & 0.820 & Gemini-2.5-Pro \\
\bottomrule
\end{tabular}
\caption{Penalised accuracy (P-Acc) summary. Higher is better; penalisation reduces scores for overlong outputs and invalid labels.($\alpha${=}0.5, $\beta${=}0.5).}
\label{tab:icu-pa}
\end{table*}

\paragraph{Decision Faithfulness}
Table~\ref{tab:adult_au_metrics} - \ref{tab:monk1_metrics} list $\Delta_{\text{LAO}}$, \textsc{Self-Faith}, and \textsc{SelfAtt@k} for every model–dataset pair.  Fig.~\ref{fig:lao_delta_app_1}, \ref{fig:lao_delta_app_2}, and \ref{fig:pima_lao_delta} visualize heatmaps of LAO performance ($\Delta_{\text{LAO}}$) across all datasets.

Average of feature--target statistical dependency metrics across all benchmark datasets is provided in Table~\ref{tab:assoc_summary}.
``Triangulated'' faithfulness across all models and datasets are provided in Tabel~\ref{tab:rho_all}.

\begin{table}[htb]
\centering
\small
\begin{adjustbox}{max width=\linewidth}
\begin{tabular}{lccccp{5.5cm}}
\toprule
\textbf{Dataset} & \textbf{Mean Cramér’s V} & \textbf{Mean NMI} & 
\textbf{Mean Pearson $r$} & \textbf{Mean Spearman $\rho$} & \textbf{Top-3 by NMI} \\
\midrule
Adult Income & 0.308 & 0.053 & 0.143 & 0.164 & 
\texttt{relationship}, \texttt{marital-status}, \texttt{capital-gain} \\
Breast Cancer & 0.162 & 0.017 & -- & -- & 
\texttt{inv-nodes}, \texttt{deg-malig}, \texttt{irradiat} \\
Car Evaluation & 0.196 & 0.072 & -- & -- & 
\texttt{persons}, \texttt{safety}, \texttt{buying} \\
COMPAS & 0.203 & 0.027 & 0.071 & 0.086 & 
\texttt{decile\_score}, \texttt{score\_text}, \texttt{priors\_count} \\
Congressional Voting & 0.503 & 0.165 & -- & -- & 
\texttt{physician-fee-freeze}, \texttt{el-salvador-aid}, \texttt{education-spending} \\
Gaussian Synthetic & -- & 0.019 & $-0.019$ & $-0.022$ & 
\texttt{gauss\_1}, \texttt{gauss\_2}, \texttt{gauss\_6} \\
German Credit & 0.025 & 0.006 & $-0.025$ & $-0.023$ & 
\texttt{other-installment-plans}, \texttt{installment-rate}, \texttt{number-credits} \\
Give Me Some Credit & 0.205 & 0.014 & $-0.031$ & 0.017 & 
\texttt{RevolvingUtilizationOfUnsecuredLines}, 
\texttt{NumberOfTimes90DaysLate}, \texttt{NumberOfTime30-59DaysPastDueNotWorse} \\
Heart Disease & 0.083 & 0.008 & 0.122 & 0.104 & 
\texttt{age}, \texttt{prevalentHyp}, \texttt{sysBP} \\
HELOC & 0.189 & 0.030 & 0.035 & 0.036 & 
\texttt{ExternalRiskEstimate}, \texttt{NetFractionRevolvingBurden}, \texttt{PercentTradesWBalance} \\
Iris & 0.633 & 0.677 & 0.866 & 0.867 & 
\texttt{petal\_length}, \texttt{petal\_width}, \texttt{sepal\_length} \\
Monk1 & 0.095 & 0.041 & -- & -- & 
\texttt{a5}, \texttt{a2}, \texttt{a1} \\
Monk2 & 0.021 & 0.012 & -- & -- & 
\texttt{a4}, \texttt{a1}, \texttt{a6} \\
Monk3 & 0.213 & 0.075 & -- & -- & 
\texttt{a5}, \texttt{a2}, \texttt{a6} \\
Pima Diabetes & 0.252 & 0.047 & 0.206 & 0.224 & 
\texttt{Glucose}, \texttt{BMI}, \texttt{Age} \\
\bottomrule
\end{tabular}
\end{adjustbox}
\caption{Average of feature--target statistical dependency metrics across all benchmark datasets. 
Values are averaged over all features within each dataset. 
Dashes indicate non-applicable metrics (e.g., Pearson/Spearman for categorical targets). 
Top-ranked features by NMI highlight dominant statistical dependencies, 
which serve as proxies for co-occurrence rather than causal relationships.}
\label{tab:assoc_summary}
\end{table}

\begin{scriptsize}
\begin{longtable}{@{}llrrr@{}}
\caption{``Triangulated'' Faithfulness Across All Models and Datasets.}
\label{tab:rho_all}\\
\toprule
Dataset & Model & $\rho(\pi_{\text{self}}, \pi_{\text{LAO}})$ & $\rho(\pi_{\text{self}}, \pi_{\text{NMI}})$ & $\rho(\pi_{\text{LAO}}, \pi_{\text{NMI}})$ \\
\midrule
\endfirsthead
\toprule
Dataset & Model & $\rho(\pi_{\text{self}}, \pi_{\text{LAO}})$ & $\rho(\pi_{\text{self}}, \pi_{\text{NMI}})$ & $\rho(\pi_{\text{LAO}}, \pi_{\text{NMI}})$ \\ \midrule
\endhead
\midrule
\multicolumn{5}{r}{\emph{Continued on next page}}\\
\midrule
\endfoot
\bottomrule
\endlastfoot

\multicolumn{5}{@{}l}{\textbf{Adult Income}}\\
 & Gemma-4B & 0.552$^\dagger$ [0.098] & 0.394 [0.260] & 0.547$^*$ [0.043] \\
 & Gemma-1B & --- [---] & --- [---] & -0.165 [0.573] \\
 & Gemini-2.5-Pro & 0.253 [0.383] & 0.477$^\dagger$ [0.085] & 0.187 [0.523] \\
 & DeepSeek-Llama-8B & 0.240 [0.409] & 0.301 [0.296] & -0.187 [0.523] \\
 & Mistral-7B & -0.545$^\dagger$ [0.083] & -0.509 [0.110] & 0.235 [0.418] \\
 & Llama3-8B & -0.045 [0.894] & 0.436 [0.180] & -0.182 [0.533] \\
 & Qwen3-8B & -0.167 [0.668] & 0.350 [0.356] & 0.481$^\dagger$ [0.081] \\
 & GPT-4.1-mini & -0.015 [0.958] & 0.240 [0.409] & -0.618$^*$ [0.019] \\
 & Llama3-3B & -0.336 [0.240] & 0.121 [0.681] & -0.415 [0.140] \\
\addlinespace

\multicolumn{5}{@{}l}{\textbf{Breast Cancer}}\\
 & DeepSeek-Llama-8B & 0.810$^*$ [0.015] & 0.586 [0.127] & 0.009 [0.982] \\
 & Llama3-3B & -0.929$^*$ [0.003] & 0.259 [0.574] & 0.183 [0.638] \\
 & GPT-4.1-mini & 0.217 [0.576] & 0.775$^*$ [0.014] & 0.000 [1.000] \\
 & Mistral-7B & 0.150 [0.700] & 0.366 [0.333] & 0.583$^\dagger$ [0.099] \\
 & Gemini-2.5-Pro & -0.607 [0.148] & 0.595 [0.159] & -0.409 [0.274] \\
 & Llama3-8B & -0.183 [0.637] & 0.522 [0.149] & -0.148 [0.704] \\
 & Gemma-1B & -1.000$^*$ [0.000] & --- [---] & 0.574 [0.106] \\
 & Qwen3-8B & 0.333 [0.420] & 0.634$^\dagger$ [0.091] & -0.392 [0.297] \\
 & Gemma-4B & -0.048 [0.911] & 0.366 [0.373] & 0.078 [0.841] \\
\addlinespace

\multicolumn{5}{@{}l}{\textbf{Car Evaluation}}\\
 & Llama3-3B & -1.000 [---] & 1.000 [---] & 0.257 [0.623] \\
 & Gemini-2.5-Pro & 0.657 [0.156] & 0.886$^*$ [0.019] & 0.314 [0.544] \\
 & GPT-4.1-mini & -0.600 [0.208] & 0.657 [0.156] & -0.943$^*$ [0.005] \\
 & DeepSeek-Llama-8B & 0.029 [0.957] & 0.143 [0.787] & -0.143 [0.787] \\
 & Llama3-8B & 0.314 [0.544] & 0.657 [0.156] & 0.429 [0.397] \\
 & Mistral-7B & -0.086 [0.872] & 0.086 [0.872] & -0.543 [0.266] \\
 & Gemma-4B & -0.943$^*$ [0.005] & -0.086 [0.872] & -0.143 [0.787] \\
 & Gemma-1B & -0.600 [0.208] & -0.314 [0.544] & -0.086 [0.872] \\
 & Qwen3-8B & 0.543 [0.266] & 0.257 [0.623] & 0.429 [0.397] \\
\addlinespace

\multicolumn{5}{@{}l}{\textbf{COMPAS}}\\
 & Gemini-2.5-Pro & -0.576 [0.082] & 0.030 [0.934] & -0.212 [0.556] \\
 & Gemma-4B & 0.500 [0.207] & -0.167 [0.693] & -0.200 [0.580] \\
 & Gemma-1B & -0.452 [0.260] & -0.119 [0.779] & -0.515 [0.128] \\
 & Llama3-8B & 0.033 [0.932] & -0.417 [0.265] & 0.406 [0.244] \\
 & Mistral-7B & 0.881$^*$ [0.004] & 0.095 [0.823] & 0.285 [0.425] \\
 & Qwen3-8B & -0.455 [0.187] & 0.103 [0.777] & -0.248 [0.489] \\
 & GPT-4.1-mini & 0.212 [0.556] & 0.394 [0.260] & 0.248 [0.489] \\
 & Llama3-3B & -0.086 [0.872] & -0.543 [0.266] & 0.103 [0.777] \\
 & DeepSeek-Llama-8B & 0.030 [0.934] & 0.636$^*$ [0.048] & 0.188 [0.603] \\
\addlinespace

\multicolumn{5}{@{}l}{\textbf{Congressional Voting}}\\
 & DeepSeek-Llama-8B & 0.248 [0.392] & -0.095 [0.748] & -0.453 [0.078] \\
 & Llama3-8B & -0.243 [0.383] & -0.013 [0.965] & 0.477$^\dagger$ [0.062] \\
 & Mistral-7B & 0.445 [0.128] & -0.058 [0.851] & 0.319 [0.228] \\
 & GPT-4.1-mini & 0.226 [0.399] & 0.350 [0.184] & -0.037 [0.892] \\
 & Gemma-4B & 0.068 [0.810] & 0.579$^*$ [0.024] & 0.306 [0.249] \\
 & Gemma-1B & 0.643 [0.119] & -0.487 [0.268] & -0.311 [0.242] \\
 & Qwen3-8B & 0.014 [0.960] & 0.465 [0.081] & 0.255 [0.341] \\
 & Llama3-3B & -0.379 [0.147] & 0.041 [0.880] & 0.284 [0.286] \\
 & Gemini-2.5-Pro & 0.435 [0.092] & 0.748$^*$ [0.001] & 0.383 [0.144] \\
\addlinespace

\multicolumn{5}{@{}l}{\textbf{Gaussian Synthetic}}\\
 & Gemma-1B & --- [---] & --- [---] & -0.061 [0.798] \\
 & Gemma-4B & -0.060 [0.801] & 0.645$^*$ [0.002] & -0.301 [0.197] \\
 & DeepSeek-Llama-8B & 0.800 [0.200] & 0.800 [0.200] & 0.271 [0.248] \\
 & Llama3-3B & -0.232 [0.326] & 0.645$^*$ [0.002] & -0.057 [0.810] \\
 & GPT-4.1-mini & -0.477$^*$ [0.034] & 0.192 [0.418] & -0.101 [0.671] \\
 & Mistral-7B & -0.356 [0.123] & 0.575$^*$ [0.008] & -0.525$^*$ [0.017] \\
 & Qwen3-8B & 0.171 [0.470] & -0.193 [0.414] & -0.031 [0.897] \\
 & Gemini-2.5-Pro & -0.059 [0.806] & 0.380 [0.098] & 0.225 [0.340] \\
 & Llama3-8B & -0.018 [0.940] & 0.645$^*$ [0.002] & -0.306 [0.190] \\
\addlinespace

\multicolumn{5}{@{}l}{\textbf{German Credit}}\\
 & Llama3-3B & 0.344 [0.137] & 0.021 [0.929] & 0.268 [0.253] \\
 & DeepSeek-Llama-8B & 0.260 [0.283] & -0.108 [0.659] & -0.430 [0.059] \\
 & Llama3-8B & 0.586$^*$ [0.008] & -0.033 [0.893] & 0.092 [0.699] \\
 & Mistral-7B & -0.164 [0.631] & 0.569 [0.068] & 0.021 [0.929] \\
 & Qwen3-8B & -0.051 [0.836] & -0.066 [0.787] & -0.066 [0.783] \\
 & GPT-4.1-mini & -0.179 [0.450] & -0.053 [0.825] & -0.095 [0.689] \\
 & Gemma-1B & 0.250 [0.369] & 0.113 [0.689] & -0.148 [0.533] \\
 & Gemma-4B & 0.162 [0.521] & 0.125 [0.622] & 0.097 [0.684] \\
 & Gemini-2.5-Pro & 0.139 [0.701] & -0.020 [0.955] & 0.389 [0.266] \\
\addlinespace

\multicolumn{5}{@{}l}{\textbf{Give Me Some Credit}}\\
 & Mistral-7B & -0.429 [0.289] & 0.452 [0.260] & -0.394 [0.260] \\
 & GPT-4.1-mini & 0.236 [0.511] & 0.709$^*$ [0.022] & 0.261 [0.467] \\
 & Gemma-4B & -0.083 [0.831] & 0.400 [0.286] & -0.139 [0.701] \\
 & Gemini-2.5-Pro & -0.152 [0.676] & 0.758$^*$ [0.011] & 0.127 [0.726] \\
 & Llama3-8B & 0.224 [0.533] & 0.224 [0.533] & -0.564 [0.090] \\
 & Gemma-1B & --- [---] & --- [---] & 0.685$^*$ [0.029] \\
 & DeepSeek-Llama-8B & 0.595 [0.120] & 0.333 [0.420] & 0.673$^*$ [0.033] \\
 & Qwen3-8B & 0.214 [0.610] & 0.857$^*$ [0.007] & -0.030 [0.934] \\
 & Llama3-3B & 0.107 [0.819] & 0.893$^*$ [0.007] & 0.406 [0.244] \\
\addlinespace

\multicolumn{5}{@{}l}{\textbf{Heart Disease}}\\
 & Mistral-7B & -0.200 [0.475] & 0.170 [0.545] & -0.435 [0.105] \\
 & Gemma-1B & 0.039 [0.889] & -0.129 [0.647] & 0.275 [0.322] \\
 & Gemma-4B & -0.154 [0.633] & 0.303 [0.339] & 0.144 [0.609] \\
 & Gemini-2.5-Pro & 0.657$^*$ [0.008] & 0.572$^*$ [0.026] & 0.061 [0.829] \\
 & Llama3-8B & 0.032 [0.909] & 0.380 [0.162] & -0.046 [0.870] \\
 & GPT-4.1-mini & 0.496 [0.060] & 0.321 [0.244] & -0.100 [0.724] \\
 & Llama3-3B & 0.393 [0.147] & -0.129 [0.647] & -0.184 [0.511] \\
 & DeepSeek-Llama-8B & 0.115 [0.707] & 0.328 [0.274] & -0.266 [0.339] \\
 & Qwen3-8B & 0.054 [0.850] & 0.245 [0.378] & 0.172 [0.541] \\
\addlinespace

\multicolumn{5}{@{}l}{\textbf{HELOC}}\\
 & Gemma-4B & 0.139 [0.536] & 0.194 [0.388] & 0.024 [0.914] \\
 & Gemma-1B & 1.000 [---] & 1.000 [---] & 0.196 [0.371] \\
 & GPT-4.1-mini & -0.229 [0.293] & 0.173 [0.430] & 0.060 [0.785] \\
 & DeepSeek-Llama-8B & -0.005 [0.984] & 0.265 [0.287] & 0.123 [0.578] \\
 & Gemini-2.5-Pro & 0.251 [0.248] & 0.696$^*$ [0.000] & 0.275 [0.205] \\
 & Llama3-3B & -0.218 [0.317] & 0.133 [0.544] & -0.009 [0.968] \\
 & Llama3-8B & 0.156 [0.564] & -0.097 [0.721] & 0.199 [0.364] \\
 & Qwen3-8B & -0.170 [0.438] & 0.042 [0.847] & 0.410 [0.052] \\
 & Mistral-7B & 0.462$^*$ [0.030] & 0.263 [0.238] & 0.190 [0.386] \\
\addlinespace

\multicolumn{5}{@{}l}{\textbf{Iris}}\\
 & Llama3-8B & 0.200 [0.800] & 0.400 [0.600] & 0.800 [0.200] \\
 & DeepSeek-Llama-8B & -0.800 [0.200] & -0.600 [0.400] & 0.000 [1.000] \\
 & Mistral-7B & 0.800 [0.200] & -0.600 [0.400] & -0.800 [0.200] \\
 & Gemini-2.5-Pro & 0.800 [0.200] & 1.000$^*$ [0.000] & 0.800 [0.200] \\
 & Gemma-1B & 0.600 [0.400] & -0.600 [0.400] & -1.000$^*$ [0.000] \\
 & Gemma-4B & 0.400 [0.600] & -0.600 [0.400] & 0.400 [0.600] \\
 & Qwen3-8B & 0.400 [0.600] & 1.000$^*$ [0.000] & 0.400 [0.600] \\
 & GPT-4.1-mini & 0.800 [0.200] & 1.000$^*$ [0.000] & 0.800 [0.200] \\
 & Llama3-3B & -0.600 [0.400] & -0.600 [0.400] & 1.000$^*$ [0.000] \\
\addlinespace

\multicolumn{5}{@{}l}{\textbf{Monk1}}\\
 & DeepSeek-Llama-8B & 0.200 [0.704] & 0.676 [0.140] & 0.845$^*$ [0.034] \\
 & Gemini-2.5-Pro & 0.943$^*$ [0.005] & 0.372 [0.468] & 0.541 [0.268] \\
 & Llama3-3B & -0.943$^*$ [0.005] & -0.101 [0.848] & 0.338 [0.512] \\
 & Mistral-7B & 1.000$^*$ [0.000] & 0.500 [0.667] & -0.068 [0.899] \\
 & Llama3-8B & -0.657 [0.156] & 0.372 [0.468] & 0.034 [0.949] \\
 & GPT-4.1-mini & 0.314 [0.544] & 0.372 [0.468] & -0.135 [0.798] \\
 & Gemma-4B & -0.900$^*$ [0.037] & -0.112 [0.858] & -0.372 [0.468] \\
 & Gemma-1B & -0.200 [0.704] & -0.101 [0.848] & -0.778 [0.069] \\
 & Qwen3-8B & 0.500 [0.667] & -1.000$^*$ [0.000] & -0.169 [0.749] \\
\addlinespace

\multicolumn{5}{@{}l}{\textbf{Monk2}}\\
 & Gemini-2.5-Pro & -0.371 [0.468] & 0.030 [0.954] & 0.152 [0.774] \\
 & Llama3-8B & 0.886$^*$ [0.019] & 0.030 [0.954] & -0.030 [0.954] \\
 & Gemma-1B & -0.771 [0.072] & 0.030 [0.954] & -0.638 [0.173] \\
 & Gemma-4B & 0.771 [0.072] & 0.030 [0.954] & -0.152 [0.774] \\
 & DeepSeek-Llama-8B & 0.000 [1.000] & -0.316 [0.684] & -0.395 [0.439] \\
 & Mistral-7B & -0.657 [0.156] & 0.030 [0.954] & 0.516 [0.295] \\
 & Qwen3-8B & 0.314 [0.544] & -0.091 [0.864] & -0.030 [0.954] \\
 & GPT-4.1-mini & 0.771 [0.072] & 0.030 [0.954] & -0.213 [0.686] \\
 & Llama3-3B & 0.086 [0.872] & 0.030 [0.954] & 0.030 [0.954] \\
\addlinespace

\multicolumn{5}{@{}l}{\textbf{Monk3}}\\
 & Mistral-7B & 0.500 [0.667] & 0.500 [0.667] & -0.395 [0.439] \\
 & Gemma-1B & -0.200 [0.704] & -0.395 [0.439] & 0.334 [0.518] \\
 & Gemma-4B & 0.500 [0.391] & -0.447 [0.450] & 0.516 [0.295] \\
 & Llama3-8B & 0.143 [0.787] & 0.395 [0.439] & -0.152 [0.774] \\
 & GPT-4.1-mini & -0.600 [0.208] & 0.395 [0.439] & -0.395 [0.439] \\
 & Qwen3-8B & 1.000$^*$ [0.000] & 0.500 [0.667] & 0.516 [0.295] \\
 & Gemini-2.5-Pro & -0.200 [0.704] & 0.577 [0.231] & 0.334 [0.518] \\
 & DeepSeek-Llama-8B & 0.714 [0.111] & 0.395 [0.439] & 0.395 [0.439] \\
 & Llama3-3B & -0.143 [0.787] & -0.395 [0.439] & -0.516 [0.295] \\
\addlinespace

\multicolumn{5}{@{}l}{\textbf{Pima Diabetes}}\\
 & Gemini-2.5-Pro & 0.429 [0.289] & 0.976$^*$ [0.000] & 0.571 [0.139] \\
 & Gemma-1B & -0.429 [0.289] & 0.429 [0.289] & 0.238 [0.570] \\
 & Gemma-4B & 0.333 [0.420] & 0.095 [0.823] & -0.571 [0.139] \\
 & Qwen3-8B & -0.429 [0.289] & 0.857$^*$ [0.007] & -0.286 [0.493] \\
 & GPT-4.1-mini & -0.095 [0.823] & 0.905$^*$ [0.002] & -0.333 [0.420] \\
 & DeepSeek-Llama-8B & 0.952$^*$ [0.000] & -0.286 [0.493] & -0.286 [0.493] \\
 & Llama3-3B & -0.393 [0.383] & -0.036 [0.939] & -0.286 [0.493] \\
 & Llama3-8B & -0.381 [0.352] & -0.262 [0.531] & -0.286 [0.493] \\
 & Mistral-7B & -0.400 [0.505] & -0.300 [0.624] & 0.048 [0.911] \\
\end{longtable}

\vspace{-0.5em}
\noindent\emph{Notes.} Dashes (---) denote undefined due to NaNs or degenerate ranks. Stars: $^*$ $p{<}.05$, $^\dagger$ $p{<}.10$. Brackets show $p$-values.
\end{scriptsize}

\end{document}

%% file: math_commands.tex
\usepackage{amsmath,amsfonts,bm}

\def\eqref#1{equation~\ref{#1}}
\def\1{\bm{1}}

\DeclareMathAlphabet{\mathsfit}{\encodingdefault}{\sfdefault}{m}{sl}
\SetMathAlphabet{\mathsfit}{bold}{\encodingdefault}{\sfdefault}{bx}{n}

%% file: main.bib
@article{chi1981categorization,
  title={Categorization and representation of physics problems by experts and novices},
  author={Chi, Michelene TH and Feltovich, Paul J and Glaser, Robert},
  journal={Cognitive science},
  volume={5},
  number={2},
  pages={121--152},
  year={1981},
  publisher={Elsevier}
}

@book{bransford2000people,
  title={How people learn},
  author={Bransford, John D and Brown, Ann L and Cocking, Rodney R and others},
  volume={11},
  year={2000},
  publisher={Washington, DC: National academy press}
}

@book{chi2014nature,
  title={The nature of expertise},
  author={Chi, Michelene TH and Glaser, Robert and Farr, Marshall J},
  year={2014},
  publisher={Psychology Press}
}

@article{abd2023large,
  title={Large language models in medical education: opportunities, challenges, and future directions},
  author={Abd-Alrazaq, Alaa and AlSaad, Rawan and Alhuwail, Dari and Ahmed, Arfan and Healy, Padraig Mark and Latifi, Syed and Aziz, Sarah and Damseh, Rafat and Alrazak, Sadam Alabed and Sheikh, Javaid},
  journal={JMIR medical education},
  volume={9},
  number={1},
  pages={e48291},
  year={2023},
  publisher={JMIR Publications Inc., Toronto, Canada}
}

@inproceedings{Tam2024speak,
    title = "Let Me Speak Freely? A Study On The Impact Of Format Restrictions On Large Language Model Performance.",
    author = "Tam, Zhi Rui  and
      Wu, Cheng-Kuang  and
      Tsai, Yi-Lin  and
      Lin, Chieh-Yen  and
      Lee, Hung-yi  and
      Chen, Yun-Nung",
    editor = "Dernoncourt, Franck  and
      Preo{\c{t}}iuc-Pietro, Daniel  and
      Shimorina, Anastasia",
    booktitle = "Proceedings of the 2024 Conference on Empirical Methods in Natural Language Processing: Industry Track",
    month = nov,
    year = "2024",
    address = "Miami, Florida, US",
    publisher = "Association for Computational Linguistics",
    url = "https://aclanthology.org/2024.emnlp-industry.91/",
    doi = "10.18653/v1/2024.emnlp-industry.91",
    pages = "1218--1236"
}

@article{xin2022exploring,
  title={Exploring the whole rashomon set of sparse decision trees},
  author={Xin, Rui and Zhong, Chudi and Chen, Zhi and Takagi, Takuya and Seltzer, Margo and Rudin, Cynthia},
  journal={Advances in neural information processing systems},
  volume={35},
  pages={14071--14084},
  year={2022}
}

@misc{asuncion2007uci,
  title={UCI machine learning repository},
  author={Asuncion, Arthur and Newman, David and others},
  year={2007},
  publisher={Irvine, CA, USA}
}

@inproceedings{smith1988using,
  title={Using the ADAP learning algorithm to forecast the onset of diabetes mellitus},
  author={Smith, Jack W and Everhart, James E and Dickson, William C and Knowler, William C and Johannes, Robert Scott},
  booktitle={Proceedings of the annual symposium on computer application in medical care},
  pages={261},
  year={1988}
}

@article{jordan2015effect,
  title={The effect of race/ethnicity on sentencing: Examining sentence type, jail length, and prison length},
  author={Jordan, Kareem L and Freiburger, Tina L},
  journal={Journal of Ethnicity in Criminal Justice},
  volume={13},
  number={3},
  pages={179--196},
  year={2015},
  publisher={Taylor \& Francis}
}

@article{yeh2009comparisons,
  title={The comparisons of data mining techniques for the predictive accuracy of probability of default of credit card clients},
  author={Yeh, I-Cheng and Lien, Che-hui},
  journal={Expert systems with applications},
  volume={36},
  number={2},
  pages={2473--2480},
  year={2009},
  publisher={Elsevier}
}

@inproceedings{Samek2017Explain,
  title={Explainable Artificial Intelligence: Understanding, Visualizing and Interpreting Deep Learning Models},
  author={Wojciech Samek and Thomas Wiegand and Klaus-Robert M\"{u}ller},
  booktitle={ITG Symposium on Image Processing},
  year={2017}
}

@article{Nauta2021Survey,
  title={An Overview of Concept-Based Interpretability in Machine Learning},
  author={Meike Nauta and Cees Snoek},
  journal={IEEE Transactions on Pattern Analysis and Machine Intelligence},
  volume={45},
  number={4},
  year={2023},
  pages={4701--4719}
}

@article{agarwal2022openxai,
  title={Openxai: Towards a transparent evaluation of model explanations},
  author={Agarwal, Chirag and Krishna, Satyapriya and Saxena, Eshika and Pawelczyk, Martin and Johnson, Nari and Puri, Isha and Zitnik, Marinka and Lakkaraju, Himabindu},
  journal={Advances in neural information processing systems},
  volume={35},
  pages={15784--15799},
  year={2022}
}

@article{Unwin2021TheID,
  title={The iris data set: In search of the source of virginica},
  author={Antony Unwin and Kim Kleinman},
  journal={Significance},
  year={2021},
  volume={18},
  url={https://api.semanticscholar.org/CorpusID:244763032}
}

@article{llama3,
  title={The llama 3 herd of models},
  author={Dubey, Abhimanyu and Jauhri, Abhinav and Pandey, Abhinav and Kadian, Abhishek and Al-Dahle, Ahmad and Letman, Aiesha and Mathur, Akhil and Schelten, Alan and Yang, Amy and Fan, Angela and others},
  journal={arXiv e-prints},
  pages={arXiv--2407},
  year={2024}
}

@article{mistral,
  title   = {Mistral 7B},
  author  = {Jiang, Albert Q. and Sablayrolles, Alexandre and Mensch, Arthur and Bamford, Chris and Chaplot, Devendra Singh and de las Casas, Diego and Bressand, Florian and Lengyel, Gianna and Lample, Guillaume and Saulnier, Lucile and Renard Lavaud, L{\'e}lio and Lachaux, Marie–Anne and Stock, Pierre and Le Scao, Teven and Lavril, Thibaut and Wang, Thomas and Lacroix, Timoth{\'e}e and El Sayed, William},
  year    = {2023},
  eprint  = {2310.06825},
  archivePrefix = {arXiv},
  primaryClass  = {cs.CL},
  url     = {https://arxiv.org/abs/2310.06825}
}

@misc{deepseek,
  title   = {DeepSeek LLMs: Efficient Distillation of Llama Models},
  author  = {DeepSeek},
  year    = {2024},
  note    = {Technical report},
  url     = {https://github.com/deepseek-ai}
}

@article{qwen,
  title={Qwen3 technical report},
  author={Yang, An and Li, Anfeng and Yang, Baosong and Zhang, Beichen and Hui, Binyuan and Zheng, Bo and Yu, Bowen and Gao, Chang and Huang, Chengen and Lv, Chenxu and others},
  journal={arXiv preprint arXiv:2505.09388},
  year={2025}
}

@article{gemma,
  title={Gemma 3 technical report},
  author={Team, Gemma and Kamath, Aishwarya and Ferret, Johan and Pathak, Shreya and Vieillard, Nino and Merhej, Ramona and Perrin, Sarah and Matejovicova, Tatiana and Ram{\'e}, Alexandre and Rivi{\`e}re, Morgane and others},
  journal={arXiv preprint arXiv:2503.19786},
  year={2025}
}

@article{gemini,
  title={Gemini: a family of highly capable multimodal models},
  author={Team, Gemini and Anil, Rohan and Borgeaud, Sebastian and Alayrac, Jean-Baptiste and Yu, Jiahui and Soricut, Radu and Schalkwyk, Johan and Dai, Andrew M and Hauth, Anja and Millican, Katie and others},
  journal={arXiv preprint arXiv:2312.11805},
  year={2023}
}

@article{gpt4,
  title={Gpt-4 technical report},
  author={Achiam, Josh and Adler, Steven and Agarwal, Sandhini and Ahmad, Lama and Akkaya, Ilge and Aleman, Florencia Leoni and Almeida, Diogo and Altenschmidt, Janko and Altman, Sam and Anadkat, Shyamal and others},
  journal={arXiv preprint arXiv:2303.08774},
  year={2023}
}

@article{thrun1991monk,
  title={The MONK's Problems: A Performance Comparison of Different Learning Algorithems},
  author={Thrun, Sebastian},
  journal={Technical Report of Carnegie Mellon University},
  year={1991}
}

@misc{congressional_voting,
  title        = {{Congressional Voting Records}},
  year         = {1987},
  howpublished = {UCI Machine Learning Repository},
  note         = {{DOI}: https://doi.org/10.24432/C5C01P}
}

@misc{freshcorn2022gmsc,
  author       = {Bryce Freshcorn},
  title        = {Give Me Some Credit :: 2011 Competition Data},
  year         = {2022},
  howpublished = {\url{https://www.kaggle.com/datasets/brycecf/give-me-some-credit-dataset}},
  note         = {Accessed: 2025-08-01}
}

@misc{who_cvd_2021,
  author       = {{World Health Organization}},
  title        = {Cardiovascular diseases (CVDs): fact sheet},
  year         = {2021},
  howpublished = {\url{https://www.who.int/news-room/fact-sheets/detail/cardiovascular-diseases-(cvds)}},
  note         = {Accessed: 2025-08-01}
}

@inproceedings{sui2024table,
  title={Table meets llm: Can large language models understand structured table data? a benchmark and empirical study},
  author={Sui, Yuan and Zhou, Mengyu and Zhou, Mingjie and Han, Shi and Zhang, Dongmei},
  booktitle={Proceedings of the 17th ACM International Conference on Web Search and Data Mining},
  pages={645--654},
  year={2024}
}

@article{zhao2023survey,
  title={A survey of large language models},
  author={Zhao, Wayne Xin and Zhou, Kun and Li, Junyi and Tang, Tianyi and Wang, Xiaolei and Hou, Yupeng and Min, Yingqian and Zhang, Beichen and Zhang, Junjie and Dong, Zican and others},
  journal={arXiv preprint arXiv:2303.18223},
  volume={1},
  number={2},
  year={2023}
}

@article{mayer1989models,
  title={Models for understanding},
  author={Mayer, Richard E},
  journal={Review of educational research},
  volume={59},
  number={1},
  pages={43--64},
  year={1989},
  publisher={Sage Publications Sage CA: Thousand Oaks, CA}
}

@book{bereiter2005education,
  title={Education and mind in the knowledge age},
  author={Bereiter, Carl},
  year={2005},
  publisher={Routledge}
}

@article{ali2023explainable,
  title={Explainable Artificial Intelligence (XAI): What we know and what is left to attain Trustworthy Artificial Intelligence},
  author={Ali, Sajid and Abuhmed, Tamer and El-Sappagh, Shaker and Muhammad, Khan and Alonso-Moral, Jose M and Confalonieri, Roberto and Guidotti, Riccardo and Del Ser, Javier and D{\'\i}az-Rodr{\'\i}guez, Natalia and Herrera, Francisco},
  journal={Information fusion},
  volume={99},
  pages={101805},
  year={2023},
  publisher={Elsevier}
}

@article{adadi2018peeking,
  title={Peeking inside the black-box: a survey on explainable artificial intelligence (XAI)},
  author={Adadi, Amina and Berrada, Mohammed},
  journal={IEEE access},
  volume={6},
  pages={52138--52160},
  year={2018},
  publisher={IEEE}
}

@article{das2020opportunities,
  title={Opportunities and challenges in explainable artificial intelligence (xai): A survey},
  author={Das, Arun and Rad, Paul},
  journal={arXiv preprint arXiv:2006.11371},
  year={2020}
}

@article{li2022interpretable,
  title={Interpretable deep learning: Interpretation, interpretability, trustworthiness, and beyond},
  author={Li, Xuhong and Xiong, Haoyi and Li, Xingjian and Wu, Xuanyu and Zhang, Xiao and Liu, Ji and Bian, Jiang and Dou, Dejing},
  journal={Knowledge and Information Systems},
  volume={64},
  number={12},
  pages={3197--3234},
  year={2022},
  publisher={Springer}
}

@article{wang2023label,
  title={Label words are anchors: An information flow perspective for understanding in-context learning},
  author={Wang, Lean and Li, Lei and Dai, Damai and Chen, Deli and Zhou, Hao and Meng, Fandong and Zhou, Jie and Sun, Xu},
  journal={arXiv preprint arXiv:2305.14160},
  year={2023}
}

@article{petroni2019language,
  title={Language models as knowledge bases?},
  author={Petroni, Fabio and Rockt{\"a}schel, Tim and Lewis, Patrick and Bakhtin, Anton and Wu, Yuxiang and Miller, Alexander H and Riedel, Sebastian},
  journal={arXiv preprint arXiv:1909.01066},
  year={2019}
}

@article{dong2022survey,
  title={A survey on in-context learning},
  author={Dong, Qingxiu and Li, Lei and Dai, Damai and Zheng, Ce and Ma, Jingyuan and Li, Rui and Xia, Heming and Xu, Jingjing and Wu, Zhiyong and Liu, Tianyu and others},
  journal={arXiv preprint arXiv:2301.00234},
  year={2022}
}

@article{liu-etal-2024-lost,
    title = "Lost in the Middle: How Language Models Use Long Contexts",
    author = "Liu, Nelson F.  and
      Lin, Kevin  and
      Hewitt, John  and
      Paranjape, Ashwin  and
      Bevilacqua, Michele  and
      Petroni, Fabio  and
      Liang, Percy",
    journal = "Transactions of the Association for Computational Linguistics",
    volume = "12",
    year = "2024",
    address = "Cambridge, MA",
    publisher = "MIT Press",
    url = "https://aclanthology.org/2024.tacl-1.9/",
    doi = "10.1162/tacl_a_00638",
    pages = "157--173"
}

@article{min2023factscore,
  title={Factscore: Fine-grained atomic evaluation of factual precision in long form text generation},
  author={Min, Sewon and Krishna, Kalpesh and Lyu, Xinxi and Lewis, Mike and Yih, Wen-tau and Koh, Pang Wei and Iyyer, Mohit and Zettlemoyer, Luke and Hajishirzi, Hannaneh},
  journal={arXiv preprint arXiv:2305.14251},
  year={2023}
}

@article{deyoung2019eraser,
  title={ERASER: A benchmark to evaluate rationalized NLP models},
  author={DeYoung, Jay and Jain, Sarthak and Rajani, Nazneen Fatema and Lehman, Eric and Xiong, Caiming and Socher, Richard and Wallace, Byron C},
  journal={arXiv preprint arXiv:1911.03429},
  year={2019}
}

@article{cobbe2021training,
  title={Training verifiers to solve math word problems},
  author={Cobbe, Karl and Kosaraju, Vineet and Bavarian, Mohammad and Chen, Mark and Jun, Heewoo and Kaiser, Lukasz and Plappert, Matthias and Tworek, Jerry and Hilton, Jacob and Nakano, Reiichiro and others},
  journal={arXiv preprint arXiv:2110.14168},
  year={2021}
}

@article{hendrycks2020measuring,
  title={Measuring massive multitask language understanding},
  author={Hendrycks, Dan and Burns, Collin and Basart, Steven and Zou, Andy and Mazeika, Mantas and Song, Dawn and Steinhardt, Jacob},
  journal={arXiv preprint arXiv:2009.03300},
  year={2020}
}

@article{wang2018glue,
  title={GLUE: A multi-task benchmark and analysis platform for natural language understanding},
  author={Wang, Alex and Singh, Amanpreet and Michael, Julian and Hill, Felix and Levy, Omer and Bowman, Samuel R},
  journal={arXiv preprint arXiv:1804.07461},
  year={2018}
}

@article{dua2019drop,
  title={DROP: A reading comprehension benchmark requiring discrete reasoning over paragraphs},
  author={Dua, Dheeru and Wang, Yizhong and Dasigi, Pradeep and Stanovsky, Gabriel and Singh, Sameer and Gardner, Matt},
  journal={arXiv preprint arXiv:1903.00161},
  year={2019}
}

@article{camburu2018snli,
  title={e-snli: Natural language inference with natural language explanations},
  author={Camburu, Oana-Maria and Rockt{\"a}schel, Tim and Lukasiewicz, Thomas and Blunsom, Phil},
  journal={Advances in Neural Information Processing Systems},
  volume={31},
  year={2018}
}

@article{brown2020language,
  title={Language models are few-shot learners},
  author={Brown, Tom and Mann, Benjamin and Ryder, Nick and Subbiah, Melanie and Kaplan, Jared D and Dhariwal, Prafulla and Neelakantan, Arvind and Shyam, Pranav and Sastry, Girish and Askell, Amanda and others},
  journal={Advances in neural information processing systems},
  volume={33},
  pages={1877--1901},
  year={2020}
}

@inproceedings{bender2021dangers,
  title={On the dangers of stochastic parrots: Can language models be too big?},
  author={Bender, Emily M and Gebru, Timnit and McMillan-Major, Angelina and Shmitchell, Shmargaret},
  booktitle={Proceedings of the 2021 ACM conference on fairness, accountability, and transparency},
  pages={610--623},
  year={2021}
}

@article{wei2022chain,
  title={Chain-of-thought prompting elicits reasoning in large language models},
  author={Wei, Jason and Wang, Xuezhi and Schuurmans, Dale and Bosma, Maarten and Xia, Fei and Chi, Ed and Le, Quoc V and Zhou, Denny and others},
  journal={Advances in neural information processing systems},
  volume={35},
  pages={24824--24837},
  year={2022}
}

@article{lanham2023measuring,
  title={Measuring faithfulness in chain-of-thought reasoning},
  author={Lanham, Tamera and Chen, Anna and Radhakrishnan, Ansh and Steiner, Benoit and Denison, Carson and Hernandez, Danny and Li, Dustin and Durmus, Esin and Hubinger, Evan and Kernion, Jackson and others},
  journal={arXiv preprint arXiv:2307.13702},
  year={2023}
}

@article{li2023evaluating,
  title={Evaluating object hallucination in large vision-language models},
  author={Li, Yifan and Du, Yifan and Zhou, Kun and Wang, Jinpeng and Zhao, Wayne Xin and Wen, Ji-Rong},
  journal={arXiv preprint arXiv:2305.10355},
  year={2023}
}

@article{zarlenga2023tabcbm,
  title={Tabcbm: Concept-based interpretable neural networks for tabular data},
  author={Zarlenga, Mateo Espinosa and Shams, Zohreh and Nelson, Michael Edward and Kim, Been and Jamnik, Mateja},
  year={2023}
}

@article{kim2024ambiguous,
  title={How Ambiguous Are the Rationales for Natural Language Reasoning? A Simple Approach to Handling Rationale Uncertainty},
  author={Kim, Hazel H},
  journal={arXiv preprint arXiv:2402.14337},
  year={2024}
}

@article{turpin2023language,
  title={Language models don't always say what they think: Unfaithful explanations in chain-of-thought prompting},
  author={Turpin, Miles and Michael, Julian and Perez, Ethan and Bowman, Samuel},
  journal={Advances in Neural Information Processing Systems},
  volume={36},
  pages={74952--74965},
  year={2023}
}

@article{lewkowycz2022solving,
  title={Solving quantitative reasoning problems with language models},
  author={Lewkowycz, Aitor and Andreassen, Anders and Dohan, David and Dyer, Ethan and Michalewski, Henryk and Ramasesh, Vinay and Slone, Ambrose and Anil, Cem and Schlag, Imanol and Gutman-Solo, Theo and others},
  journal={Advances in neural information processing systems},
  volume={35},
  pages={3843--3857},
  year={2022}
}

@article{wang2024chain,
  title={Chain-of-table: Evolving tables in the reasoning chain for table understanding},
  author={Wang, Zilong and Zhang, Hao and Li, Chun-Liang and Eisenschlos, Julian Martin and Perot, Vincent and Wang, Zifeng and Miculicich, Lesly and Fujii, Yasuhisa and Shang, Jingbo and Lee, Chen-Yu and others},
  journal={arXiv preprint arXiv:2401.04398},
  year={2024}
}

@article{wei2023larger,
  title={Larger language models do in-context learning differently},
  author={Wei, Jerry and Wei, Jason and Tay, Yi and Tran, Dustin and Webson, Albert and Lu, Yifeng and Chen, Xinyun and Liu, Hanxiao and Huang, Da and Zhou, Denny and others},
  journal={arXiv preprint arXiv:2303.03846},
  year={2023}
}

@article{min2022rethinking,
  title={Rethinking the role of demonstrations: What makes in-context learning work?},
  author={Min, Sewon and Lyu, Xinxi and Holtzman, Ari and Artetxe, Mikel and Lewis, Mike and Hajishirzi, Hannaneh and Zettlemoyer, Luke},
  journal={arXiv preprint arXiv:2202.12837},
  year={2022}
}

@article{akyurek2022learning,
  title={What learning algorithm is in-context learning? investigations with linear models},
  author={Aky{\"u}rek, Ekin and Schuurmans, Dale and Andreas, Jacob and Ma, Tengyu and Zhou, Denny},
  journal={arXiv preprint arXiv:2211.15661},
  year={2022}
}

@article{liu2023towards,
  title={Towards understanding in-context learning with contrastive demonstrations and saliency maps},
  author={Liu, Fuxiao and Xu, Paiheng and Li, Zongxia and Feng, Yue and Song, Hyemi},
  journal={arXiv preprint arXiv:2307.05052},
  year={2023}
}

@article{chen2023self,
  title={Self-icl: Zero-shot in-context learning with self-generated demonstrations},
  author={Chen, Wei-Lin and Wu, Cheng-Kuang and Chen, Yun-Nung and Chen, Hsin-Hsi},
  journal={arXiv preprint arXiv:2305.15035},
  year={2023}
}

@article{zhang2023tablellama,
  title={Tablellama: Towards open large generalist models for tables},
  author={Zhang, Tianshu and Yue, Xiang and Li, Yifei and Sun, Huan},
  journal={arXiv preprint arXiv:2311.09206},
  year={2023}
}

@article{jacovi2020towards,
  title={Towards faithfully interpretable NLP systems: How should we define and evaluate faithfulness?},
  author={Jacovi, Alon and Goldberg, Yoav},
  journal={arXiv preprint arXiv:2004.03685},
  year={2020}
}

@article{herzig2020tapas,
  title={TaPas: Weakly supervised table parsing via pre-training},
  author={Herzig, Jonathan and Nowak, Pawe{\l} Krzysztof and M{\"u}ller, Thomas and Piccinno, Francesco and Eisenschlos, Julian Martin},
  journal={arXiv preprint arXiv:2004.02349},
  year={2020}
}

@article{deng2022turl,
  title={Turl: Table understanding through representation learning},
  author={Deng, Xiang and Sun, Huan and Lees, Alyssa and Wu, You and Yu, Cong},
  journal={ACM SIGMOD Record},
  volume={51},
  number={1},
  pages={33--40},
  year={2022},
  publisher={ACM New York, NY, USA}
}

@article{yu2024natural,
  title={Natural language reasoning, a survey},
  author={Yu, Fei and Zhang, Hongbo and Tiwari, Prayag and Wang, Benyou},
  journal={ACM Computing Surveys},
  volume={56},
  number={12},
  pages={1--39},
  year={2024},
  publisher={ACM New York, NY}
}

@article{holter2018fico,
  title={Fico explainable machine learning challenge},
  author={Holter, Steffen and Gomez, Oscar and Bertini, Enrico},
  journal={FICO COmmunity},
  year={2018}
}

@inproceedings{koh2017understanding,
  title={Understanding black-box predictions via influence functions},
  author={Koh, Pang Wei and Liang, Percy},
  booktitle={International conference on machine learning},
  pages={1885--1894},
  year={2017},
  organization={PMLR}
}

@article{barez2025chain,
  title={Chain-of-thought is not explainability},
  author={Barez, Fazl and Wu, Tung-Yu and Arcuschin, Iv{\'a}n and Lan, Michael and Wang, Vincent and Siegel, Noah and Collignon, Nicolas and Neo, Clement and Lee, Isabelle and Paren, Alasdair and others},
  journal={Preprint, alphaXiv},
  pages={v1},
  year={2025}
}

@article{arcuschin2025chain,
  title={Chain-of-thought reasoning in the wild is not always faithful},
  author={Arcuschin, Iv{\'a}n and Janiak, Jett and Krzyzanowski, Robert and Rajamanoharan, Senthooran and Nanda, Neel and Conmy, Arthur},
  journal={arXiv preprint arXiv:2503.08679},
  year={2025}
}

@article{hollmann2025tabpfn,
 title={Accurate predictions on small data with a tabular foundation model},
 author={Hollmann, Noah and M{\"u}ller, Samuel and Purucker, Lennart and
         Krishnakumar, Arjun and K{\"o}rfer, Max and Hoo, Shi Bin and
         Schirrmeister, Robin Tibor and Hutter, Frank},
 journal={Nature},
 year={2025},
 month={01},
 day={09},
 doi={10.1038/s41586-024-08328-6},
 publisher={Springer Nature},
 url={https://www.nature.com/articles/s41586-024-08328-6},
}
